\documentclass[sigconf, nonacm]{acmart}

\usepackage{algorithm}
\usepackage{algorithmic}

\usepackage{amsmath}
\usepackage{amssymb}
\usepackage{mathtools}
\usepackage{amsthm}

\usepackage{adjustbox}
\usepackage{booktabs}
\usepackage{multirow}

\usepackage{microtype}
\usepackage{graphicx}
\usepackage{subfigure}
\usepackage{booktabs} % for professional tables

\usepackage{hyperref}
\usepackage{placeins}

\theoremstyle{plain}
\newtheorem{theorem}{Theorem}[section]
\newtheorem{proposition}[theorem]{Proposition}
\newtheorem{lemma}[theorem]{Lemma}
\newtheorem{corollary}[theorem]{Corollary}
\theoremstyle{definition}
\newtheorem{definition}[theorem]{Definition}

\theoremstyle{remark}

\newcommand{\eg}{\emph{e.g.}}
\newcommand{\ie}{\emph{i.e.}}
\AtBeginDocument{%
  }

\setcopyright{rightsretained}
\copyrightyear{}
\acmYear{}
\acmDOI{}
\acmConference[]{}{}{}
\acmISBN{978-1-4503-XXXX-X/2018/06}

\begin{document}

\title{Post-Hoc FREE Calibrating on Kolmogorov–Arnold Networks}

\author{Wenhao Liang}
\email{w.liang@adelaide.edu.au}
\affiliation{%
  \institution{Adelaide University}
  \city{Adelaide}
  \state{SA}
  \country{AU}
}

\author{Wei Emma Zhang}
% \email{wei.emma.zhang@adelaide.edu.au}
\affiliation{%
  \institution{Adelaide University}
  \city{Adelaide}
  \state{SA}
  \country{AU}
}

\author{Lin Yue}
% \email{wei.emma.zhang@adelaide.edu.au}
\affiliation{%
  \institution{Adelaide University}
  \city{Adelaide}
  \state{SA}
  \country{AU}
}

\author{Miao Xu}
\affiliation{%
  \institution{The University of Queensland}
  \city{Brisbane}
  \state{QLD}
  \country{AU}
}

\author{Olaf Maennel}
% \email{wei.emma.zhang@adelaide.edu.au}
\affiliation{%
  \institution{Adelaide University}
  \city{Adelaide}
  \state{SA}
  \country{AU}
}

\author{Weitong Chen}
% \email{wei.emma.zhang@adelaide.edu.au}
\affiliation{%
  \institution{Adelaide University}
  \city{Adelaide}
  \state{SA}
  \country{AU}
}
% \affiliation{%

% \affiliation{%

% \affiliation{%

% \affiliation{%

% \affiliation{%


\begin{abstract}
  % Kolmogorov-Arnold Networks (KANs) are neural architectures inspired by the Kolmogorov-Arnold representation theorem, featuring B-spline parameterizations for flexible representation learning. However, they often suffer from over- and underconfidence, leading to biased confidence estimates. Through extensive experiments, we identify four critical hyperparameters—Layer Width, Grid Order, Shortcut Function, and Grid Range—that strongly impact calibration. We then investigate various calibration strategies, including our proposed Temperature-Scaled Loss (TSL). Both theoretical and empirical results show that TSL significantly reduces miscalibration across diverse settings. Overall, our findings clarify how architectural and parameter choices affect KAN calibration and establish TSL as a robust, loss-agnostic solution for enhancing confidence reliability in probabilistic predictions involving KANs.
  Kolmogorov-Arnold Networks (KANs) are neural architectures inspired by the Kolmogorov-Arnold representation theorem that leverage B-spline parameterizations for flexible, locally adaptive function approximation. Although KANs can capture complex nonlinearities beyond those modeled by standard Multi-Layer Perceptrons (MLPs), they frequently exhibit miscalibrated confidence estimates—manifesting as overconfidence in dense data regions and underconfidence in sparse areas. In this work, we systematically examine the impact of four critical hyperparameters—Layer Width, Grid Order, Shortcut Function, and Grid Range—on the calibration of KANs. Furthermore, we introduce a novel Temperature-Scaled Loss (TSL) that integrates a temperature parameter directly into the training objective, dynamically adjusting the predictive distribution during learning. Both theoretical analysis and extensive empirical evaluations on standard benchmarks demonstrate that TSL significantly reduces calibration errors, thereby improving the reliability of probabilistic predictions. Overall, our study provides actionable insights into the design of spline-based neural networks and establishes TSL as a robust, loss-agnostic solution for enhancing calibration.

\end{abstract}


% \keywords{Do, Not, Us, This, Code, Put, the, Correct, Terms, for,


\maketitle

% \section{Introduction}
% ACM's consolidated article template, introduced in 2017, provides a
% consistent \LaTeX\ style for use across ACM publications, and
% incorporates accessibility and metadata-extraction functionality
% necessary for future Digital Library endeavors. Numerous ACM and
% SIG-specific \LaTeX\ templates have been examined, and their unique
% features incorporated into this single new template.

% If you are new to publishing with ACM, this document is a valuable
% guide to the process of preparing your work for publication. If you
% have published with ACM before, this document provides insight and
% instruction into more recent changes to the article template.

% The ``\verb|acmart|'' document class can be used to prepare articles
% for any ACM publication --- conference or journal, and for any stage
% of publication, from review to final ``camera-ready'' copy, to the
% author's own version, with {\itshape very} few changes to the source.

\section{Introduction}
\label{intro}
Accurate confidence calibration is essential for safety-critical and high-stakes applications such as medical diagnosis \cite{huang2020atutorial}, autonomous driving \cite{feng2021review}, and risk-sensitive finance \cite{liu2019neural}. Deep neural architectures such as Multi-Layer Perceptrons (MLPs) \cite{goodfellow2016deep} and Kolmogorov-Arnold Networks (KANs) \cite{liu2024kan} are both designed to model complex input-output relationships. While MLPs rely on fixed activation functions (\eg, ReLU or sigmoid) applied at each neuron, KANs relocate learnable activations to the network’s \emph{edges} via parameterized basis functions (\eg, B-splines). This architectural choice endows KANs with increased flexibility to adapt to local variations in the input space, potentially yielding richer function approximations than their fixed-activation counterparts.

Despite their promise, KANs encounter two notable challenges. First, the learnable spline functions can overfit in regions with abundant data while underfitting in sparser regions \cite{hastie2005elements, aguilera2013comparative, perperoglou2019review}, leading to inconsistent predictive quality. Second, the calibration properties of KANs—\ie, the alignment between predicted probabilities and actual outcomes—have not been thoroughly investigated, even though calibration is critical for risk-sensitive applications \cite{guo2017calibration}. Previous work on KANs has primarily focused on accuracy and interpretability in controlled, often physics-based benchmarks, leaving open questions regarding their behavior on widely used datasets such as MNIST \cite{deng2012mnist, cohen2017emnist, xiao2017fashion} and CIFAR-10 \cite{krizhevsky2009learning}.

\begin{figure}[ht]
\vskip 0.2in
\begin{center}
\resizebox{0.7\linewidth}{!}{%
\centerline{\includegraphics[width=\columnwidth]{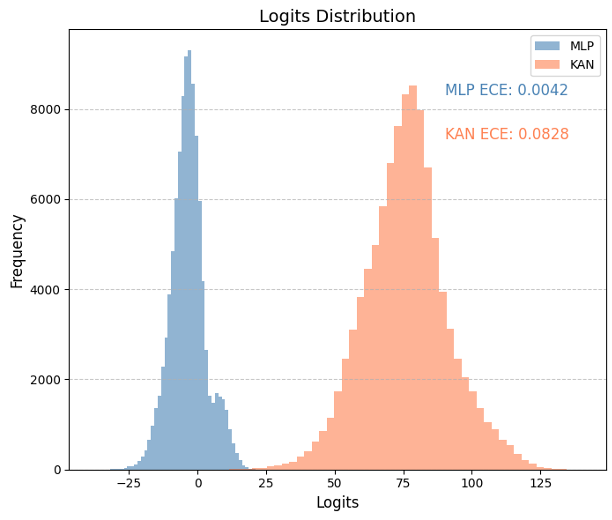}}
}
\caption{Logits distribution between the MLP and KAN models on the MNIST dataset with KAN producing a broader range of logits compared to the more centered logits of MLP.}
\Description{tbc}
\label{fig:logit_mlp_kan}
\end{center}
\vskip -0.2in
\end{figure}

\paragraph{Motivation.}
Our empirical studies (see Table~\ref{tab:model_comparison_limited}) compare MLPs and KANs over a wide range of hyperparameter settings under fair parameter budgets (up to 120k parameters) \cite{yu2024kan}. The results show that MLPs attain higher average accuracy (95.67\% vs.\ 81.09\%) and exhibit relatively lower calibration errors across the measured metrics. By contrast, KANs have more variability in both accuracy and calibration (notably higher standard deviations), indicating a greater susceptibility to over- and underconfidence. As further illustrated in Figure~\ref{fig:logit_mlp_kan}, KANs produce broader logit distributions, which can exacerbate miscalibration. This motivates the development of calibration-enhancing strategies that specifically address KANs' spline-based transformations.

\paragraph{Contributions.}
To address these challenges, we propose integrating a temperature parameter \(\tau\) directly into the training phase via a Temperature-Scaled Loss (TSL). Unlike post-hoc temperature scaling, our TSL approach dynamically adjusts the sharpness of the predictive distribution during learning, thereby mitigating miscalibration as spline functions are updated. Our contributions are:
\begin{itemize}
    \item \textbf{Fair Calibration Analysis.} We perform the first extensive comparison between MLPs and KANs under matched parameter budgets, revealing inherent trade-offs between calibration and performance.
    \item \textbf{Hyperparameter Ablations.} We analyze how key KAN hyperparameters (\eg, \emph{Grid Order}, \emph{Layer Width}, \emph{Shortcut Functions}) affect calibration metrics (ECE, AdaECE, Class-wise ECE, Smooth ECE), and offer practical guidelines.
    \item \textbf{Temperature-Scaled Loss (TSL).} We introduce a unified training framework that incorporates \(\tau\) into standard loss functions (\eg, cross-entropy, Brier score), demonstrating both theoretically and empirically that this integration significantly reduces calibration errors.
\end{itemize}

\begin{table*}[t]
\caption{MLP and KAN models trained on MNIST across various performance metrics for models with parameters in the range [0, 120,000]. All values represent mean $\pm$ standard deviation across 1752 runs with cross entropy.}
\label{tab:model_comparison_limited}
\vskip 0.15in
\begin{center}
\begin{small}
\begin{sc}
\begin{tabular}{lccccccc}
\toprule
Model & Test Accuracy (\%) & ECE & ADAECE & CECE & SMECE & NLL & Train Time \\
\midrule
MLP   & 95.67 $\pm$ 2.71 & 0.046 $\pm$ 0.064 & 0.046 $\pm$ 0.064 & 0.010 $\pm$ 0.013 & 0.047 $\pm$ 0.064 & 0.178 $\pm$ 0.129 & 161.57 $\pm$ 152.62 \\
KAN   & 81.09 $\pm$ 14.31 & 0.066 $\pm$ 0.065 & 0.067 $\pm$ 0.065 & 0.017 $\pm$ 0.013 & 0.065 $\pm$ 0.064 & 0.618 $\pm$ 0.370 & 211.99 $\pm$ 128.66 \\
\bottomrule
\end{tabular}
\end{sc}
\end{small}
\end{center}
\vskip -0.1in
\end{table*}

\section{Related Work}
\label{sec:related_work}

\paragraph{Post-hoc Calibration.}
Modern deep neural networks are known to exhibit overconfidence, leading to probability estimates that often do not reflect true outcome frequencies \cite{guo2017calibration}. Calibration metrics such as the Expected Calibration Error (ECE) and its variants (\eg, Smooth ECE \cite{blasiok2023smooth}) have been developed to quantify this misalignment between confidence and accuracy. Early post-hoc calibration methods, including Platt Scaling \citep{platt1999probabilistic} and Isotonic Regression \cite{niculescu2005predicting}, adjust model outputs after training using a held-out validation set. Temperature Scaling \cite{guo2017calibration} generalizes this idea by uniformly scaling logits, offering a simple yet effective means to improve calibration. However, such post-hoc methods are inherently limited by their reliance on a separate calibration dataset and may struggle to adapt in scenarios with limited data or during distribution shifts \citep{tomani2021post}.

\paragraph{In-training and Temperature Related Calibration.}
To overcome the limitations of post-hoc techniques, several recent studies have integrated calibration directly into the training process. For example, Kumar et al. \cite{kumar2018trainable} introduced a kernel-based Maximum Mean Calibration Error (MMCE) penalty, while Label Smoothing \cite{muller2019does} softens the target distributions to reduce overconfidence. Focal Loss \cite{lin2017focal}, originally proposed for addressing class imbalance, has been adapted to penalize overconfident predictions \cite{charoenphakdee2021focal}, and Dual Focal Loss \cite{tao2023dual} further refines this approach by balancing over- and underconfidence through a margin-maximization term. Most recently, Focal Calibration Loss \cite{liang2024calibrating} combines focal penalties with an euclidean calibration objective, preserving accuracy and improving probability estimates. More recent studies have explored temperature scaling within ensemble methods and uncertainty estimation frameworks. Lakshminarayanan et al.~\cite{lak2017simple} showed that deep ensembles offer scalable, reliable uncertainty estimation with minimal tuning, outperforming Bayesian methods on large-scale tasks. Similarly, Ovadia et al.~\cite{ovadia2019can} showed that traditional post-hoc temperature scaling often fails under dataset shift, highlighting the need for adaptive uncertainty estimation methods for improved robustness. Zhang et al.~\cite{zhang2020mix} proposed Mix-n-Match calibration strategies to enhance post-hoc calibration, addressing limitations of traditional temperature scaling. Kukleva et al.~\cite{kukleva2023temperature} explore dynamic temperature scheduling, demonstrating that adjusting \(tau\) during training can lead to improved representations by balancing instance-level and group-level discrimination . 

\paragraph{Calibration in Kolmogorov-Arnold Networks (KANs).}
Kolmogorov-Arnold Networks (KANs) \cite{liu2024kan} extend traditional architectures by replacing fixed activation functions with spline-based, learnable transformations on network edges. While this design enhances expressive power and interpretability, it also introduces unique calibration challenges. In particular, the flexible spline-based layers can generate logits with broader or more variable distributions, especially when the underlying B-spline grids are coarse or misconfigured \cite{hastie2005elements,aguilera2013comparative,perperoglou2019review}. Our work bridges this gap by systematically analyzing the calibration behavior of KANs and introducing a method that mitigates spline-induced miscalibration during training.

%-------------------------------------------------------------------------
\section{Preliminaries}
\label{sec:preliminaries}

In this section, we lay the theoretical groundwork for our study on calibration and Kolmogorov-Arnold Networks (KANs). We begin by defining key calibration concepts for multi-class classification (\S\ref{subsec:basics_calibration}) and reviewing the widely used post-hoc temperature scaling method (\S\ref{subsec:calibration_scaling}). We then provide an overview of the Kolmogorov-Arnold representation theorem, which motivates the design of KANs (\S\ref{subsec:kan_basics}), and discuss KAN-specific calibration challenges (\S\ref{subsec:kan_calibration_challenges}).

%-------------------------------------------------------------------------
\subsection{Basics of Model Calibration}
\label{subsec:basics_calibration}

Consider a $K$-class classification problem with label \(\mathcal{Y} = \{1, \dots, K\}\) and input space \(\mathcal{X}\). A model learns a function  
\begin{equation}
  g(\mathbf{x}; \theta): \mathcal{X} \to \mathbb{R}^K,
\end{equation}  
which produces a logit vector \(\mathbf{g}(\mathbf{x}) = [g_1(\mathbf{x}), \dots, g_K(\mathbf{x})]^\top\) for each input \(\mathbf{x} \in \mathcal{X}\) and \(\theta\) denotes the set of learnable parameters of the model (\eg, weights and biases). The logits are typically mapped to probabilities using the softmax function:
\begin{equation}
  f(\mathbf{x}) = \sigma\bigl(\mathbf{g}(\mathbf{x}; \theta)\bigr),
\end{equation} 
where \(f(\mathbf{x}) \in \Delta_{K-1}\) lies on the \((K-1)\)-dimensional simplex.

\paragraph{Calibration.}
A model is considered well-calibrated if its predicted probabilities match the true conditional probabilities; that is, among all instances where the model predicts a confidence of 0.8, approximately 80\% of those predictions are correct \cite{guo2017calibration, kumar2019calibration}. For a given input, let $\hat{f}(\mathbf{x}) = \max_k f_k(\mathbf{x})$ denote the top-class confidence and $\hat{y}(\mathbf{x}) = \arg\max_k f_k(\mathbf{x})$ the predicted label. A calibration error quantifies the discrepancy between $\hat{f}(\mathbf{x})$ and the empirical accuracy.

\begin{definition}[Calibration Error, \citealp{guo2017calibration, naeini2015obtaining}]
\label{def:calibration_error}
For a predictor $f$, the top-label calibration error is defined as
\begin{equation}
  \mathrm{CE}(f) = \sqrt{\,
    \mathbb{E}\Bigl[\Bigl(\Pr\bigl(y=\hat{y}(\mathbf{x})\mid\hat{f}(\mathbf{x})\bigr) - \hat{f}(\mathbf{x})\Bigr)^2\Bigr]},
\end{equation}
with $\mathrm{CE}(f)=0$ indicating perfect calibration.
\end{definition}

\paragraph{Expected Calibration Error (ECE)}
Since direct computation of calibration error is challenging, the Expected Calibration Error (ECE) \citep{naeini2015obtaining} is often used as an empirical proxy. By partitioning the confidence range $[0,1]$ into $M$ bins, one defines
\begin{definition}[Expected Calibration Error]
\label{def:ECE}
Let $B_m = \{i \mid c_{m-1} < \hat{f}(\mathbf{x}_i) \le c_m\}$ for bins $(c_{m-1}, c_m]$. Then,
\begin{equation}
  \mathrm{ECE}(f) = \sum_{m=1}^M \frac{|B_m|}{N} \left|\mathrm{acc}_m - \mathrm{conf}_m\right|,
  \label{eq:ECE}
\end{equation}
where 
\[
\mathrm{acc}_m = \frac{1}{|B_m|}\sum_{i\in B_m}\mathbb{I}\{\hat{y}(\mathbf{x}_i)=y_i\}, \quad \mathrm{conf}_m = \frac{1}{|B_m|}\sum_{i\in B_m}\hat{f}(\mathbf{x}_i).
\]
\end{definition}

\paragraph{Smooth Expected Calibration Error.}
To avoid binning artifacts, Smooth Expected Calibration Error (smoothECE) \cite{blasiok2023smooth} computes the calibration error continuously:
\begin{definition}[Smooth ECE, \citealp{blasiok2023smooth}]
\label{def:smoothECE}
Let $u \in [0,1]$ be a confidence level, and define
\begin{equation}
   \mathrm{smoothECE}(f)
   = \int_{0}^{1} \left|\mathrm{acc}(u) - u\right| \,\omega(u)\,du,
   \label{eq:smoothECE}
\end{equation}
where $\mathrm{acc}(u)=\Pr\!\bigl(\hat{y}(\mathbf{x})=y\mid\hat{f}(\mathbf{x})=u\bigr)$ and $\omega(u)$ is a smoothing kernel. Lower values indicate better calibration.
\end{definition}

Additional metrics such as AdaECE and class-wise ECE further dissect calibration performance, and we provide details in Appendix~\ref{appendix: cal_metr}.

%-------------------------------------------------------------------------
\subsection{Post-Hoc Calibration Temperature Scaling}
\label{subsec:calibration_scaling}

Temperature scaling is a widely adopted post-hoc calibration technique that adjusts a model’s logits after training. For a trained model with logits \(\mathbf{g}(\mathbf{x})\), temperature scaling replaces them with $\mathbf{g}(\mathbf{x})/T$, where $T>0$ is a scalar. The calibrated probabilities become:
\begin{equation}
  \varphi(T)\circ f(\mathbf{x})
  = \sigma\!\Bigl(\frac{\mathbf{g}(\mathbf{x})}{T}\Bigr).
  \label{eq:TS}
\end{equation}
An optimal $T$ is typically determined by minimizing the negative log-likelihood (NLL) on a validation set $D_{\mathrm{val}}$:
\begin{equation}
  \min_{T}\; -\sum_{(\mathbf{x}_i,y_i)\in D_{\mathrm{val}}} \ln\!\Bigl(\sigma_{y_i}\!\Bigl(\frac{\mathbf{g}({x}_i)}{T}\Bigr)\Bigr).
  \label{eq:T_opt}
\end{equation}

% Temperature scaling is a widely used post-hoc calibration technique without altering the model weights. For a trained model with logits \(\mathbf{g}(\mathbf{x}) = [g_1(\mathbf{x}), \dots, g_K(\mathbf{x})]^\top\), temperature scaling modifies these logits by dividing them by a positive scalar temperature \(T > 0\). The calibrated probabilities are computed as:  
% \(
%   f_T(\mathbf{x}) = \sigma\!\Bigl(\frac{\mathbf{g}(\mathbf{x})}{T}\Bigr),
%   \quad \text{where } f_T(\mathbf{x}) \in \Delta_{K-1}.
% \) 
% This is expressed explicitly as:  
% \(
%   f_{T,k}(\mathbf{x}) = \frac{\exp\bigl(g_k(\mathbf{x}) / T\bigr)}{\sum_{j=1}^K \exp\bigl(g_j(\mathbf{x}) / T\bigr)}.
% \)
% The optimal temperature \(T^*\) is typically determined by minimizing the negative log-likelihood (NLL) on a held-out validation set \(D_{\mathrm{val}} = \{(\mathbf{x}_i, y_i)\}\):
% \[
%   \min_{T} \; -\sum_{(\mathbf{x}_i, y_i) \in D_{\mathrm{val}}} \ln\!\Bigl(f_{T, y_i}(\mathbf{x}_i)\Bigr),
% \]  
% where \(f_{T, y_i}(\mathbf{x}_i)\) denotes the predicted probability assigned to the correct label \(y_i\) under the temperature-scaled distribution.
This procedure sharpens ($T<1$) or flattens ($T>1$) the output distribution while keeping the original model weights unchanged.

%-------------------------------------------------------------------------
\subsection{Kolmogorov-Arnold Representation KANs}
\label{subsec:kan_basics}

\paragraph{Kolmogorov’s Superposition Theorem.}
Kolmogorov’s superposition theorem \cite{kolmogorov1957representation} asserts that any multivariate continuous function defined on a bounded domain can be represented as finite compositions of univariate continuous functions and summations. This foundational result underpins the design of Kolmogorov-Arnold Networks (KANs) \cite{liu2024kan}.

\paragraph{KAN Layer Structure.}
A single KAN layer transforms an input vector $\mathbf{x}_l \in \mathbb{R}^{n_l}$ to an output vector $\mathbf{x}_{l+1} \in \mathbb{R}^{n_{l+1}}$ via:
\begin{equation}
  x_{l+1,i} = \sum_{j=1}^{n_l} \phi_{l,i,j}\bigl(x_{l,j}\bigr), \quad \forall\, i=1,\dots,n_{l+1},
  \label{eq:KAN_layer}
\end{equation}
where each $\phi_{l,i,j}(\cdot)$ is a learnable univariate function, typically parameterized by B-splines. Stacking $L$ such layers yields the overall network:
\begin{equation}
\label{eq:kan_composed}
  \mathbf{g} = \mathrm{KAN}(\mathbf{x}_{0}) = \bigl(\mathbf{\Phi}_{L-1} \circ \cdots \circ \mathbf{\Phi}_{0}\bigr)\,\mathbf{x}_{0}.
\end{equation}
For classification tasks, a softmax activation is applied to $\mathbf{g}$ to obtain probability estimates.

%-------------------------------------------------------------------------
\subsection{KAN-Specific Calibration Challenges}
\label{subsec:kan_calibration_challenges}

\paragraph{Grid-Induced Variability.}
Unlike MLPs, KANs employ B-spline functions that rely on a predefined grid of knots to approximate nonlinearities. If the grid is too coarse or improperly configured, it can lead to “grid bias” in the learned functions, resulting in overfitting in data-dense regions and underfitting in sparse regions \cite{hastie2005elements,aguilera2013comparative,perperoglou2019review}. This phenomenon often produces a broader or more erratic logit distribution, as seen in Figure~\ref{fig:logit_mlp_kan}.

\begin{proposition}[Spline Order and Calibration Error]
\label{prop:spline_calib}
Let \( \phi_{l,i,j} \) be a B-spline of order \( s \) with \( G \) knots. For a fixed number of knots \( G \), the variance of the logits, \( \mathbb{V}[\theta(\mathbf{x})] \), tends to increase with the spline order \( s \), which in turn exacerbates the ECE (see proof in Appendix~\ref{app:proof_order_ece}).
\end{proposition}

\begin{figure}[ht]
\vskip 0.2in
\begin{center}
\includegraphics[width=\columnwidth]{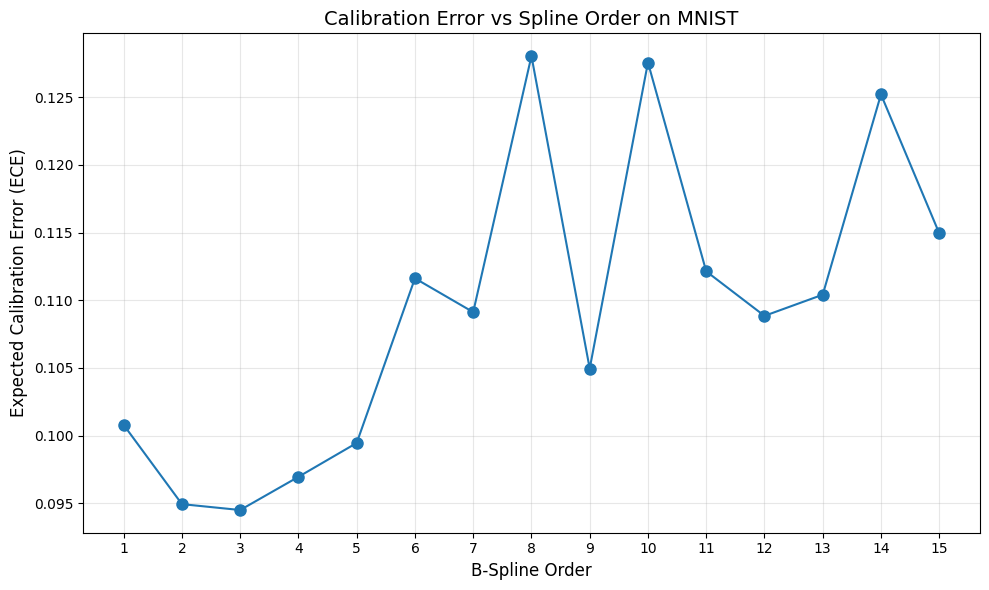}
\caption{\textbf{ECE vs. Spline Order on MNIST}}
\Description{tbc}
\label{fig:ece_spline_order_MNIST}
\end{center}
\vskip -0.2in
\end{figure}

\paragraph{Overconfidence and Underfitting.}
The design of the B-spline grids introduces two primary calibration risks:
\begin{itemize}
  \item \textbf{Overconfidence in Sparse Regions:} In areas with low sample density, coarse grids can lead to abrupt extrapolation, yielding overly confident predictions.
  \item \textbf{Underfitting in Dense Regions:} Conversely, in regions with abundant data, an over-regularized spline (high spline order with few knots) may fail to capture local variations, resulting in underconfident predictions.
\end{itemize}
% Empirical evidence in Figure~\ref{fig:ece_spline_order_MNIST} shows that, on MNIST, ECE increases by approximately 35\% as the spline order grows from 3 to 8, supporting Proposition~\ref{prop:spline_calib}.
\paragraph{Empirical Validation.} 
Figure~\ref{fig:ece_spline_order_MNIST} shows ECE vs. Spline Order on MNIST using KANs (G=5). ECE increases by 34.98\% as s grows from 3 to 8, confirming the Proposition~\ref{prop:spline_calib}.

\begin{figure*}[ht]
\vskip 0.2in
\begin{center}
\centerline{\includegraphics[width=\linewidth]{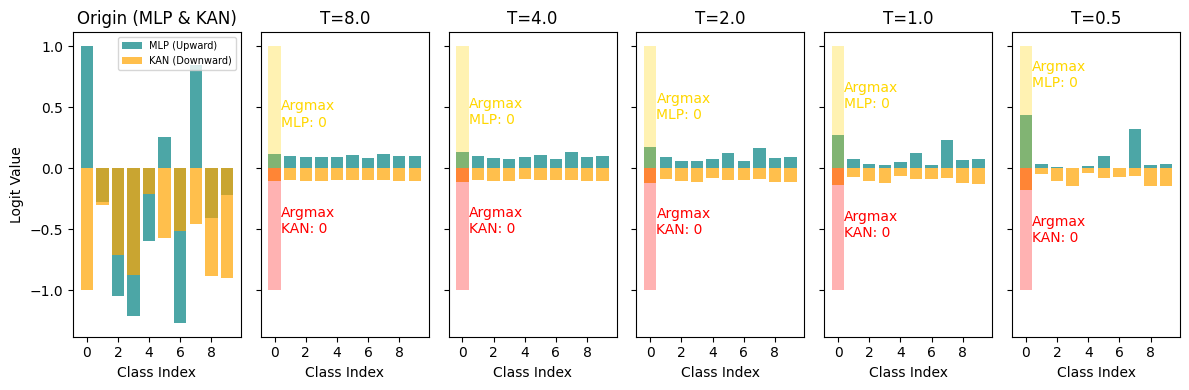}}
\caption{Visualization of temperature scaling applied to MLP (upward) and KAN (downward) logits for different temperature values. The first plot displays the original logits for both models. Each subsequent plot shows the probability distributions scaled by temperatures T=8.0,4.0,2.0,1.0,0.5, with the respective argmax classes highlighted. Gold/Red bars indicate the MLP/KAN argmax. \textbf{Higher \(T\)} simulates noisier distributions, encouraging robustness to uncertainty.
While \textbf{Lower \(T\)} focuses on sharpening predictions, reducing overconfidence.}
\Description{tbc}
\label{temp_mlp_kan}
\end{center}
\vskip -0.2in
\end{figure*}

\paragraph{Visualizing Temperature Effects.}
Figure~\ref{temp_mlp_kan} illustrates how temperature scaling affects the logit distributions for both MLPs and KANs. Higher temperature values yield noisier, more distributed logits, which can help mitigate overconfidence, while lower values sharpen the predictions, reducing unwarranted confidence.

\begin{figure*}[ht]
\vskip 0.2in
\begin{center}
\centerline{\includegraphics[width=\linewidth]{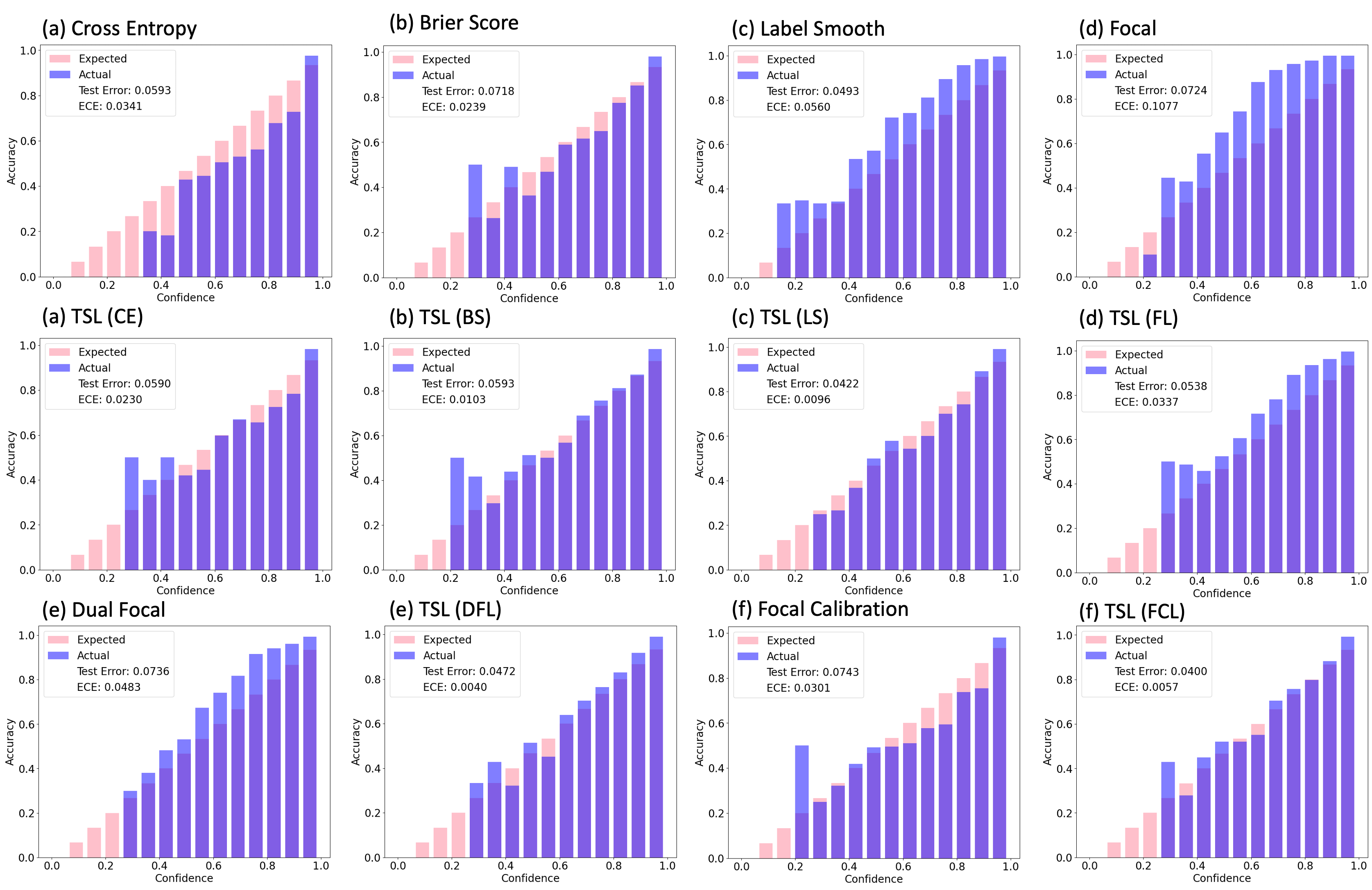}}
\caption{Reliability Diagram for KAN Model: Evaluation of calibration performance for a KAN model trained with 8 SOTA loss functions on the MNIST dataset under identical hyperparameter settings.}
\label{fig: 6losses_kan}
\Description{tbc}
\end{center}
\vskip -0.2in
\end{figure*}

\paragraph{Loss Function Comparisons.}
In Figure~\ref{fig: 6losses_kan}, we present reliability diagrams for a KAN model trained with 6 state-of-the-art loss functions on the MNIST dataset. The diagram underscores the variability in calibration performance under different loss functions, further motivating the need for a unified calibration approach like our proposed Temperature-Scaled Loss (TSL).

\begin{figure}[ht]
\vskip 0.2in
\begin{center}
\includegraphics[width=\columnwidth]{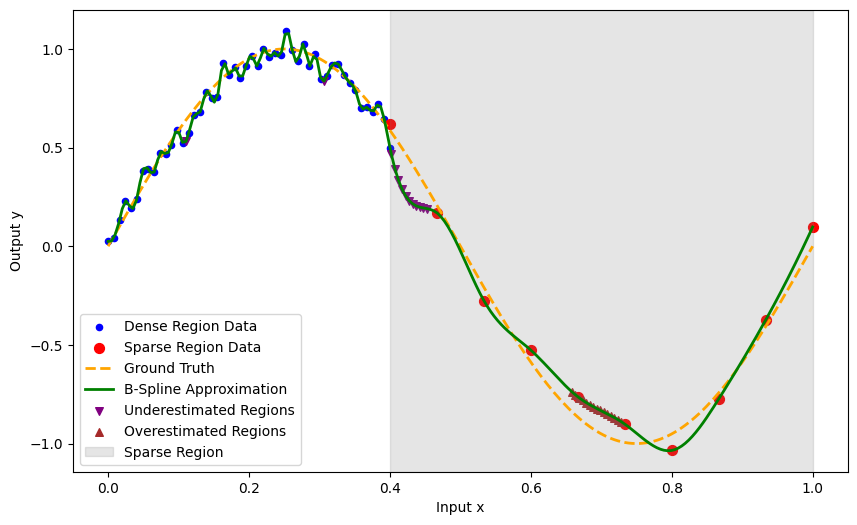}
\caption{\textbf{B-Spline Approximation in Dense vs. Sparse Regions.} A B-spline (green) approximates a sine wave (orange, dashed), with $\mathbf{x}\in[0,1]$ subdivided into a dense region $\mathbf{x}\in[0,0.4]$ and a sparse region $\mathbf{x}\in[0.4,1.0]$ (shaded in gray). Over- or under-smoothing can arise from uneven grid usage.}
\Description{tbc}
\label{bspline_oc_uc}
\end{center}
\vskip -0.2in
\end{figure}

% \begin{figure*}[ht]
% \vskip 0.2in
% \begin{center}
% \resizebox{\textwidth}{!}{%
% \begin{tabular}{ccccc}
%     \subfigure[Layer Width]{
%         \includegraphics[width=0.18\linewidth]{imgs/ece_width.png}
%     } &
%     \subfigure[Grid Range]{
%         \includegraphics[width=0.18\linewidth]{imgs/ece_range.png}
%     } &
%     \subfigure[Gird Order]{
%         \includegraphics[width=0.18\linewidth]{imgs/ece_order.png}
%     } &
%     \subfigure[Shortcut]{
%         \includegraphics[width=0.18\linewidth]{imgs/ece_shortcut.png}
%     } &
%     \subfigure[Number Params($10^4$)]{
%         \includegraphics[width=0.18\linewidth]{imgs/ece_param.png}
%     }
% \end{tabular}
% }
% \caption{KANs Key Parameters vs. ECE.}
% \Description{tbc}
% \label{fig:multiple_factors}
% \end{center}
% \vskip -0.2in
% \end{figure*}

\section{Methodology}
\label{sec:methodology}

In this section, we formalize our supervised classification setup (\S\ref{subsec:problem_setup}), review the standard post-hoc \emph{Temperature Scaling} method (\S\ref{subsec:temp_scaling}), and introduce our novel \emph{Temperature-Scaled Loss (TSL)} that integrates the temperature parameter directly into the training objective (\S\ref{subsec:tsl}). We conclude with a detailed description of the algorithmic procedure for TSL (\S\ref{subsec:alg_tsl}).

%-------------------------------------------------------------------------
\subsection{Problem Setup}
\label{subsec:problem_setup}

Consider a supervised classification problem with training samples 
\begin{equation}
  \bigl\{ (\mathbf{x}_i, y_i) \bigr\}_{i=1}^N,
\end{equation}
where each feature vector \(\mathbf{x}_i \in \mathbb{R}^d\) and label \(y_i \in \{1, \dots, K\}\). Let \(f(\mathbf{x};\,\theta)\) denote a neural network (parameterized by \(\theta\)) that produces a logit vector \(\mathbf{g}_i \in \mathbb{R}^K\) for input \(\mathbf{x}_i\). The corresponding class probabilities are obtained by applying the softmax function:
\begin{equation}
  \hat{p}_{ik} \;=\; \frac{\exp\!\bigl(g_{ik}\bigr)}{\sum_{j=1}^K \exp\!\bigl(g_{ij}\bigr)}, \quad k=1,\dots,K,
  \label{eq:prob_softmax}
\end{equation}
where \(g_{ik}\) denotes the \(k\)-th component of \(\mathbf{g}_i\).

%-------------------------------------------------------------------------
\subsection{Temperature Scaling (TS)}
\label{subsec:temp_scaling}

\emph{Temperature Scaling} (TS) \cite{guo2017calibration} is a widely used post-hoc calibration method. Given the original logits \(\mathbf{g}_i\) from a trained model, TS rescales them with a positive scalar \(\tau>0\):
\begin{equation}
  \tilde{p}_{ik} \;=\; \frac{\exp\!\bigl(g_{ik}/\tau\bigr)}{\sum_{j=1}^K \exp\!\bigl(g_{ij}/\tau\bigr)}, \quad k = 1, \dots, K.
  \label{eq:ts_definition}
\end{equation}
A larger \(\tau\) produces a flatter (more uniform) distribution, thereby reducing overconfidence, whereas a smaller \(\tau\) sharpens the distribution, potentially mitigating underconfidence.

\paragraph{Optimal Temperature.}
In standard post-hoc TS, the network parameters \(\theta\) are fixed and only \(\tau\) is optimized on a held-out validation set \(\mathcal{D}_{\mathrm{val}}\) by solving:
\begin{equation}
  \tau^{\ast} \;=\; \arg\min_{\tau>0}\; -\sum_{(\mathbf{x}_i,y_i)\in\mathcal{D}_{\mathrm{val}}} \ln\!\Bigl(\mathrm{Softmax}\bigl(\mathbf{g}_i/\tau\bigr)_{y_i}\Bigr).
  \label{eq:ts_optimal}
\end{equation}
Although effective, this approach does not allow the network to adjust its internal representations for accuracy and calibration.

%-------------------------------------------------------------------------
\subsection{Temperature-Scaled Loss (TSL)}
\label{subsec:tsl}

To address the limitations of post-hoc TS, we propose \emph{Temperature-Scaled Loss (TSL)}, which incorporates the temperature parameter \(\tau\) as a trainable variable within the learning process. In contrast to post-hoc methods, TSL updates both the network parameters \(\theta\) and the temperature \(\tau\) simultaneously.

Let \(\mathcal{L}_{\mathrm{base}}\) denote a standard training loss (\eg, cross-entropy, focal loss). For each input \(\mathbf{x}_i\) with logits \(\mathbf{g}_i = f(\mathbf{x}_i;\,\theta)\), we define the rescaled logits as:
\begin{equation}
  \tilde{\mathbf{g}}_i \;=\; \frac{\mathbf{g}_i}{\tau}.
  \label{eq:tsl_logits}
\end{equation}
The Temperature-Scaled Loss is then given by
\begin{equation}
  \mathcal{L}_{\mathrm{TSL}} \;=\; \mathcal{L}_{\mathrm{base}}\!\Bigl(\tilde{\mathbf{g}},\,y\Bigr).
  \label{eq:tsl_def}
\end{equation}
Minimizing \(\mathcal{L}_{\mathrm{TSL}}\) allows the parameters \(\theta\) and \(\tau\) to co-adapt, thereby yielding logits that are better aligned with the true class probabilities.

\paragraph{Joint Optimization of \(\tau\) and \(\theta\).}
We update \(\tau\) along with \(\theta\) using backpropagation. For instance, a gradient descent update on \(\tau\) is performed as
\begin{equation}
  \tau \;\gets\; \tau - \eta_{\tau}\,\frac{\partial \mathcal{L}_{\mathrm{TSL}}}{\partial \tau},
  \label{eq:tsl_grad}
\end{equation}
where \(\eta_{\tau}\) is the learning rate for \(\tau\). A simple projection (\eg, \(\tau \leftarrow \max(\varepsilon, \tau)\) for some small \(\varepsilon>0\)) ensures that \(\tau\) remains strictly positive.

\paragraph{Benefits for Calibration.}
Incorporating \(\tau\) as a trainable parameter enables the model to:
(i) Dynamically penalize overly peaked (overconfident) or excessively flat (underconfident) output distributions during training. (ii) Adjust gradient updates such that miscalibrated predictions incur a higher loss, thus guiding \(\tau\) (and \(\theta\)) toward reducing calibration error. (iii) Eliminate the need for a separate post-hoc calibration step, as the network inherently learns to produce well-calibrated probabilities.

%-------------------------------------------------------------------------
\subsection{Algorithmic Steps for TSL}
\label{subsec:alg_tsl}

Algorithm~\ref{alg:tsl} details the TSL procedure. For each minibatch, the algorithm computes the logits, rescales them by \(\tau\), computes the Temperature-Scaled Loss, and then updates both \(\theta\) and \(\tau\) via backpropagation.

\begin{algorithm}[ht]
\caption{Temperature-Scaled Loss (TSL)}
\label{alg:tsl}
\begin{algorithmic}[1]
   \STATE \textbf{Input:} Model \(f(\cdot;\,\theta)\), dataset \(\mathcal{D}=\{(\mathbf{x}_i,y_i)\}_{i=1}^{N}\), base loss \(\mathcal{L}_{\text{base}}\), initial parameters \(\theta_0\) and \(\tau_0>0\), learning rates \(\eta\) and \(\eta_\tau\), number of epochs \(E\).
   \STATE \textbf{Output:} Learned parameters \((\theta,\,\tau)\).
   \STATE Initialize \((\theta,\,\tau) \gets (\theta_0,\,\tau_0)\).
   \FOR{epoch \(=1\) to \(E\)}
       \STATE Shuffle and partition \(\mathcal{D}\) into minibatches \(\{\mathcal{B}_k\}_{k=1}^M\).
       \FOR{each minibatch \(\mathcal{B}_k\)}
          \STATE \textbf{(i) Compute logits:} For each \((\mathbf{x}_i,y_i)\in \mathcal{B}_k\), set
          \[
            \mathbf{g}_i \gets f(\mathbf{x}_i;\,\theta).
          \]
          \STATE \textbf{(ii) Rescale logits:} Compute
          \(
            \tilde{\mathbf{g}}_i \gets \frac{\mathbf{g}_i}{\tau}.
          \)
          \STATE \textbf{(iii) Compute TSL:} Evaluate
          \(
            \mathcal{L}_{\mathrm{TSL}} \gets \mathcal{L}_{\mathrm{base}}\!\bigl(\tilde{\mathbf{g}},\,y\bigr).
          \)
          \STATE \textbf{(iv) Backpropagate and update parameters:}
          \[
            (\nabla_{\theta}, \nabla_{\tau}) \gets \nabla_{\theta,\tau}\,\mathcal{L}_{\mathrm{TSL}},
          \]
          \[
            \theta \gets \theta - \eta\,\nabla_{\theta}\mathcal{L}_{\mathrm{TSL}}, \quad
            \tau \gets \tau - \eta_{\tau}\,\nabla_{\tau}\mathcal{L}_{\mathrm{TSL}}.
          \]
          \STATE \textbf{(v) Enforce \(\tau > 0\):} Update
          \(
            \tau \gets \max(\varepsilon, \tau).
          \)
       \ENDFOR
   \ENDFOR 
   \STATE \RETURN \((\theta,\,\tau)\)
\end{algorithmic}
\end{algorithm}

\section{Theoretical Evidence}
\label{sec:theoretical_evidence}

In this section, we establish key theoretical properties of TSL. We first show that TSL preserves the strict properness of the base loss (\S\ref{subsec:strict_properness}). Next, we demonstrate that the gradient with respect to \(\tau\) adjusts for over- and underconfidence (\S\ref{subsec:gradient_adjustments}). Finally, we present local convergence guarantees along with a reduction in calibration error (\S\ref{subsec:convergence_calibration} and \S\ref{subsec:cor_reduction_ece}). Additional proofs and extensions (\eg, using Riemann–Stieltjes integration and maximum entropy arguments) are provided in Appendix~\ref{appendix:extended_theory}.

\subsection{Preservation of Strict Properness}
\label{subsec:strict_properness}

\begin{proposition}
\label{prop:tsl_strict_properness}
Let \(\mathcal{L}_{\mathrm{base}}(p,y)\) be a strictly proper scoring rule, where \(p\) represents the predicted probability vector and \(y\) the true label \citep{kumar2018trainable, guo2017calibration}. Define
\begin{equation}
  \mathcal{L}_{\mathrm{TSL}}(\theta,\tau)
  \;=\;
  \mathcal{L}_{\mathrm{base}}\!\Bigl(\mathrm{softmax}\!\Bigl(\frac{g(\theta)}{\tau}\Bigr),\,y\Bigr), \quad \tau>0.
\end{equation}
Then \(\mathcal{L}_{\mathrm{TSL}}\) remains strictly proper with respect to the true conditional distribution \(\Pr(Y\mid X)\). See Appenidx~\ref{appendix:proof_strict_properness} for detail.
\end{proposition}

\subsection{\texorpdfstring{Gradient-Based Adjustments of $\tau$}{Gradient-Based Adjustments of tau}}

\label{subsec:gradient_adjustments}

\begin{lemma}[Monotonic Gradient Updates]
\label{lemma:tau_updates_correct_miscalibration}
Consider the softmax probabilities defined as
\begin{equation}
  \hat{p}_{ik}(\theta,\tau)
  \;=\; 
  \frac{\exp\!\bigl(g_{ik}(\theta)/\tau\bigr)}
       {\sum_{j=1}^K \exp\!\bigl(g_{ij}(\theta)/\tau\bigr)},
\end{equation}
and let
\[
\mathcal{L}_{\mathrm{TSL}}(\theta,\tau) = \mathcal{L}_{\mathrm{base}}\bigl(\hat{p}(\theta,\tau),y\bigr).
\]
Then the gradient \(\frac{\partial \mathcal{L}_{\mathrm{TSL}}}{\partial \tau}\) is such that if \(\hat{p}_{ik}(\theta,\tau)\) exceeds the true label indicator \(y_{ik}\) (indicating overconfidence), the update pushes \(\tau\) upward; conversely, if \(\hat{p}_{ik}(\theta,\tau)\) is too low (indicating underconfidence), the update pushes \(\tau\) downward. (See Appendix~\ref{appendix:gradient_adjustments} for details.)
\end{lemma}

\subsection{Local Convergence and Calibration}
\label{subsec:convergence_calibration}

\begin{theorem}[Local Convergence and Calibration Improvement~\citep{ghadimi2013stochastic}]
\label{thm:tsl_local_minimum}
Assume that \(\mathcal{L}_{\mathrm{TSL}}(\theta,\tau)\) is continuous, differentiable, and bounded. Let \(\{(\theta_t,\tau_t)\}\) be the sequence of parameters generated by (stochastic) gradient descent with an appropriate learning rate. Then, under standard regularity conditions, \((\theta_t,\tau_t)\) converges to a local minimum of \(\mathcal{L}_{\mathrm{TSL}}\). Moreover, the resulting model is at least as well-calibrated as a model trained with fixed temperature \(\tau=1\).
\end{theorem}

\paragraph{Local vs. Global Minima.}
While Theorem~\ref{thm:tsl_local_minimum} guarantees convergence to a local minimum, we do not claim global optimality. Empirically, local minima in deep networks often yield strong calibration performance. (See proof in Appendix~\ref{appendix:convergence_calibration}.)

\subsection{Reduction in Expected Calibration Error}
\label{subsec:cor_reduction_ece}

\begin{corollary}[Reduction of ECE]
\label{cor:ece_reduction}
Let \(\mathrm{acc}(B_i)\) denote the accuracy and \(\mathrm{conf}(B_i)\) the average confidence in bin \(B_i\) for an unscaled model, and let \(\mathrm{conf}(B_i;\tau_i)\) denote the corresponding value after temperature scaling. Then, for each bin \(B_i\),
\begin{equation}
  \Bigl|\mathrm{acc}(B_i) - \mathrm{conf}(B_i;\tau_i)\Bigr|
  \;\le\;
  \Bigl|\mathrm{acc}(B_i) - \mathrm{conf}(B_i)\Bigr|.
\end{equation}
Summing over all bins, we have
\begin{equation}
  \mathrm{ECE}_{\mathrm{TSL}} \;\le\; \mathrm{ECE}_{\mathrm{base}},
\end{equation}
where \(\mathrm{ECE}_{\mathrm{TSL}}\) and \(\mathrm{ECE}_{\mathrm{base}}\) denote calibration errors under temperature-scaled and unscaled training, respectively. (See proof in Appendix~\ref{appendix:ece_reduction}.)
\end{corollary}

%-------------------------------------------------------------------------
% (Include tables and figures here as needed)
\begin{figure}[]
\vskip 0.2in
\begin{center}
\resizebox{\linewidth}{!}{%
\includegraphics[width=\columnwidth]{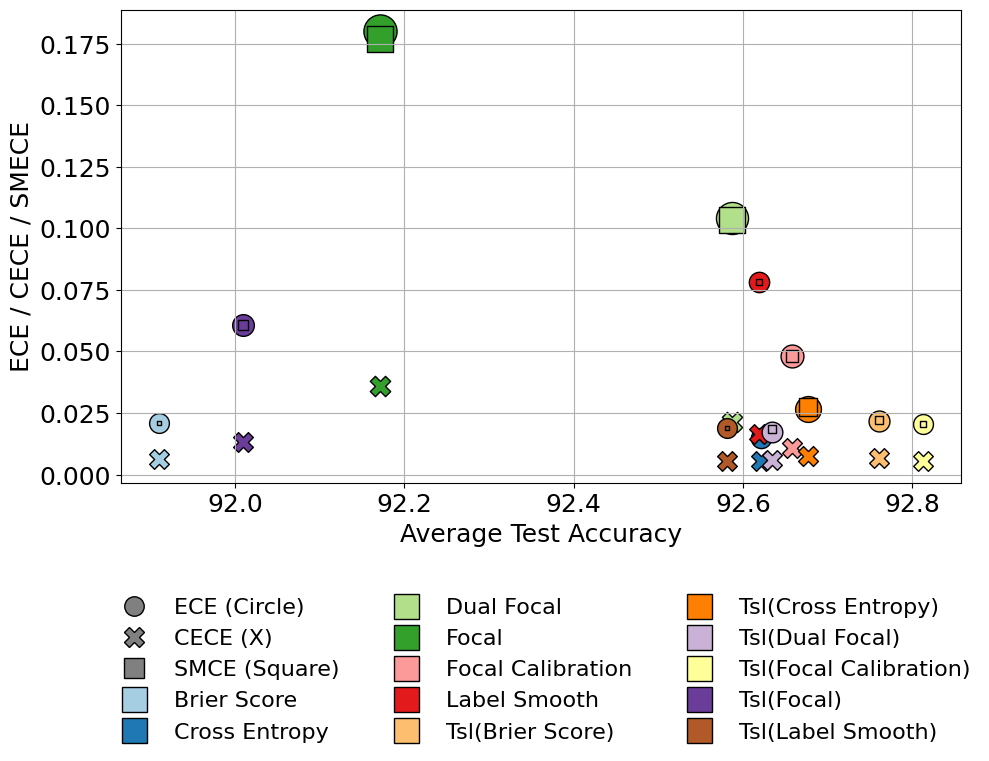}
}
\caption{\textbf{Calibration Metrics vs. Average Test Accuracy.} Marker size indicates variance. Lower calibration errors and higher accuracies suggest more effective training.}
\Description{tbc}
\label{fig:ece_loss}
\end{center}
\vskip -0.2in
\end{figure}

\begin{table*}[t]
\caption{Comparison of test accuracy and calibration metrics across different loss functions on MNIST dataset.\protect\footnotemark[1]}
\label{tab: loss-comparison-table}
% \vskip 0.15in
\begin{center}
\begin{small}
\begin{sc}
\resizebox{\textwidth}{!}{%
\begin{tabular}{lcccccccccc}
\toprule
Loss & *Acc\(\uparrow\) & †Acc\(\uparrow\) & *ECE\(\downarrow\) & †ECE\(\downarrow\) & *CECE\(\downarrow\) & †CECE\(\downarrow\) & *AECE\(\downarrow\) & †AECE\(\downarrow\) & *SMECE\(\downarrow\) & †SMECE\(\downarrow\) \\

\midrule
CE\cite{guo2017calibration}           & 92.62 $\pm$ 2.04 & 96.38 & 1.50 $\pm$ 0.12 & 0.28 & 0.50 $\pm$ 0.03 & 0.28 & 1.40 $\pm$ 0.12 & 0.18 & 1.60 $\pm$ 0.11 & 0.66 \\
BS\cite{brier1950verification}           & 91.91 $\pm$ 1.96 & 94.18 & 2.20 $\pm$ 2.67 & 0.87 & 0.70 $\pm$ 0.07 & 0.40 & 2.10 $\pm$ 2.65 & 0.83 & 2.20 $\pm$ 2.51 & 1.04 \\
FL\cite{lin2017focal}           & 92.16 $\pm$ 2.11 & 95.79 & 18.00 $\pm$ 3.69 & 3.32 & 3.60 $\pm$ 1.37 & 0.81 & 18.00 $\pm$ 3.69 & 3.31 & 17.70 $\pm$ 3.15 & 3.30 \\
LS\cite{szegedy2016rethinking}           & 92.62 $\pm$ 2.28 & 96.23 & 7.80 $\pm$ 1.50 & 4.63 & 1.60 $\pm$ 0.05 & 1.07 & 7.80 $\pm$ 1.51 & 4.58 & 7.80 $\pm$ 1.50 & 4.59 \\
DFL\cite{tao2023dual}          & 92.58 $\pm$ 2.05 & 96.42 & 10.40 $\pm$ 3.32 & 1.81 & 2.10 $\pm$ 1.25 & 0.48 & 10.40 $\pm$ 3.33 & 1.59 & 10.30 $\pm$ 3.14 & 1.75 \\
FCL\cite{liang2024calibrating}          & 92.66 $\pm$ 2.18 & 96.29 & 4.80 $\pm$ 0.72 & 0.55 & 1.10 $\pm$ 0.24 & 0.28 & 4.80 $\pm$ 0.72 & 0.39 & 4.80 $\pm$ 0.72 & 0.70 \\
TSL(CE)      & 92.68 $\pm$ 2.57 & \textbf{96.44} & 2.70 $\pm$ 1.46 & 0.25 (+11.97\%) & 0.70 $\pm$ 0.48 & 0.27 (+2.74\%) & 2.60 $\pm$ 1.47 & \textbf{0.18 (+0.01\%)} & 2.70 $\pm$ 1.40 & 0.62 (+4.89\%) \\
TSL(BS)      & 92.76 $\pm$ 0.83 & 96.30 & 2.10 $\pm$ 0.25 & 0.32 (+63.40\%) & 0.60 $\pm$ 0.01 & 0.26 (+34.83\%) & 2.10 $\pm$ 0.26 & 0.27 (+67.32\%) & 2.10 $\pm$ 0.23 & \textbf{0.55 (+46.58\%)} \\
TSL(FL)      & 92.01 $\pm$ 2.01 & 95.19 & 6.10 $\pm$ 0.44 & 3.27 (+1.25\%) & 1.30 $\pm$ 0.15 & 0.73 (+9.80\%) & 6.10 $\pm$ 0.44 & 3.17 (+4.26\%) & 6.10 $\pm$ 0.44 & 3.20 (+2.78\%) \\
TSL(LS)      & 92.58 $\pm$ 2.09 & 95.78 & 1.90 $\pm$ 0.25 & 0.83 (+82.03\%) & 0.50 $\pm$ 0.01 & 0.31 (+70.79\%) & 1.80 $\pm$ 0.25 & 0.67 (+85.43\%) & 1.90 $\pm$ 0.22 & 0.88 (+80.88\%) \\
TSL(DFL)     & 92.63 $\pm$ 2.31 & 95.99 & 1.70 $\pm$ 2.22 & \textbf{0.22 (+87.89\%)} & 0.60 $\pm$ 0.05 & 0.27 (+42.96\%) & 1.70 $\pm$ 2.19 & 0.21 (+86.60\%) & 1.90 $\pm$ 1.96 & 0.58 (+66.97\%) \\
TSL(FCL)     & 92.81 $\pm$ 2.37 & 96.18 & 2.10 $\pm$ 0.47 & 0.42 (+22.43\%) & 0.60 $\pm$ 0.02 & \textbf{0.25 (+10.12\%)} & 2.00 $\pm$ 0.47 & 0.23 (+40.63\%) & 2.10 $\pm$ 0.40 & 0.63 (+10.10\%) \\
\bottomrule

\end{tabular}
} % Resizebox end
\end{sc}
\end{small}
\end{center}
\vskip -0.1in
% \begin{flushleft}
% \footnotesize{
% \textbf{Note:} `*` denotes the average metric (mean $\pm$ var), and `†` denotes the best metric. All values are in percentage. Improvements in calibration metrics for TSL losses are shown in parentheses compared with its base losses.
% }
% \end{flushleft}
\end{table*}
\footnotetext[1]{
`*` denotes the average metric (mean $\pm$ var), and `†` denotes the best metric. All values are in percentage. Improvements in calibration metrics for TSL losses are shown in parentheses compared with its base losses.
}

\begin{table*}[ht]
\begin{small}
\centering
\caption{Comparison of test accuracy and calibration metrics across different loss functions on the MNIST datasets.}
\label{tab:combined_results}
\begin{adjustbox}{width=\textwidth}
\begin{tabular}{llccccc}
\toprule
\textbf{Dataset} & \textbf{Loss Function} & \textbf{†Test Acc} & \textbf{†ECE} & \textbf{†AdaECE} & \textbf{†CECE} & \textbf{†SMECE} \\
\midrule
\multirow{8}{*}{EMNIST-Balanced} 
    % & brier\_score & 2.1277 & 0.0004 & 0.0000 & 0.0001 & 0.0004 \\
    % & cross\_entropy & 72.9840 & 0.1316 & 0.1316 & 0.0063 & 0.1315 \\
    & Dual Focal~\citep{tao2023dual}  & 74.2128 & 0.1942 & 0.1942 & 0.0088 & 0.1922 \\
    & Focal Loss~\citep{lin2017focal}  & 72.3883 & 0.2339 & 0.2339 & 0.0102 & 0.2271 \\
    & Focal Calibration Loss~\citep{liang2024calibrating} & 73.5691 & 0.2053 & 0.2053 & 0.0091 & 0.2022 \\
    & Label Smooth~\citep{szegedy2016rethinking} & 72.1277 & 0.1831 & 0.1831 & 0.0086 & 0.1818 \\
    % & tsl(brier\_score) & 2.1277 & 0.0002 (36.54\%) & 0.0000 & 0.0001 (26.25\%) & 0.0002 (36.54\%) \\
    % & tsl(cross\_entropy) & 69.6436 & 0.0712 (45.92\%) & 0.0712 (45.93\%) & 0.0041 (34.92\%) & 0.0713 (45.79\%) \\
    & TSL(Dual Focal) & 72.5904 & 0.0145 (92.52\%) & 0.0159 (91.81\%) & 0.0028 (68.13\%) & 0.0151 (92.14\%) \\
    & TSL(Focal Loss) & 68.5479 & 0.0497 (78.77\%) & 0.0499 (78.69\%) & 0.0033 (67.88\%) & 0.0495 (78.22\%) \\
    & TSL(Focal Calibration Loss) & 72.9574 & 0.0387 (81.13\%) & 0.0387 (81.13\%) & 0.0029 (68.43\%) & 0.0388 (80.82\%) \\
    & TSL(Label Smooth) & 68.6915 & 0.0595 (67.53\%) & 0.0591 (67.72\%) & 0.0036 (57.81\%) & 0.0592 (67.44\%) \\
\midrule
\multirow{8}{*}{EMNIST-Letters} 
    % & brier\_score & 3.8462 & 0.0098 & 0.0000 & 0.0109 & 0.0098 \\
    % & cross\_entropy & 79.7260 & 0.0977 & 0.0977 & 0.0076 & 0.0977 \\
    & Dual Focal~\citep{tao2023dual}  & 81.3606 & 0.1747 & 0.1746 & 0.0129 & 0.1739 \\
    & Focal Loss~\citep{lin2017focal}  & 79.3173 & 0.2188 & 0.2188 & 0.0157 & 0.2144 \\
    & Focal Calibration Loss~\citep{liang2024calibrating} & 79.7885 & 0.1829 & 0.1829 & 0.0135 & 0.1819 \\
    & Label Smooth~\citep{szegedy2016rethinking} & 79.2933 & 0.1542 & 0.1542 & 0.0112 & 0.1541 \\
    % & tsl(brier\_score) & 3.8462 & 0.0104 (-5.23\%) & 0.0000 & 0.0114 (-4.53\%) & 0.0104 (-5.23\%) \\
    % & tsl(cross\_entropy) & 76.8510 & 0.0720 (26.32\%) & 0.0720 (26.33\%) & 0.0063 (17.33\%) & 0.0721 (26.23\%) \\
    & TSL(Dual Focal) & 79.9760 & 0.0163 (90.68\%) & 0.0170 (90.26\%) & 0.0034 (73.25\%) & 0.0163 (90.61\%) \\
    & TSL(Focal Loss) & 75.2885 & 0.0682 (68.81\%) & 0.0679 (68.95\%) & 0.0057 (63.75\%) & 0.0667 (68.87\%) \\
    & TSL(Focal Calibration Loss) & 79.7115 & 0.0357 (80.49\%) & 0.0357 (80.49\%) & 0.0041 (69.67\%) & 0.0356 (80.41\%) \\
    & TSL(Label Smooth) & 76.6106 & 0.0552 (64.17\%) & 0.0552 (64.17\%) & 0.0053 (52.43\%) & 0.0554 (64.07\%) \\
\midrule
\multirow{8}{*}{FMNIST} 
    % % & brier\_score & 72.4700 & 0.2513 & 0.2513 & 0.0590 & 0.2413 \\
    % & cross\_entropy & 85.4500 & 0.0710 & 0.0710 & 0.0153 & 0.0709 \\
    & Dual Focal~\citep{tao2023dual}  & 85.6700 & 0.0920 & 0.0920 & 0.0197 & 0.0921 \\
    & Focal Loss~\citep{lin2017focal}  & 84.9400 & 0.1612 & 0.1611 & 0.0319 & 0.1611 \\
    & Focal Calibration Loss~\citep{liang2024calibrating} & 85.7300 & 0.1154 & 0.1153 & 0.0226 & 0.1153 \\
    & Label Smooth~\citep{szegedy2016rethinking} & 85.2600 & 0.0651 & 0.0650 & 0.0155 & 0.0652 \\
    % & tsl(brier\_score) & 60.9100 & 0.1076 (57.19\%) & 0.1083 (56.91\%) & 0.0483 (18.09\%) & 0.1069 (55.69\%) \\
    % & tsl(cross\_entropy) & 85.1100 & 0.0246 (188.12\%) & 0.0242 (192.99\%) & 0.0075 (103.75\%) & 0.0247 (187.32\%) \\
    & TSL(Dual Focal) & 85.2900 & 0.0084 (90.83\%) & 0.0104 (88.68\%) & 0.0077 (61.16\%) & 0.0115 (87.52\%) \\
    & TSL(Focal Loss) & 81.8200 & 0.0943 (41.51\%) & 0.0939 (41.75\%) & 0.0201 (36.97\%) & 0.0940 (41.65\%) \\
    & TSL(Focal Calibration Loss) & 85.7300 & 0.0419 (63.65\%) & 0.0421 (63.51\%) & 0.0101 (55.26\%) & 0.0421 (63.50\%) \\
    & TSL(Label Smooth) & 84.7200 & 0.0636 (2.31\%) & 0.0637 (2.05\%) & 0.0141 (9.15\%) & 0.0639 (2.01\%) \\
\midrule
\multirow{8}{*}{KMNIST} 
    % & brier\_score & 46.2800 & 0.2142 & 0.2174 & 0.0603 & 0.2096 \\
    % & cross\_entropy & 78.6700 & 0.0333 & 0.0312 & 0.0128 & 0.0295 \\
    & Dual Focal~\citep{tao2023dual}  & 79.3300 & 0.1125 & 0.1125 & 0.0255 & 0.1125 \\
    & Focal Loss~\citep{lin2017focal}  & 77.7500 & 0.1573 & 0.1572 & 0.0340 & 0.1570 \\
    & Focal Calibration Loss~\citep{liang2024calibrating} & 78.8400 & 0.1222 & 0.1213 & 0.0280 & 0.1214 \\
    & Label Smooth~\citep{szegedy2016rethinking} & 75.3900 & 0.0911 & 0.0910 & 0.0226 & 0.0911  \\
    % & tsl(brier\_score) & 29.0700 & 0.0891 (58.40\%) & 0.0971 (55.32\%) & 0.0430 (28.72\%) & 0.0890 (57.54\%) \\
    % & tsl(cross\_entropy) & 76.3100 & 0.1075 (-222.41\%) & 0.1076 (-245.29\%) & 0.0249 (-95.29\%) & 0.1075 (-263.88\%) \\
    & TSL(Dual Focal) & 77.8700 & 0.0144 (87.24\%) & 0.0155 (86.22\%) & 0.0132 (48.21\%) & 0.0148 (86.84\%) \\
    & TSL(Focal Loss) & 73.1000 & 0.0506 (67.83\%) & 0.0488 (68.97\%) & 0.0190 (44.02\%) & 0.0487 (68.97\%) \\
    & TSL(Focal Calibration Loss) & 78.9600 & 0.0611 (50.03\%) & 0.0610 (49.69\%) & 0.0160 (43.06\%) & 0.0611 (49.66\%) \\
    & TSL(Label Smooth) & 77.9100 &  0.0886(2.84\%) & 0.0885 (2.80\%) & 0.0214 (5.58\%) & 0.0886(2.86\%) \\
\midrule

\bottomrule
\end{tabular}
\end{adjustbox}
\end{small}
\end{table*}

\section{Experimental Setup}
\label{sec:experimental_setup}

We evaluate our proposed Temperature-Scaled Loss (TSL) on seven standard vision benchmarks: \textbf{MNIST} \cite{deng2012mnist}, \textbf{EMNIST-Balance} \cite{cohen2017emnist}, \textbf{EMNIST-Letter} \cite{cohen2017emnist}, \textbf{FMNIST} \cite{xiao2017fashion}, \textbf{KMNIST} \cite{hastie2005elements}, \textbf{CIFAR-10} \cite{krizhevsky2009learning}, and \textbf{SVHN} \cite{Netzer2011ReadingDI}. We experiment with two network architectures:

\begin{itemize}
  \item \textbf{MLP:} Fully-connected networks with hidden-layer widths chosen from \(\{32, 64, 128, 256, 512, 1024\}\) and using either GELU or ReLU activations (without normalization).
  \item \textbf{KAN:} Configured with hidden-layer widths from \(\{2, 4, 8, 16\}\), B-spline grid order from \(\{3, 5, 10, 20\}\), spline degrees from \(\{2, 3, 5\}\) and grid range in \(\{[-1, 1], [-2, 2], [-4, 4]\}\).
\end{itemize}

All models are trained for \textbf{20 epochs} using the Adam optimizer with a batch size of \textbf{128}. The learning rate is initialized at \(10^{-3}\) and decayed to \(10^{-4}\) mid-training. Throughout training, we monitor both classification accuracy and calibration error on the test set, recording the best observed accuracy and the minimum calibration error (\eg, Expected Calibration Error (ECE) along with its variants such as AdaECE, Class-wise ECE, and Smooth ECE). For the MNIST and CIFAR-10 datasets, we further investigate KANs under various loss functions, including Cross-Entropy, Brier Score, Label Smoothing, Focal Loss, Dual Focal Loss, and Focal Calibration Loss. For each model–dataset configuration, the reported metrics are test Accuracy and the minimum Calibration Error achieved.

\section{Results}
\label{sec:results}

\subsection{Temperature Scaling on the Logit Space}
\label{subsec:temp_scaling_visual}

In Figure~\ref{temp_mlp_kan}, we vary the temperature parameter \(\tau\in \{8.0, 4.0, 2.0, 1.0,\\ 0.5\}\) and show how it affects the logits for MLP and KAN. The subplots reveal that higher \(\tau\) disperses the probability distribution, lowering overconfidence but increasing overall uncertainty. Conversely, when \(\tau\) is small, the predicted probabilities concentrate more sharply on the argmax class, enhancing confidence but risking overestimation. By examining the bars in each subplot, one can see that, although argmax classes remain consistent around moderate \(\tau\) values (\eg, \(\tau=1.0\)), the predicted class can fluctuate at extreme ends of the temperature range. This underscores a central trade-off: moderate temperatures can balance between maintaining reasonable confidence levels and controlling miscalibration.

\subsection{Impact of KAN Hyperparameters}
\label{subsec:kan_params_ece}

Figure~\ref{fig:multiple_factors_combined} in Appendix~\ref{app:Kan_hyper} reports calibration metrics under different KAN hyperparameter settings (Layer Width, Grid Range, Grid Order, Shortcut Function, and Number of Parameters). Each column in the figure corresponds to a specific hyperparameter, and each row shows a different calibration metric. These suggest that fine-tuning hyperparameter is essential for a well-calibrated KANs.
\begin{itemize}
    % \item \textbf{Layer Width:} Panels (a), (f), and (k) show that increasing the layer width yields broader calibration distributions—indicating a higher propensity for overconfident predictions—while the plots also exhibit narrower vertical spans, suggesting these calibration values remain more concentrated within that range.
    % \item \textbf{Grid Range:} Panels (b), (g), and (l) indicate that very large or negative grid boundaries amplify calibration errors, emphasizing the need for balanced range selection.
    % \item \textbf{Grid Order:} Panels (c), (h), and (m) show that higher-order splines allow more flexible transformations but can raise ECE variance. Selecting an appropriate spline order is thus crucial for stable calibration.
    % \item \textbf{Shortcut Functions:} Panels (d), (i), and (n) reveal that including identity or other bypass mappings generally narrows the ECE distribution, pointing to enhanced stability in the learned transformations.
    % \item \textbf{Model Size:} Panels (e), (j), and (o) demonstrate that larger models can achieve lower mean ECE, but may exhibit wider variance unless properly regularized.

    \item \textbf{Layer Width (a, f, k):} Widening the layers increases the spread of calibration errors, signifying heightened overconfidence; however, the lower regions of these violin plots suggest that tuning can still yield moderate miscalibration.
    \item \textbf{Grid Range (b, g, l):} Very large or negative ranges produce notably taller and wider violins, reflecting elevated mean and variance in calibration errors. In contrast, moderate ranges result in more compressed distributions.
    \item \textbf{Grid Order (c, h, m):} Heightened spline orders broaden the violin plots, revealing that while addition can capture complex patterns, it also raises the risk of miscalibration.
    \item \textbf{Shortcut Functions (d, i, n):} Adding shortcut connections imply that bypass routes help smooth the spline transformations and thus mitigate erratic calibration behavior.
    \item \textbf{Number of Parameters (e, j, o):} Larger model sizes sometimes yield lower average calibration errors, yet the broader widths indicate greater variance and underscore a potential for overconfidence if not adequately regularized.
\end{itemize}

\subsection{Comparison of Calibration Metrics}
\label{subsec:loss_comparison}

In Table~\ref{tab: loss-comparison-table} and Figure~\ref{fig:ece_loss}, we compare SOTA losses with their TSL-augmented counterparts. Notably, TSL consistently lowers ECE across all setups while retaining competitive test accuracy. For example, in Figure~\ref{fig:ece_loss}, the TSL variants exhibit visibly narrower confidence gaps, reflecting reduced overconfidence. By jointly optimizing the temperature parameter \(\tau\), models learn to align predicted confidence with empirical accuracy more effectively.

\subsection{Reliability Diagrams for KAN Models}
\label{subsec:kan_reliability_diagrams}

Figure~\ref{fig: 6losses_kan} shows reliability diagrams for a KAN model trained on MNIST, with all other hyperparameters held constant. Each subplot compares predicted accuracy (blue) to the ideal diagonal (pink). Standard methods, like Cross-Entropy and Brier Score, exhibit moderate calibration but display pronounced misalignment in the mid-confidence ranges. Label Smoothing and Focal Loss reduce overconfidence yet can still leave noticeable gaps. By contrast, TSL versions yield substantially tighter alignment across confidence bins, often cutting ECE by over 40\% relative to the corresponding standard methods. Although Dual Focal (panel (e)) and Focal Calibration (panel (f)) improve reliability, TSL-based variants typically show the smallest calibration errors, highlighting the value of learning \(\tau\).
%-------------------------------------------------------------------------
\section{Discussion}
\label{sec:discussion}
Kolmogorov-Arnold Networks (KANs), with their flexible spline-based transformations, exhibit greater calibration challenges compared to standard MLPs due to their dynamic logit distributions. Architectural factors such as grid range, spline order, and shortcut functions significantly influence calibration. By embedding a learnable temperature parameter \(\tau\) within the training objective using TSL, we directly address over- and underconfidence, eliminating the need for post-hoc calibration. This joint optimization enhances calibration performance, making TSL particularly effective for models with adaptive activations like KANs.
% Our experimental results reveal that Kolmogorov-Arnold Networks (KANs), which leverage learnable spline-based transformations, are inherently more prone to calibration issues than standard MLPs due to their flexible logit distributions. Key architectural hyperparameters—such as grid range, spline order, and shortcut function—play a significant role in determining calibration performance. By integrating a learnable temperature parameter \(\tau\) directly into the training objective via TSL, our approach effectively narrows the gap between predicted confidence and true accuracy. This joint optimization not only mitigates over- and underconfidence but also eliminates the need for separate post-hoc calibration procedures. Overall, TSL offers a robust and versatile method for improving the calibration of deep neural networks, particularly in architectures with adaptive activation functions such as KANs.

\section{Conclusion}
\label{sec:conclusion}
We explored calibration issues in Kolmogorov-Arnold Networks (KANs) and proposed the Temperature-Scaled Loss (TSL) to jointly optimize network parameters and the temperature \(\tau\). TSL consistently improves calibration performance across multiple benchmarks, demonstrating its effectiveness in reducing miscalibrated predictions. Future work will extend TSL to other deep neural architectures, refining spline formulations, and applying the method in domains such as medical imaging and risk-sensitive applications.

\newpage
\bibliographystyle{ACM-Reference-Format}
\bibliography{base}

\newpage
\appendix
\onecolumn

%%%%%%%%%%%%%%%%%%%%%%%%%%%%%%%%%%%%%%%%%%%%%%%%%%%%%%%%%%%%%%%%%%%%%%%%%%%%%%
\section{Kolmogorov-Arnold Network (KAN) Implementation}
\label{sec:kan_implementation}

Kolmogorov-Arnold Networks (KANs) approximate a multivariate function 
\begin{equation}
  f(\mathbf{x}) 
  \;=\; 
  \sum_{q=1}^{2n+1} 
  \Phi_q\!\Bigl(\,\sum_{p=1}^{n} \phi_{q,p}(x_p)\Bigr),
\end{equation}
where each \(\phi_{q,p}\) and \(\Phi_q\) is a univariate continuous function. In practice, we replace these functions with parameterized building blocks (\eg, spline-based layers), assuming the splines are continuously differentiable so that automatic differentiation applies.

\subsection{Input Transformation (Inner Layer with Splines)}
\label{subsec:inner_layer}

Each input coordinate \(x_p \in \mathbb{R}\) is mapped through a univariate function \(\psi_{q,p}\) (often denoted \(\phi_{q,p}\) in the theorem). In a spline-based KAN, we represent:
\begin{equation}
  \psi_{q,p}(x_p) 
  \;=\;
  \mathrm{Spline}_{\boldsymbol{\alpha}_{q,p}}\bigl(x_p\bigr),
\end{equation}
where \(\mathrm{Spline}_{\boldsymbol{\alpha}_{q,p}}(\cdot)\) is parameterized by a set of control points (knots) \(\boldsymbol{\alpha}_{q,p}\). In practice, regularization or smoothing priors may be applied to the spline coefficients to balance expressiveness against overfitting.

\subsection{Feature Aggregation}
\label{subsec:feature_aggregation}

After transforming each input coordinate, KAN sums these transformed features:
\begin{equation}
  z_q
  \;=\;
  \sum_{p=1}^{n} \psi_{q,p}(x_p).
\end{equation}
Here, the index \(q\) indicates each summation path corresponding to one outer function, with \(z_q\) representing an aggregated feature.

\subsection{Output Mapping (Outer Layer with Temperature Scaling)}
\label{subsec:outer_layer}

Each aggregated value \(z_q\) is passed through another univariate function \(\Phi_q\), implemented as a spline or a small neural layer:
\begin{equation}
  u_q
  \;=\;
  \Phi_q(z_q)
  \;=\;
  \mathrm{Spline}_{\boldsymbol{\beta}_{q}}(z_q),
\end{equation}
with \(\boldsymbol{\beta}_q\) as the learnable parameters. The final KAN output is then given by:
\begin{equation}
  f(\mathbf{x})
  \;=\;
  \sum_{q=1}^{2n+1} u_q.
\end{equation}
For classification, \(f(\mathbf{x})\) yields logits \(\mathbf{g} = [g_1, \dots, g_K]\) that are scaled by a learnable temperature parameter \(\tau > 0\):
\begin{equation}
  \tilde{g}_k 
  \;=\;
  \frac{g_k}{\tau}, 
  \quad k = 1, \dots, K.
\end{equation}
Thus, the temperature-scaled probabilities become:
\begin{equation}
  \hat{p}_k 
  \;=\;
  \frac{\exp(g_k / \tau)}{\sum_{j=1}^K \exp(g_j / \tau)}.
\end{equation}

\subsection{Complete Architecture with Splines}
\label{subsec:complete_arch_splines}

The overall KAN architecture can be summarized as:
\begin{equation}
  \mathbf{x}
  \;\xrightarrow{\text{(splines)}}\;
  \Bigl[\underbrace{\psi_{1,1}(x_1) + \cdots + \psi_{1,n}(x_n)}_{z_1}, \dots,
        \underbrace{\psi_{Q,1}(x_1) + \cdots + \psi_{Q,n}(x_n)}_{z_Q}\Bigr]
  \;\xrightarrow{\text{(splines)}}\;
  \Bigl[\Phi_1(z_1), \dots, \Phi_Q(z_Q)\Bigr]
  \;\xrightarrow{\text{(sum)}}\;
  f(\mathbf{x}),
\end{equation}
where \(Q=2n+1\) (adjustable in practice). Note that \(\tau\) is trained jointly with the spline parameters.

\subsection{Training Procedure (with Brier Score Base Loss)}
\label{subsec:training_procedure}

\paragraph{Brier Score Loss.}  
For a classification problem with \(K\) classes, given logits \(f(\mathbf{x})\) converted to probabilities \(\hat{p}_k(\mathbf{x})\) via softmax, the Brier score for a sample \((\mathbf{x}, y)\) is defined as:
\begin{equation}
  \ell_{\mathrm{Brier}}( \hat{\mathbf{p}}, y )
  \;=\;
  \sum_{k=1}^K \Bigl(\hat{p}_k(\mathbf{x}) - \mathbb{I}\{y=k\}\Bigr)^2.
\end{equation}

The training procedure involves:
\begin{enumerate}
  \item \textbf{Forward Pass:} Compute each \(\psi_{q,p}(x_p)\) to obtain \(z_q\), then compute \(\Phi_q(z_q)\) and sum to obtain logits \(g_k\). Scale the logits: \(\tilde{g}_k = g_k / \tau\), and convert to probabilities \(\hat{p}_k\).
  \item \textbf{Loss Computation:} Calculate \(\ell_{\mathrm{Brier}}(\hat{\mathbf{p}}, y)\) for each sample (or average over a mini-batch).
  \item \textbf{Backward Pass:} Backpropagate gradients through both the inner (spline) layers and outer layers to update the spline parameters \(\{\boldsymbol{\alpha}_{q,p}, \boldsymbol{\beta}_q\}\) as well as \(\tau\).
  \item \textbf{Parameter Updates:} Use an optimizer (\eg, Adam) to update all parameters. A projection such as \(\tau \gets \max(\varepsilon, \tau)\) (with a small \(\varepsilon>0\)) ensures \(\tau\) remains positive.
\end{enumerate}

This procedure can be adapted to other losses (\eg, cross-entropy, focal) or to the Temperature-Scaled Loss (TSL) framework described next.

\subsection{Temperature-Scaled Loss (TSL) Integration}
\label{subsec:tsl_integration}

To improve calibration, we introduce a learnable temperature \(\tau > 0\) that rescales the logits:
\begin{equation}
  \tilde{f}(\mathbf{x})
  \;=\;
  \frac{f(\mathbf{x})}{\tau}.
\end{equation}
We then define the \emph{Temperature-Scaled Loss (TSL)} by replacing \(f(\mathbf{x})\) with \(\tilde{f}(\mathbf{x})\) in the base loss. For example, using the Brier score:
\begin{equation}
  \mathcal{L}_{\mathrm{TSL}}
  \;=\;
  \frac{1}{N}\sum_{i=1}^N
  \ell_{\mathrm{Brier}}\!\Bigl(\mathrm{softmax}\Bigl(\frac{f(\mathbf{x}_i)}{\tau}\Bigr),\,y_i\Bigr).
\end{equation}
Both the spline parameters and \(\tau\) are updated jointly via gradient descent.

\paragraph{Implementation Notes.}  
Each spline layer is implemented as a collection of 1D transformations, and automatic differentiation handles the piecewise definitions given the assumed continuity. Hyperparameters such as spline degree, number of knots, and layer widths are tuned to balance expressiveness and overfitting.

%%%%%%%%%%%%%%%%%%%%%%%%%%%%%%%%%%%%%%%%%%%%%%%%%%%%%%%%%%%%%%%%%%%%%%%%%%%%%%
\section{Proof: Temperature-Scaled Loss as a Constrained Optimization Problem}
\label{app:proof_contrained_opt}

\subsection{Problem Formulation}
Given \(N\) samples with logits \(\mathbf{g}_i \in \mathbb{R}^K\) and labels \(y_i \in \{1, \dots, K\}\), we wish to maximize the entropy of the output distribution \(q_{ik}\) (interpreted as calibrated probabilities) subject to:
\begin{equation}
\max_{q} \;\; - \sum_{i=1}^N \sum_{k=1}^K q_{ik} \log q_{ik},
\end{equation}
subject to:
\begin{enumerate}
    \item \( q_{ik} \geq 0, \; \forall i, k,\)
    \item \( \sum_{k=1}^K q_{ik} = 1, \; \forall i,\)
    \item \( \sum_{i=1}^N g_i^{(y_i)} \;=\; \sum_{i=1}^N \sum_{k=1}^K g_{ik}\, q_{ik}\) (preservation of expected logits).
\end{enumerate}

\subsection{Lagrangian Formulation and Solution}
The Lagrangian is
\begin{equation}
\begin{aligned}
L &= - \sum_{i=1}^N \sum_{k=1}^K q_{ik} \log q_{ik} 
  + \lambda \Bigl( \sum_{i=1}^N \sum_{k=1}^K g_{ik}\,q_{ik} - \sum_{i=1}^N g_i^{(y_i)} \Bigr) \\
  &\quad + \sum_{i=1}^N \beta_i \Bigl( \sum_{k=1}^K q_{ik} - 1 \Bigr),
\end{aligned}
\end{equation}
where \(\lambda\) enforces the expected logits constraint and \(\beta_i\) enforces normalization. Setting \(\frac{\partial L}{\partial q_{ik}} = 0\) gives
\begin{equation}
\log q_{ik} = \lambda\,g_{ik} + \beta_i - 1,
\end{equation}
or equivalently,
\begin{equation}
q_{ik} = e^{\lambda g_{ik} + \beta_i - 1}.
\end{equation}
Normalization requires
\begin{equation}
\sum_{k=1}^K e^{\lambda g_{ik} + \beta_i - 1} = 1, \quad \text{so} \quad e^{\beta_i - 1} = \frac{1}{\sum_{k=1}^K e^{\lambda g_{ik}}}.
\end{equation}
Thus,
\begin{equation}
q_{ik} = \frac{e^{\lambda g_{ik}}}{\sum_{j=1}^K e^{\lambda g_{ij}}}.
\end{equation}
Comparing with temperature scaling
\begin{equation}
q_{ik} = \frac{\exp(g_{ik}/\tau)}{\sum_{j=1}^K \exp(g_{ij}/\tau)},
\end{equation}
we deduce that \(\lambda = \frac{1}{\tau}\).

\paragraph{Conclusion.}  
This derivation shows that TSL can be interpreted as solving an entropy maximization problem under constraints that preserve the expected logits. The temperature \(\tau\) thus dynamically adjusts the output distribution.

%%%%%%%%%%%%%%%%%%%%%%%%%%%%%%%%%%%%%%%%%%%%%%%%%%%%%%%%%%%%%%%%%%%%%%%%%%%%%%
\section{Proof of Proposition~\ref{prop:tsl_strict_properness}}
\label{appendix:proof_strict_properness}

\begin{proposition}[Strict Properness of TSL]
Let \(\mathcal{L}_{\mathrm{base}}(p, y)\) be a strictly proper scoring rule, and define
\begin{equation}
  \mathcal{L}_{\mathrm{TSL}}(\theta,\tau)
  \;=\;
  \mathcal{L}_{\mathrm{base}}
  \Bigl(\mathrm{softmax}\Bigl(\frac{g(\theta)}{\tau}\Bigr),\,y\Bigr),
  \quad \tau > 0.
\end{equation}
Then \(\mathcal{L}_{\mathrm{TSL}}\) is strictly proper with respect to the true conditional distribution \(\Pr(Y\mid X)\).
\end{proposition}

\begin{proof}[Proof]
Since \(\mathcal{L}_{\mathrm{base}}(p, y)\) is strictly proper, its unique minimizer occurs when the predicted distribution \(p\) equals the true conditional distribution. Because scaling logits by \(1/\tau\) is a monotonic, invertible reparameterization (for \(\tau > 0\)), the unique minimizer is preserved. Thus, no incorrect distribution \(\tilde{p}\neq q\) can yield a lower loss, ensuring that \(\mathcal{L}_{\mathrm{TSL}}\) remains strictly proper.
\end{proof}

%%%%%%%%%%%%%%%%%%%%%%%%%%%%%%%%%%%%%%%%%%%%%%%%%%%%%%%%%%%%%%%%%%%%%%%%%%%%%%
\section{Proof of Lemma~\ref{lemma:tau_updates_correct_miscalibration}}
\label{appendix:gradient_adjustments}

\begin{lemma}[Monotonic Gradient Updates]
Consider the softmax probabilities
\begin{equation}
  \hat{p}_{ik}(\theta,\tau)
  =
  \frac{\exp\bigl(g_{ik}(\theta)/\tau\bigr)}{\sum_{j=1}^K \exp\bigl(g_{ij}(\theta)/\tau\bigr)},
\end{equation}
and let 
\(\mathcal{L}_{\mathrm{TSL}}(\theta,\tau) = \mathcal{L}_{\mathrm{base}}(\hat{p}(\theta,\tau),y)\).  
Then the derivative \(\frac{\partial \mathcal{L}_{\mathrm{TSL}}}{\partial \tau}\) adjusts \(\tau\) so as to counteract over- or underconfidence.
\end{lemma}

\begin{proof}[Proof]
By the chain rule, we have
\begin{equation}
  \frac{\partial \mathcal{L}_{\mathrm{TSL}}}{\partial \tau}
  =
  \sum_{i=1}^N \sum_{k=1}^K \frac{\partial \mathcal{L}_{\mathrm{base}}}{\partial \hat{p}_{ik}}\,\frac{\partial \hat{p}_{ik}}{\partial \tau}.
\end{equation}
A more detailed derivation shows that when \(\hat{p}_{ik}\) is too high relative to the true label (overconfidence), the gradient \(\frac{\partial \mathcal{L}_{\mathrm{base}}}{\partial \hat{p}_{ik}}\) is positive, and the corresponding \(\frac{\partial \hat{p}_{ik}}{\partial \tau}\) leads to an increase in \(\tau\) (flattening the distribution). Conversely, underconfidence produces a negative gradient, prompting a decrease in \(\tau\) (sharpening the distribution). Thus, the update on \(\tau\) acts to reduce miscalibration.
\end{proof}

%%%%%%%%%%%%%%%%%%%%%%%%%%%%%%%%%%%%%%%%%%%%%%%%%%%%%%%%%%%%%%%%%%%%%%%%%%%%%%
\section{Proof of Theorem~\ref{thm:tsl_local_minimum}}
\label{appendix:convergence_calibration}

\begin{theorem}[Local Convergence \& Improved Calibration]
Assume that \(\mathcal{L}_{\mathrm{TSL}}(\theta,\tau)\) is continuous, differentiable, and bounded below, and that its gradients are Lipschitz continuous. Let \(\{(\theta_t,\tau_t)\}\) be the iterates produced by (stochastic) gradient descent with suitable diminishing step sizes. Then, \((\theta_t,\tau_t)\) converges to a local minimum of \(\mathcal{L}_{\mathrm{TSL}}\). Moreover, the resulting model is at least as well-calibrated as one trained with a fixed temperature \(\tau=1\).
\end{theorem}

\begin{proof}[Proof Sketch]
Under standard assumptions (\eg, Lipschitz continuity of the gradients and proper learning rate schedules), gradient descent converges to a local minimum in nonconvex settings \citep{ghadimi2013stochastic}. By Lemma~\ref{lemma:tau_updates_correct_miscalibration}, the gradient update for \(\tau\) systematically reduces calibration mismatches. Consequently, the final prediction \(\mathrm{softmax}(g(\theta^*)/\tau^*)\) exhibits lower calibration error than \(\mathrm{softmax}(g(\theta_{\mathrm{base}}))\). 
\end{proof}

%%%%%%%%%%%%%%%%%%%%%%%%%%%%%%%%%%%%%%%%%%%%%%%%%%%%%%%%%%%%%%%%%%%%%%%%%%%%%%
\section{Calibration Metrics}
\label{appendix: cal_metr}

We summarize common metrics to quantify calibration:
\begin{enumerate}
    \item \textbf{Expected Calibration Error (ECE):}
    \begin{equation}
    \mathrm{ECE} = \sum_{m=1}^M \frac{|B_m|}{n} \Bigl| \mathrm{acc}(B_m) - \mathrm{conf}(B_m) \Bigr|,
    \end{equation}
    where \(B_m\) denotes the set of predictions in bin \(m\), \(\mathrm{acc}(B_m)\) is the bin accuracy, and \(\mathrm{conf}(B_m)\) is the mean confidence.
    
    \item \textbf{Adaptive ECE (AdaECE):}
    \begin{equation}
    \mathrm{AdaECE} = \frac{1}{M} \sum_{m=1}^M \Bigl| \mathrm{acc}(B_m) - \mathrm{conf}(B_m) \Bigr|,
    \end{equation}
    using bins that adaptively contain \(\frac{n}{M}\) points each.
    
    \item \textbf{Maximum Calibration Error (MCE):}
    \begin{equation}
    \mathrm{MCE} = \max_{m=1}^M \Bigl| \mathrm{acc}(B_m) - \mathrm{conf}(B_m) \Bigr|.
    \end{equation}
    
    \item \textbf{Brier Score:}
    \begin{equation}
    \mathrm{Brier\,Score} = \frac{1}{n} \sum_{i=1}^n \sum_{k=1}^K \Bigl(y_{ik} - \hat{p}_{ik}\Bigr)^2,
    \end{equation}
    where \(y_{ik}\) is the one-hot encoding of the true label and \(\hat{p}_{ik}\) is the predicted probability.
    
    \item \textbf{Negative Log-Likelihood (NLL):}
    \begin{equation}
    \mathrm{NLL} = - \frac{1}{n} \sum_{i=1}^n \log \hat{p}_{y_i}.
    \end{equation}
\end{enumerate}

%%%%%%%%%%%%%%%%%%%%%%%%%%%%%%%%%%%%%%%%%%%%%%%%%%%%%%%%%%%%%%%%%%%%%%%%%%%%%%
\section{Theorem: Calibration Consistency}
\label{app:calibration_consistency}

\begin{theorem}
For any base loss \(\mathcal{L}_{\mathrm{base}}\) (\eg, Cross-Entropy or Brier Score), the inclusion of a learnable temperature \(\tau\) in TSL guarantees a reduction in the calibration gap \(\bigl| \mathrm{conf}(B_m) - \mathrm{acc}(B_m) \bigr|\), provided that \(\tau\) is dynamically updated during training.
\end{theorem}

\begin{proof}
Let the logits be \(\mathbf{g}_i\) and the scaled logits \(\tilde{\mathbf{g}}_i = \mathbf{g}_i / \tau\). The predicted probabilities are:
\begin{equation}
\hat{p}_{ik} 
= 
\frac{\exp(g_{ik}/\tau)}{\sum_{j=1}^K \exp(g_{ij}/\tau)}.
\end{equation}
For a bin \(B_m\), define the calibration gap as:
\begin{equation}
\Delta_m 
= 
\Bigl| \mathrm{conf}(B_m) - \mathrm{acc}(B_m) \Bigr|,
\end{equation}
where
\begin{equation}
\mathrm{conf}(B_m) 
= 
\frac{1}{|B_m|} \sum_{i \in B_m} \max_k \hat{p}_{ik}, \quad
\mathrm{acc}(B_m) 
= 
\frac{1}{|B_m|} \sum_{i \in B_m} \mathbb{I}\{\arg\max_k \hat{p}_{ik} = y_i\}.
\end{equation}
By appropriately adjusting \(\tau\) (\ie, selecting \(\tau_i = \arg\min_\tau |\mathrm{acc}(B_i) - \mathrm{conf}(B_i; \tau)|\)), we have
\begin{equation}
\Bigl|\mathrm{acc}(B_i) - \mathrm{conf}(B_i; \tau_i)\Bigr|
\;\le\;
\Bigl|\mathrm{acc}(B_i) - \mathrm{conf}(B_i)\Bigr|.
\end{equation}
Weighting over all bins yields
\begin{equation}
\mathrm{ECE}_{\mathrm{TSL}}
=
\sum_{i=1}^M \frac{|B_i|}{n} \Bigl|\mathrm{acc}(B_i) - \mathrm{conf}(B_i; \tau_i)\Bigr|
\;\le\;
\mathrm{ECE}_{\mathrm{base}}.
\end{equation}
Thus, TSL reduces the overall calibration error.
\end{proof}

%%%%%%%%%%%%%%%%%%%%%%%%%%%%%%%%%%%%%%%%%%%%%%%%%%%%%%%%%%%%%%%%%%%%%%%%%%%%%%
\section{\texorpdfstring{Lemma: Monotonicity of \(\tau\) Adjustment}{Lemma: Monotonicity of tau Adjustment}}

\label{app:monotonic_tau}

\begin{lemma}
The gradient-based update of \(\tau\) in TSL ensures monotonic adjustments:
\begin{itemize}
    \item \(\tau\) increases when the model is overconfident (\(\mathrm{conf}(B_m) > \mathrm{acc}(B_m)\)).
    \item \(\tau\) decreases when the model is underconfident (\(\mathrm{conf}(B_m) < \mathrm{acc}(B_m)\)).
\end{itemize}
\end{lemma}

\begin{proof}
Recall the softmax probability:
\begin{equation}
\hat{p}_{ik} 
= 
\frac{\exp(g_{ik}/\tau)}{\sum_{j=1}^K \exp(g_{ij}/\tau)}.
\end{equation}
Differentiating with respect to \(\tau\) yields:
\begin{equation}
\frac{\partial \hat{p}_{ik}}{\partial \tau} 
= 
\hat{p}_{ik} \left(\frac{g_{ik}}{\tau^2} - \frac{ \sum_{j=1}^K \hat{p}_{ij}\,g_{ij}}{\tau^2}\right).
\end{equation}
Thus, if the model is overconfident, the gradient of the TSL loss with respect to \(\tau\) is such that \(\tau\) is updated upward (flattening the output probabilities). Conversely, underconfidence leads to a negative gradient, reducing \(\tau\) and sharpening the distribution. Therefore, the updates are monotonic with respect to the calibration gap.
\end{proof}

%%%%%%%%%%%%%%%%%%%%%%%%%%%%%%%%%%%%%%%%%%%%%%%%%%%%%%%%%%%%%%%%%%%%%%%%%%%%%%
\section{Corollary: Reduction of ECE}
\label{appendix:ece_reduction}

\begin{corollary}
Let \(\mathrm{acc}(B_i)\) denote the accuracy and \(\mathrm{conf}(B_i)\) the mean confidence in bin \(B_i\) for an unscaled model, and let \(\mathrm{conf}(B_i; \tau_i)\) denote the corresponding value after temperature scaling. Then, for each bin \(B_i\),
\begin{equation}
\Bigl|\mathrm{acc}(B_i) - \mathrm{conf}(B_i; \tau_i)\Bigr|
\;\le\;
\Bigl|\mathrm{acc}(B_i) - \mathrm{conf}(B_i)\Bigr|.
\end{equation}
Summing over all bins gives
\begin{equation}
\mathrm{ECE}_{\mathrm{TSL}}
\;\le\;
\mathrm{ECE}_{\mathrm{base}},
\end{equation}
where \(\mathrm{ECE}_{\mathrm{TSL}}\) and \(\mathrm{ECE}_{\mathrm{base}}\) denote the calibration errors under TSL and the base loss, respectively.
\end{corollary}

\subsection{Notation and Proof}
\begin{enumerate}
    \item \textbf{Bin Accuracy:}  
   \(
   \mathrm{acc}(B_i) 
   = 
   \frac{1}{|B_i|} \sum_{t \in B_i} \mathbb{1}(y_t = \hat{y}_t).
   \)
    \item \textbf{Unscaled Confidence:}  
   \(
   \mathrm{conf}(B_i)
   = 
   \frac{1}{|B_i|} \sum_{t \in B_i} \max_{k} \hat{p}_{tk},
   \)
   where \(\hat{p}_{tk} = \frac{\exp(g_{tk})}{\sum_{j=1}^K \exp(g_{tj})}\).
    \item \textbf{Scaled Confidence:}  
   \(
   \mathrm{conf}(B_i;\tau_i) 
   = 
   \frac{1}{|B_i|} \sum_{t \in B_i} \max_{k} \frac{\exp(g_{tk}/\tau_i)}{\sum_{j=1}^K \exp(g_{tj}/\tau_i)}.
   \)
    \item \textbf{Choosing \(\tau_i\):}  
   \(
   \tau_i = \arg\min_\tau \Bigl|\mathrm{acc}(B_i) - \mathrm{conf}(B_i;\tau)\Bigr|.
   \)
\end{enumerate}

Since \(\mathrm{conf}(B_i;\tau)\) is monotonic in \(\tau\) (decreasing from 1 as \(\tau \to 0^+\) to \(1/K\) as \(\tau \to \infty\)), there exists a \(\tau_i\) such that \(\mathrm{acc}(B_i) = \mathrm{conf}(B_i;\tau_i)\). Comparing with the unscaled case (\(\tau = 1\)) shows that the calibration gap is reduced, and thus
\begin{equation}
\mathrm{ECE}_{\mathrm{TSL}} \le \mathrm{ECE}_{\mathrm{base}}.
\end{equation}

%%%%%%%%%%%%%%%%%%%%%%%%%%%%%%%%%%%%%%%%%%%%%%%%%%%%%%%%%%%%%%%%%%%%%%%%%%%%%%
\section{Extended Theoretical Evidence}
\label{appendix:extended_theory}

This section augments our theoretical treatment of TSL with additional insights.

\subsection{Riemann–Stieltjes Approximation of ECE}
\label{app:stieltjes_ece}
The Expected Calibration Error (ECE) can be seen as a Riemann–Stieltjes sum approximating the integral:
\begin{equation}
\mathbb{E}\Bigl[\Pr(\hat{Y}=Y \mid \hat{P}=p) - p\Bigr]
=
\int_{0}^{1}
\Bigl(\Pr(\hat{Y}=Y \mid \hat{P}=p) - p\Bigr)\, dF_{\hat{P}}(p),
\end{equation}
where \(F_{\hat{P}}(p)\) is the distribution function of the predicted confidence. In practice, this is approximated by partitioning \([0,1]\) into \(M\) bins. Reducing \(\mathrm{ECE}\) via TSL therefore corresponds to a true reduction in miscalibration.

\subsection{Max-Entropy Perspective on TSL}
\label{app:max_entropy_tsl}
\begin{theorem}[Max-Entropy Perspective on TSL]
\label{thm:max_entropy_tsl_appendix}
Let \(\{\mathbf{g}_i\}_{i=1}^N\) be logits for \(N\) samples with labels \(y_i \in \{1,\dots,K\}\), and introduce a learnable \(\tau>0\) so that the rescaled logits are \(\mathbf{g}_i/\tau\). Then, TSL is equivalent to solving the constrained optimization problem:
\begin{equation}
\max_{q}
\quad -\sum_{i=1}^N \sum_{k=1}^K q(\mathbf{g}_i)(k)\,\ln\,q(\mathbf{g}_i)(k)
\end{equation}
subject to:
\[
\begin{cases}
q(\mathbf{g}_i)(k)\ge0,\quad \sum_{k=1}^K q(\mathbf{g}_i)(k)=1, \\
\sum_{i=1}^N g_i^{(y_i)} = \sum_{i=1}^N\sum_{k=1}^K g_{ik}\,q(\mathbf{g}_i)(k).
\end{cases}
\]
Using Lagrange multipliers, one finds \(\lambda = 1/\tau\) so that
\begin{equation}
q(\mathbf{g}_i)(k)
=
\frac{\exp\!\bigl(g_{ik}/\tau\bigr)}
     {\sum_{j=1}^K \exp\!\bigl(g_{ij}/\tau\bigr)}.
\end{equation}
\end{theorem}

\begin{proof}[Sketch]
The solution follows the standard method of Lagrange multipliers, as in Section~\ref{app:proof_contrained_opt}. Mapping \(\lambda\) to \(1/\tau\) recovers the softmax formulation.
\end{proof}

\subsection{Bounding the Temperature}
\label{app:tau_bounding}
To prevent numerical instability, we bound \(\tau\) within \([\tau_{\min},\tau_{\max}]\) (\eg, 0 to 10) using:
\begin{equation}
\tau
\leftarrow
\Pi_{[\tau_{\min}, \tau_{\max}]}\Bigl(
\tau - \eta_{\tau}\,\nabla_{\tau}\,\mathcal{L}_{\mathrm{TSL}}(\theta,\tau)
\Bigr),
\end{equation}
where \(\Pi\) is a projection operator. This ensures that \(\tau\) does not approach 0 or \(\infty\) during training.

\subsection{Multiclass Extension Notes}
\label{app:multiclass_extension}
In multiclass settings, scaling logits by \(1/\tau\) does not change the \(\arg\max\) decision, thereby preserving accuracy. Our local convergence arguments extend naturally to this case, and one can also consider vector or matrix scaling if needed.

%%%%%%%%%%%%%%%%%%%%%%%%%%%%%%%%%%%%%%%%%%%%%%%%%%%%%%%%%%%%%%%%%%%%%%%%%%%%%%
\section{Expected Calibration Error and Temperature Scaling}
\label{appendix:ece_tau}

\begin{figure}[ht]
\vskip 0.2in
\begin{center}
\includegraphics[width=0.6\columnwidth]{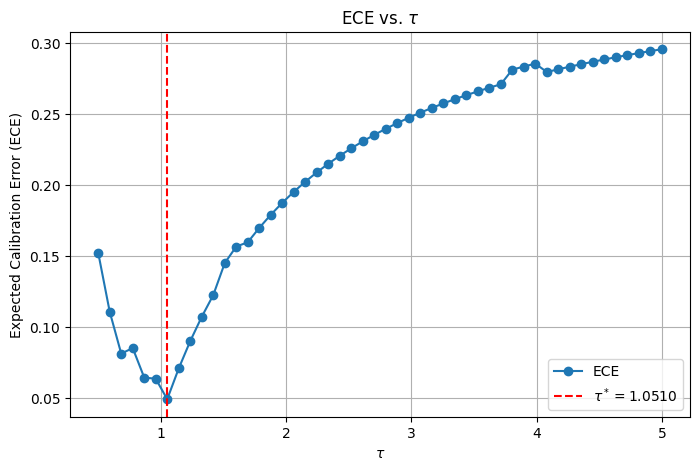}
\caption{Expected Calibration Error (ECE) as a function of the temperature parameter \(\tau\) for a trained neural network. The red dashed line represents the optimal temperature \(\tau^* = 1.0510\). The U-shaped curve illustrates the trade-off between overconfidence (low \(\tau\)) and underconfidence (high \(\tau\)).}
\Description{tbc}
\label{fig:ece_tau_1.05}
\end{center}
\vskip -0.2in
\end{figure}

To empirically study the impact of temperature scaling on calibration, we generate a synthetic classification dataset (500 samples, 20 features, 3 classes with imbalance) using Scikit-learn's \texttt{make\_classification}. A KAN model is trained in PyTorch using cross-entropy loss and the Adam optimizer. By varying the temperature \(\tau\) in the range \([0.5, 5.0]\) on a held-out validation set, we compute the ECE for each \(\tau\) using standard binning methods. Figure~\ref{fig:ece_tau_1.05} shows a U-shaped curve, with the optimal temperature \(\tau^* = 1.0510\) (marked by a red dashed line) minimizing ECE.

%%%%%%%%%%%%%%%%%%%%%%%%%%%%%%%%%%%%%%%%%%%%%%%%%%%%%%%%%%%%%%%%%%%%%%%%%%%%%%
\section{\texorpdfstring{Toy Example: Gradient Sign on \(\tau\)}{Toy Example: Gradient Sign on tau}}

\label{appendix:toy_example_tau}

\begin{figure}[ht]
\vskip 0.2in
\begin{center}
\includegraphics[width=0.6\columnwidth]{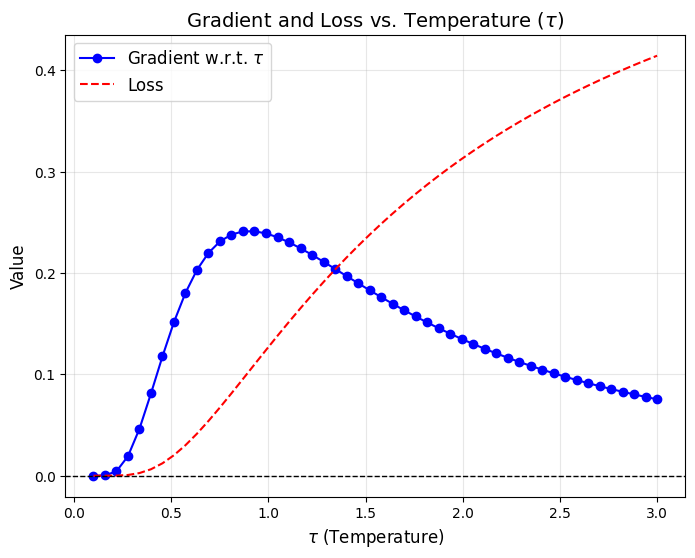}
\caption{Gradient of Temperature-Scaled Loss (TSL) with respect to the temperature parameter \(\tau\) for an overconfident example. The positive gradient at lower \(\tau\) indicates that increasing \(\tau\) reduces overconfidence.}
\Description{tbc}
\label{fig:tsl_gridient_tau}
\end{center}
\vskip -0.2in
\end{figure}

Figure~\ref{fig:tsl_gridient_tau} visualizes the gradient of the Temperature-Scaled Loss (TSL) with respect to \(\tau\) for an overconfident example. Initially, the gradient is positive, indicating that increasing \(\tau\) will flatten the probability distribution and reduce overconfidence. As \(\tau\) increases, the gradient eventually becomes negative, indicating that further increases would lead to underconfidence. This toy example demonstrates the adaptive behavior of \(\tau\) during training.

%%%%%%%%%%%%%%%%%%%%%%%%%%%%%%%%%%%%%%%%%%%%%%%%%%%%%%%%%%%%%%%%%%%%%%%%%%%%%%
\section{Proof of Proposition~\ref{prop:spline_calib} (Spline Order and Calibration Error)}
\label{app:proof_order_ece}

\textbf{Proposition:}  
Let \(\phi_{l,i,j}\) be a B-spline of order \(s\) (degree \(d=s-1\)) with \(G\) knots. For fixed \(G\), the variance of the logit \(\theta(\mathbf{x})\) grows with \(s\), thereby increasing the Expected Calibration Error (ECE).

\textbf{Proof Outline:}
\begin{enumerate}
    \item \textbf{B-Spline Parameterization:}  
    A B-spline is expressed as
    \begin{equation}
    \phi(x) = \sum_{k=1}^{G-s} c_k\,B_k(x; s),
    \end{equation}
    where \(B_k(x; s)\) are basis functions and \(c_k\) are learnable coefficients.
    
    \item \textbf{Logit Variance:}  
    In a KAN layer, the logit is
    \begin{equation}
    \theta(\mathbf{x}) = \sum_{j=1}^{n_l} \phi_{l,i,j}(x_j) = \sum_{j=1}^{n_l} \sum_{k=1}^{G-s} c_{j,k}\,B_k(x_j; s).
    \end{equation}
    Assuming the \(c_{j,k}\) are independent, zero-mean with variance \(\sigma_c^2\), for a single input dimension we have
    \begin{equation}
    \mathbb{V}[\theta(x)] \propto \sigma_c^2 \sum_{k=1}^{G-s} B_k(x; s)^2.
    \end{equation}
    
    \item \textbf{Basis Function Overlap:}  
    Higher \(s\) implies broader support of each basis function, increasing overlap. Although the partition of unity implies
    \begin{equation}
    \sum_{k} B_k(x; s) = 1,
    \end{equation}
    the sum of squares \(\sum_{k} B_k(x; s)^2\) empirically behaves like \(\mathcal{O}(1/s)\).
    
    \item \textbf{Coefficient Scaling:}  
    To maintain approximation capacity, higher-order splines require larger coefficients, such that \(\sigma_c^2 \propto s^2\) (see, \eg, \citealp{de1972calculating}). Thus,
    \begin{equation}
    \mathbb{V}[\theta(x)] \propto s^2 \cdot \frac{1}{s} = s.
    \end{equation}
    
    \item \textbf{Link to ECE:}  
    Following \citet{guo2017calibration}, the miscalibration tends to increase with the variance of the logits, leading to
    \begin{equation}
    \mathrm{ECE} \propto \sqrt{\mathbb{V}[\theta(x)]} \propto \sqrt{s}.
    \end{equation}
\end{enumerate}
Thus, higher spline order \(s\) increases ECE.

%%%%%%%%%%%%%%%%%%%%%%%%%%%%%%%%%%%%%%%%%%%%%%%%%%%%%%%%%%%%%%%%%%%%%%%%%%%%%%
\section{Formal Proof Sketch: TSL Reduces Smooth Calibration Error}
\label{app:tsl_smce_proof}

We provide a proof sketch showing that minimizing the Temperature-Scaled Loss (TSL) reduces the smooth calibration error (\(\mathrm{smCE}\)), as defined in \cite{blasiok2023smooth}.

\subsection{TSL Setup}
\begin{itemize}
    \item \textbf{Model Logits:} \(f(\mathbf{x}; \theta)\) produces logits \(\mathbf{g}(\mathbf{x}; \theta)\in\mathbb{R}^K\).
    \item \textbf{Temperature Scaling:} Rescaled logits are given by \(\tilde{\mathbf{g}}(\theta,\tau) = \mathbf{g}(\theta)/\tau\).
    \item \textbf{Predicted Probabilities:} \(\hat{p}_{i,k}(\theta,\tau) = \mathrm{softmax}\bigl(\tilde{\mathbf{g}}_i(\theta,\tau)\bigr)_k\).
    \item \textbf{TSL Objective:}
    \begin{equation}
    \mathcal{L}_{\mathrm{TSL}}(\theta,\tau) = \sum_{i=1}^N \mathcal{L}_{\mathrm{base}}\!\Bigl(\tilde{\mathbf{g}}_i(\theta,\tau), y_i\Bigr).
    \end{equation}
\end{itemize}

\subsection{Gradient Penalizes Calibration Mismatch}
Denote \(\Delta_{i,k}(\theta,\tau)=y_{i,k}-\hat{p}_{i,k}(\theta,\tau)\). For a strictly proper loss (\eg, cross-entropy), the derivative with respect to the rescaled logits is proportional to \(\Delta_{i,k}\). The chain rule implies that the gradient with respect to \(\tau\) contains terms that are proportional to \(\Delta_{i,k}\). In effect:
\begin{itemize}
    \item When \(\Delta_{i,k} < 0\) (overconfidence), the gradient pushes \(\tau\) upward (flattening probabilities).
    \item When \(\Delta_{i,k} > 0\) (underconfidence), the gradient pushes \(\tau\) downward (sharpening probabilities).
\end{itemize}
Thus, the TSL update systematically reduces \(|\Delta_{i,k}|\).

\subsection{Reduction in smCE}
Recall that the smooth calibration error (smCE) is defined as
\begin{equation}
\mathrm{smCE}(f)
\;=\;
\sup_{\phi\in \mathcal{H}}
\;\mathbb{E}\Bigl[\sum_{k} \Delta_{i,k}\,\phi\bigl(\hat{p}_{i,k}\bigr)\Bigr],
\end{equation}
where \(\mathcal{H}\) is the set of 1-Lipschitz functions. As TSL training reduces \(|\Delta_{i,k}|\) for all \(i,k\), the supremum over \(\phi\) becomes smaller. Thus, \(\mathrm{smCE}\) decreases as TSL minimizes \(\mathcal{L}_{\mathrm{TSL}}\).

\paragraph{Local Minima Argument.}  
At a local minimizer \((\theta^*,\tau^*)\), the gradients vanish, and the mismatches \(\Delta_{i,k}\) are small. Hence, no 1-Lipschitz function \(\phi\) can accumulate a large error, resulting in a lower smCE compared to a poorly calibrated model.

\section{More Tables}

\begin{table}[ht]
\centering
\caption{Comparison of test accuracy and calibration metrics across different loss functions on the Machine Learning datasets \protect\footnotemark[2].}
\label{tab:combined_results_ml}
\begin{adjustbox}{width=\textwidth}
\begin{tabular}{llccccc}
\toprule
\textbf{Dataset} & \textbf{Loss Function} & \textbf{Best Test Acc} & \textbf{Best ECE} & \textbf{Best AdaECE} & \textbf{Best CECE} & \textbf{Best SMECE} \\
\midrule
\multirow{10}{*}{SVHN} 
    % & brier\_score & 19.5874 & 0.0147 & 0.0307 & 0.0064 & 0.0147 \\
    % & cross\_entropy & 66.8869 & 0.0540 & 0.0540 & 0.0126 & 0.0540 \\
    & Dual Focal~\citep{tao2023dual}  & 68.2122 & 0.1039 & 0.1039 & 0.0201 & 0.1039 \\
    & Focal Loss~\citep{lin2017focal}  & 65.6077 & 0.1492 & 0.1492 & 0.0281 & 0.1487 \\
    & Focal Calibration Loss~\citep{liang2024calibrating} & 67.4516 & 0.1311 & 0.1311 & 0.0242 & 0.1307 \\
    & Label Smooth~\citep{szegedy2016rethinking} & 65.1352 & 0.0973 & 0.0973 & 0.0236 & 0.0973 \\
    % & tsl(brier\_score) & 19.5874 & 0.0622 (-324.26\%) & 0.0648 (-111.09\%) & 0.0141 (-120.51\%) & 0.0622 (-323.86\%) \\
    % & tsl(cross\_entropy) & 66.1225 & 0.1215 (-124.90\%) & 0.1214 (-124.96\%) & 0.0275 (-118.18\%) & 0.1211 (-124.37\%) \\
    & TSL(Dual Focal) & 66.7371 & 0.0525 (49.45\%) & 0.0525 (49.51\%) & 0.0168 (16.02\%) & 0.0529 (49.11\%) \\
    & TSL(Focal Loss) & 62.5768 & 0.0443 (70.32\%) & 0.0443 (70.33\%) & 0.0144 (48.57\%) & 0.0442 (70.26\%) \\
    & TSL(Focal Calibration Loss) & 67.3171 & 0.0680 (48.13\%) & 0.0672 (48.77\%) & 0.0176 (27.44\%) & 0.0672 (48.60\%) \\
    & TSL(Label Smooth) & 65.2159 & 0.0932 (4.38\%) & 0.0931 (4.51\%) & 0.0184 (28.09\%) & 0.0930 (4.62\%) \\
\midrule
\multirow{10}{*}{Bean} 
    & Brier Score & 89.4985 & 0.0687 & 0.0652 & 0.0298 & 0.0657 \\
    % & cross\_entropy & 93.3866 & 0.0145 & 0.0106 & 0.0070 & 0.0148 \\
    & Dual Focal~\citep{tao2023dual}  & 93.4593 & 0.0785 & 0.0768 & 0.0237 & 0.0773 \\
    & Focal Loss~\citep{lin2017focal}  & 93.3140 & 0.1356 & 0.1353 & 0.0392 & 0.1359 \\
    & Focal Calibration Loss~\citep{liang2024calibrating} & 93.4230 & 0.0841 & 0.0842 & 0.0254 & 0.0843 \\
    & Label Smooth~\citep{szegedy2016rethinking} & 93.3140 & 0.0640 & 0.0599 & 0.0210 & 0.0606 \\
    & TSL(Brier Score) & 92.8052 & 0.0088 (87.18\%) & 0.0064 (90.14\%) & 0.0088 (70.58\%) & 0.0112 (82.88\%) \\
    % & tsl(cross\_entropy) & 93.2049 & 0.0376 (-158.95\%) & 0.0363 (-241.01\%) & 0.0119 (-70.79\%) & 0.0363 (-144.57\%) \\
    & TSL(Dual Focal) & 93.2049 & 0.0189 (75.89\%) & 0.0139 (81.87\%) & 0.0078 (67.09\%) & 0.0178 (76.97\%) \\
    & TSL(Focal Loss) & 92.6599 & 0.0915 (32.55\%) & 0.0891 (34.13\%) & 0.0272 (30.78\%) & 0.0893 (34.27\%) \\
    & TSL(Focal Calibration Loss) & 93.4230 & 0.0197 (76.60\%) & 0.0189 (77.57\%) & 0.0080 (68.44\%) & 0.0194 (76.99\%) \\
    & TSL(Label Smooth) & 93.1323 & 0.0301 (52.99\%) & 0.0306 (49.01\%) & 0.0100 (52.46\%) & 0.0302 (50.24\%) \\
\midrule

\multirow{10}{*}{Rice} 
    & Brier Score & 92.4773 & 0.0390 & 0.0464 & 0.0454 & 0.0390 \\
    % & cross\_entropy & 92.6070 & 0.0218 & 0.0244 & 0.0341 & 0.0166 \\
    & Dual Focal~\citep{tao2023dual}  & 92.3476 & 0.0738 & 0.0754 & 0.0769 & 0.0739 \\
    & Focal Loss~\citep{lin2017focal}  & 91.8288 & 0.1661 & 0.1654 & 0.1709 & 0.1661  \\
    & Focal Calibration Loss~\citep{liang2024calibrating} & 92.6070 & 0.0887 & 0.0880 & 0.0879 & 0.0888 \\
    & Label Smooth~\citep{szegedy2016rethinking} & 92.6070 & 0.0628 & 0.0621 & 0.0582 & 0.0630 \\
    & TSL(Brier Score) & 92.4773 & 0.0295 (24.43\%) & 0.0316 (31.85\%) & 0.0306 (32.55\%) & 0.0254 (34.90\%) \\
    % & tsl(cross\_entropy) & 92.3476 & 0.0392 (-79.66\%) & 0.0321 (-31.87\%) & 0.0390 (-14.52\%) & 0.0385 (-131.36\%) \\
    & TSL(Dual Focal) & 92.2179 & 0.0329 (55.46\%) & 0.0408 (45.84\%) & 0.0451 (41.30\%) & 0.0328 (55.56\%) \\
    & TSL(Focal Loss) & 92.6070 & 0.1538 (7.98\%) & 0.1531 (8.02\%) & 0.1480 (15.48\%) & 0.1541(7.78\%) \\
    & TSL(Focal Calibration Loss) & 92.6070 & 0.0245 (72.42\%) & 0.0291 (66.97\%) & 0.0231 (73.77\%) & 0.0254 (71.36\%) \\
    & TSL(Label Smooth) & 92.4773 & 0.0346 (44.83\%) & 0.0283 (54.50\%) & 0.0376 (35.41\%) & 0.0294 (53.33\%) \\
\midrule
\multirow{10}{*}{Spam} 
    & Brier Score & 94.9079 & 0.0586 & 0.0580 & 0.0556 & 0.0587 \\
    % & cross\_entropy & 94.9079 & 0.0193 & 0.0202 & 0.0225 & 0.0202 \\
    & Dual Focal~\citep{tao2023dual}  & 95.4496 & 0.0802 & 0.0797 & 0.0772 & 0.0804 \\
    & Focal Loss~\citep{lin2017focal}  & 94.2579 & 0.1896 & 0.1891 & 0.1838 & 0.1887 \\
    & Focal Calibration Loss~\citep{liang2024calibrating} & 95.2329 & 0.0845 & 0.0840 & 0.0848 & 0.0846 \\
    & Label Smooth~\citep{szegedy2016rethinking} & 94.7996 & 0.0707 & 0.0702 & 0.0702 & 0.0709 \\
    & TSL(Brier Score) & 94.7996 & 0.0198 (66.21\%) & 0.0172 (70.38\%) & 0.0252 (54.63\%) & 0.0209 (64.40\%) \\
    % & tsl(cross\_entropy) & 94.9079 & 0.0253 (-31.20\%) & 0.0201 (0.51\%) & 0.0276 (-22.63\%) & 0.0254 (-25.88\%) \\
    & TSL(Dual Focal) & 94.0412 & 0.0440 (45.20\%) & 0.0360 (54.80\%) & 0.0438 (43.25\%) & 0.0407 (49.35\%) \\
    & TSL(Focal Loss) & 94.3662 & 0.1215 (56.09\%) & 0.1209 (56.34\%) &  0.1229 (49.54\%) & 0.1216 (55.10\%) \\
    & TSL(Focal Calibration Loss) & 95.2329 & 0.0214 (74.71\%) & 0.0216 (74.28\%) & 0.0245 (71.10\%) & 0.0204 (75.94\%) \\
    & TSL(Label Smooth) & 94.7996 & 0.0162 (77.08\%) & 0.0148 (78.88\%) & 0.0185 (73.61\%) & 0.0174 (75.40\%) \\
% \midrule
% \multirow{12}{*}{Telescope} 
%     & brier\_score & 86.0660 & 0.0101 & 0.0136 & 0.0167 & 0.0128 \\
%     & cross\_entropy & 86.4589 & 0.0113 & 0.0128 & 0.0122 & 0.0145 \\
%     & dual\_focal & 85.9874 & 0.0674 & 0.0724 & 0.0708 & 0.0674 \\
%     & focal & 85.9612 & 0.1553 & 0.1552 & 0.1529 & 0.1553 \\
%     & focal\_calibration & 86.2232 & 0.0908 & 0.0909 & 0.0929 & 0.0909 \\
%     & label\_smooth & 86.1708 & 0.0316 & 0.0321 & 0.0331 & 0.0311 \\
%     & tsl(brier\_score) & 85.6469 & 0.0732 (-625.26\%) & 0.0702 (-416.07\%) & 0.0733 (-339.29\%) & 0.0719 (-460.67\%) \\
%     & tsl(cross\_entropy) & 86.1708 & 0.0775 (-584.47\%) & 0.0761 (-495.47\%) & 0.0811 (-564.74\%) & 0.0755 (-420.40\%) \\
%     & TSL(Dual Focal) & 84.7302 & 0.0436 (35.33\%) & 0.0528 (27.11\%) & 0.0495 (30.11\%) & 0.0379 (43.80\%) \\
%     & TSL(Focal Loss) & 84.5731 & 0.1794 (-15.46\%) & 0.1792 (-15.47\%) & 0.1798 (-17.54\%) & 0.1786 (-14.94\%) \\
%     & TSL(Focal Calibration Loss) & 86.3279 & 0.0507 (44.15\%) & 0.0488 (46.38\%) & 0.0529 (43.02\%) & 0.0506 (44.32\%) \\
%     & TSL(Label Smooth) & 86.1708 & 0.0614 (-94.22\%) & 0.0587 (-82.81\%) & 0.0629 (-89.87\%) & 0.0606 (-94.58\%) \\

\midrule
\multirow{2}{*}{CIFAR10} 
    & TSL(Dual Focal) & 50.0300 & 0.0266  & 0.0273 & 0.0098 & 0.0260 \\   
    & TSL(Focal Calibration Loss) & 50.1500 &  0.1076 &  0.1077  & 0.0280 & 0.1077 \\
\midrule

\bottomrule
\end{tabular}
\end{adjustbox}
\end{table}

\footnotetext[2]{For CIFAR-10, only two losses achieved a test accuracy above 50\% within 20 epoches. Losses with test accuracy below 50\% are omitted, as they are considered meaningless and perform worse than random guessing.}

\section{KANs Hyperparameter vs. Calibration Metrics}
\label{app:Kan_hyper}

\begin{figure*}[ht]
\vskip 0.2in
\begin{center}
\resizebox{\textwidth}{!}{%
\begin{tabular}{ccccc}
    \subfigure[AdaECE: Layer Width]{
        \includegraphics[width=0.18\linewidth]{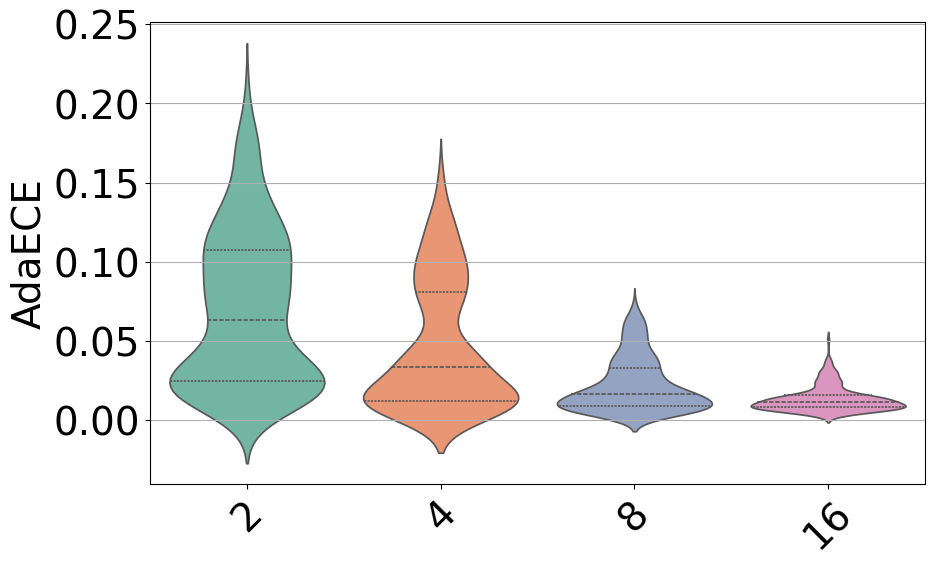}
    } &
    \subfigure[AdaECE: Grid Range]{
        \includegraphics[width=0.18\linewidth]{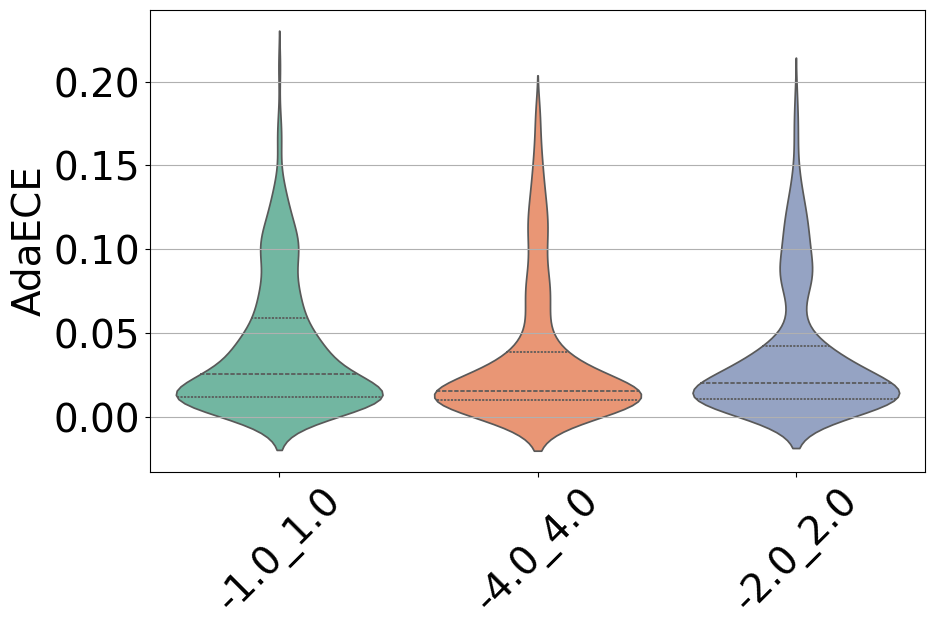}
    } &
    \subfigure[AdaECE: Grid Order]{
        \includegraphics[width=0.18\linewidth]{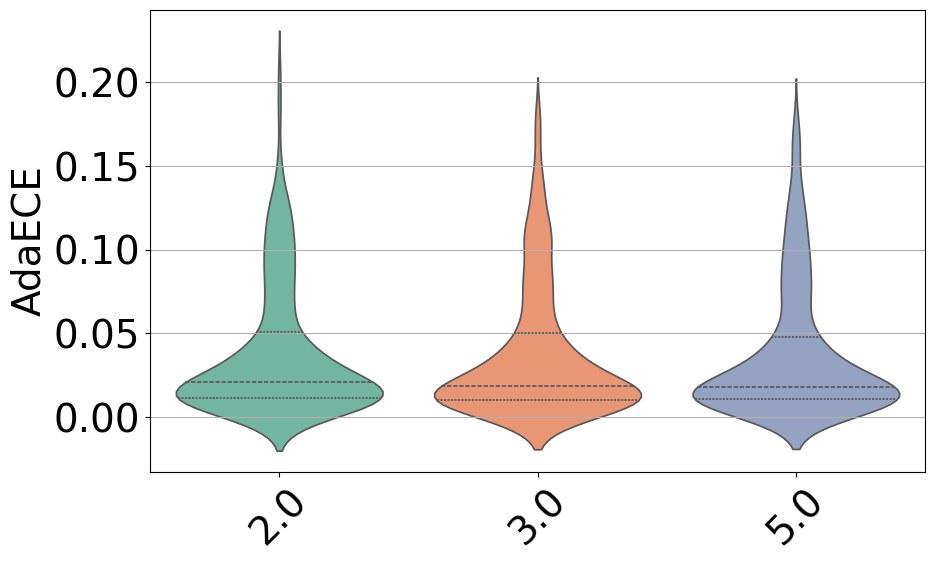}
    } &
    \subfigure[AdaECE: Shortcut]{
        \includegraphics[width=0.18\linewidth]{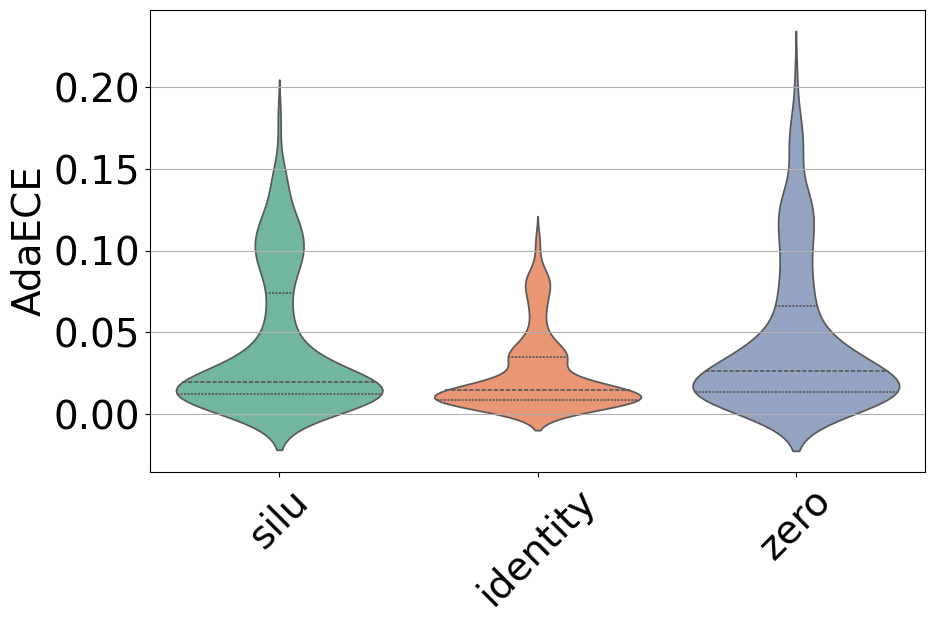}
    } &
    \subfigure[AdaECE:Params]{
        \includegraphics[width=0.18\linewidth]{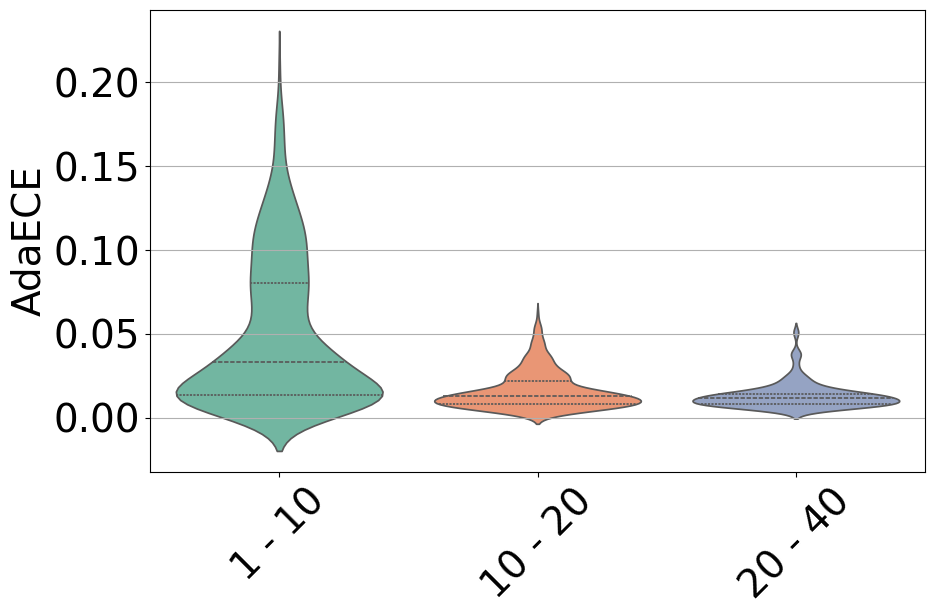}
    } \\
    \subfigure[CECE: Layer Width]{
        \includegraphics[width=0.18\linewidth]{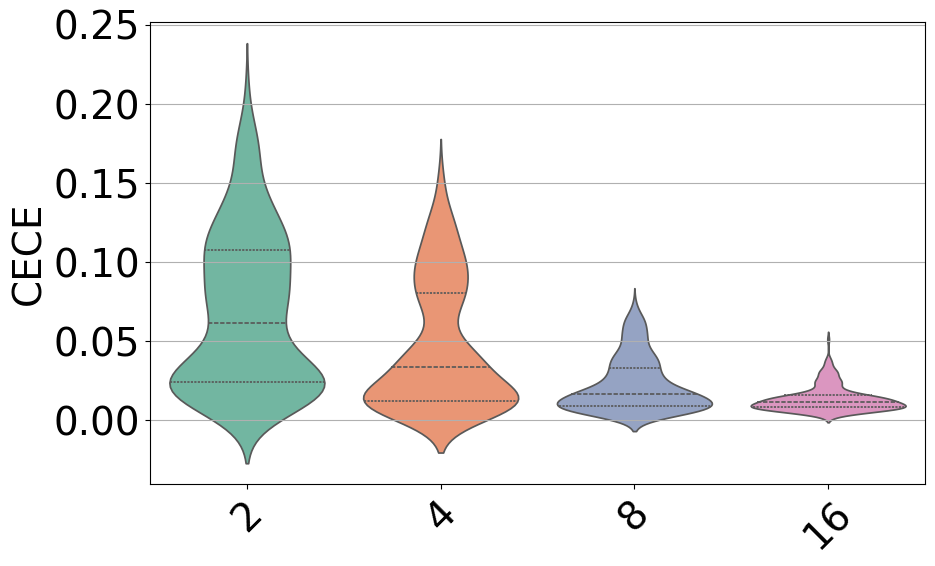}
    } &
    \subfigure[CECE: Grid Range]{
        \includegraphics[width=0.18\linewidth]{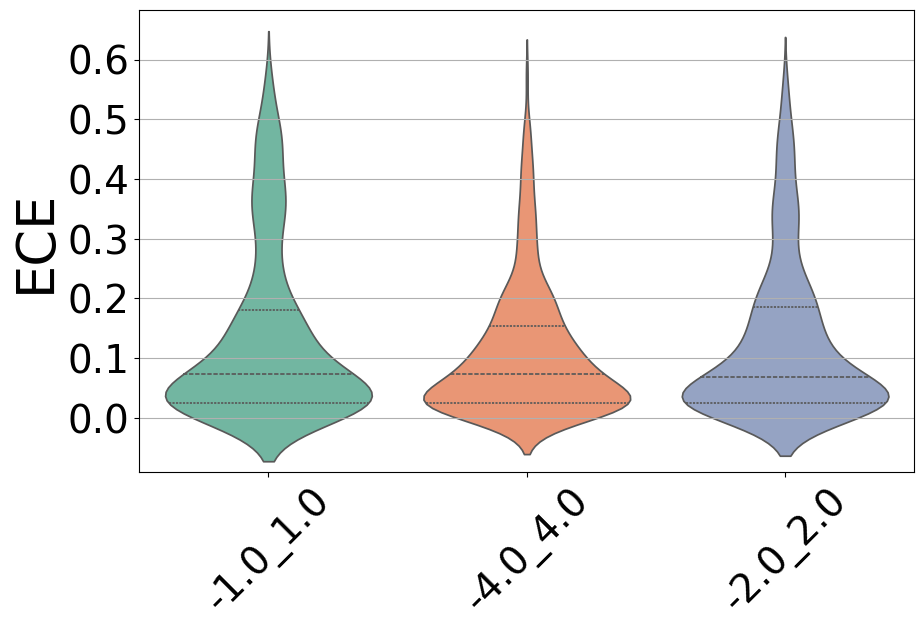}
    } &
    \subfigure[CECE: Grid Order]{
        \includegraphics[width=0.18\linewidth]{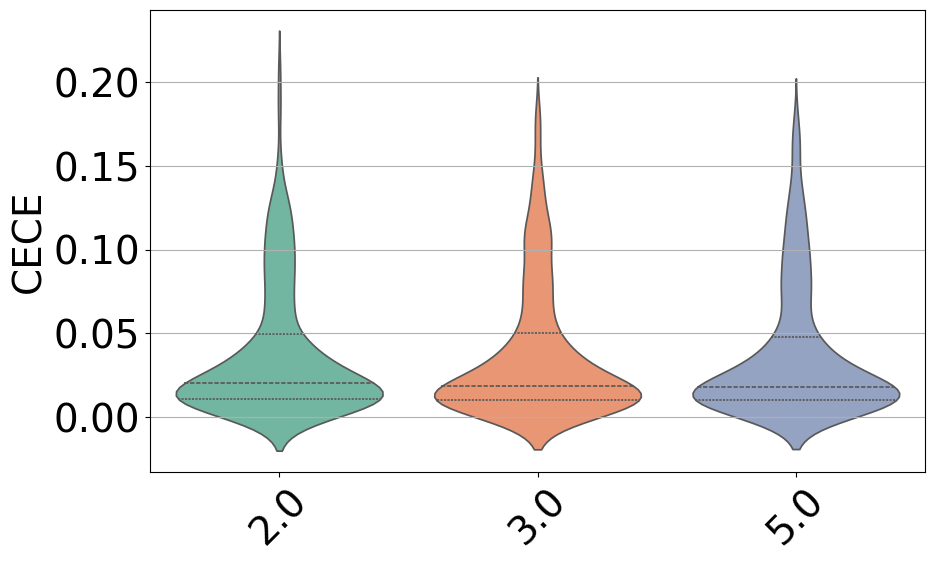}
    } &
    \subfigure[CECE: Shortcut]{
        \includegraphics[width=0.18\linewidth]{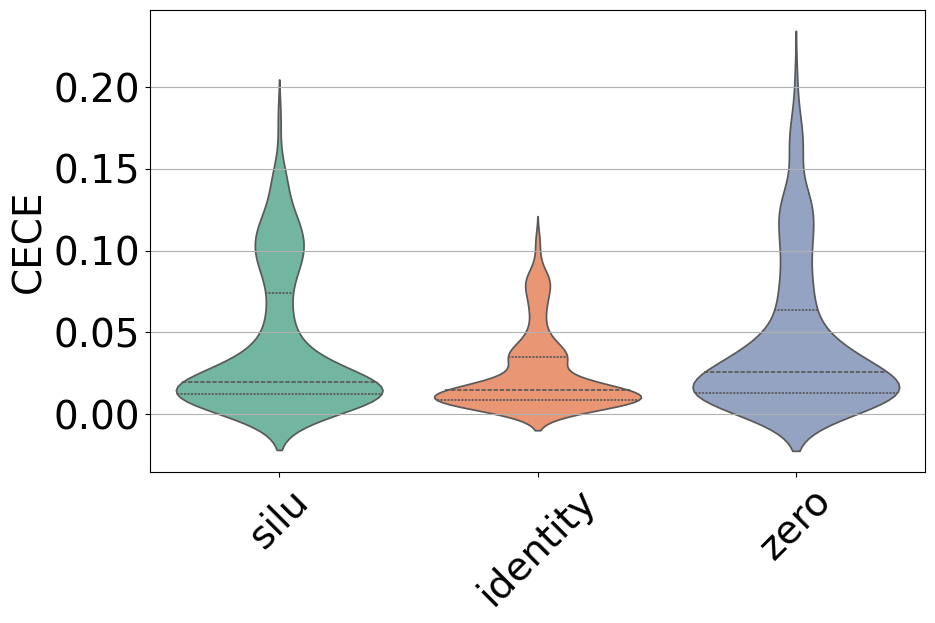}
    } &
    \subfigure[CECE:Params]{
        \includegraphics[width=0.18\linewidth]{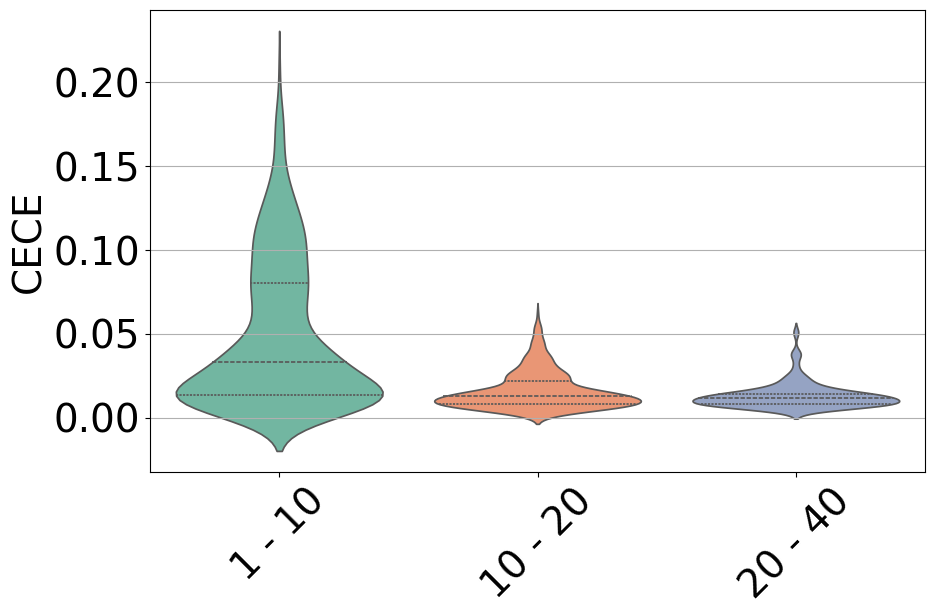}
    } \\
    \subfigure[SmECE: Layer Width]{
        \includegraphics[width=0.18\linewidth]{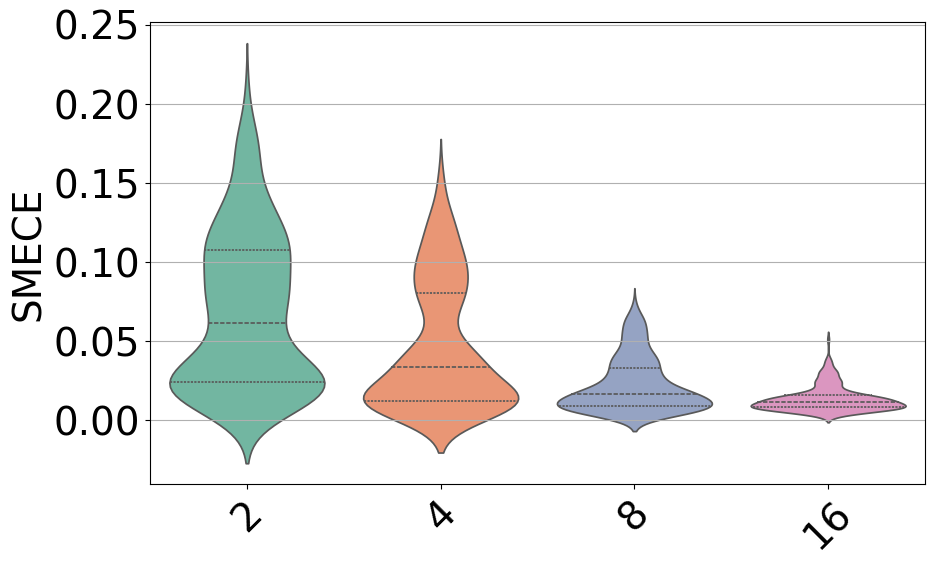}
    } &
    \subfigure[SmECE: Grid Range]{
        \includegraphics[width=0.18\linewidth]{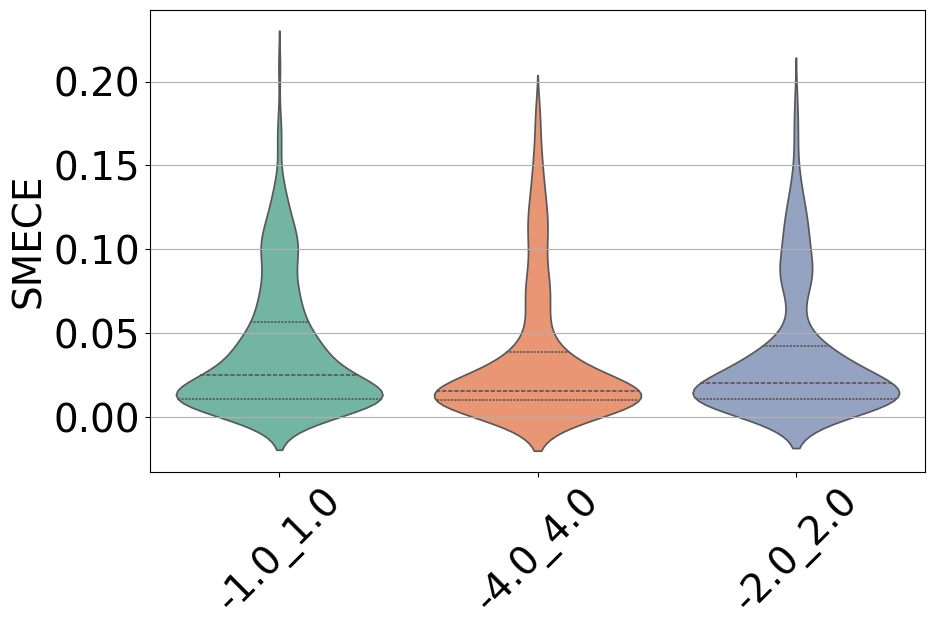}
    } &
    \subfigure[SmECE: Grid Order]{
        \includegraphics[width=0.18\linewidth]{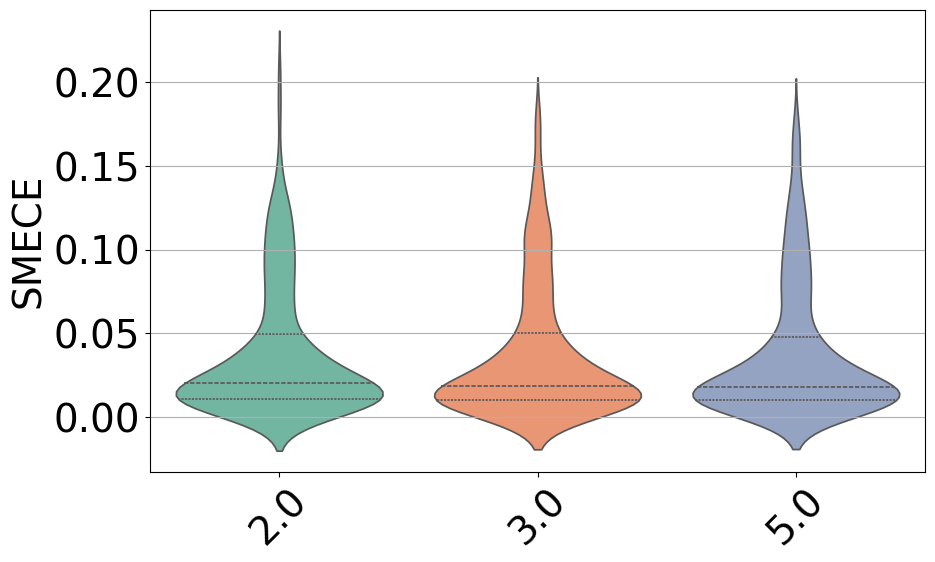}
    } &
    \subfigure[SmECE: Shortcut]{
        \includegraphics[width=0.18\linewidth]{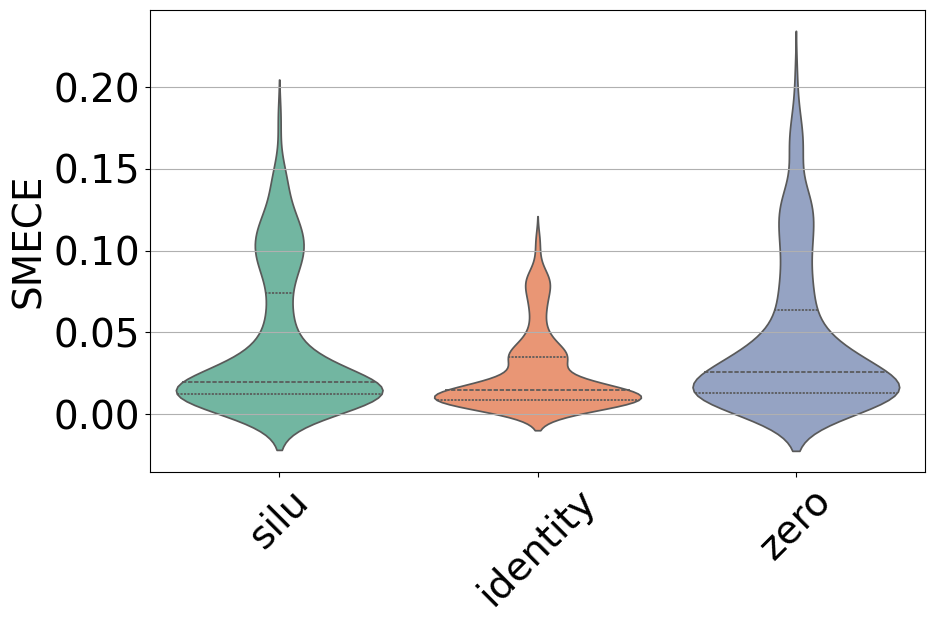}
    } &
    \subfigure[SmECE: Params]{
        \includegraphics[width=0.18\linewidth]{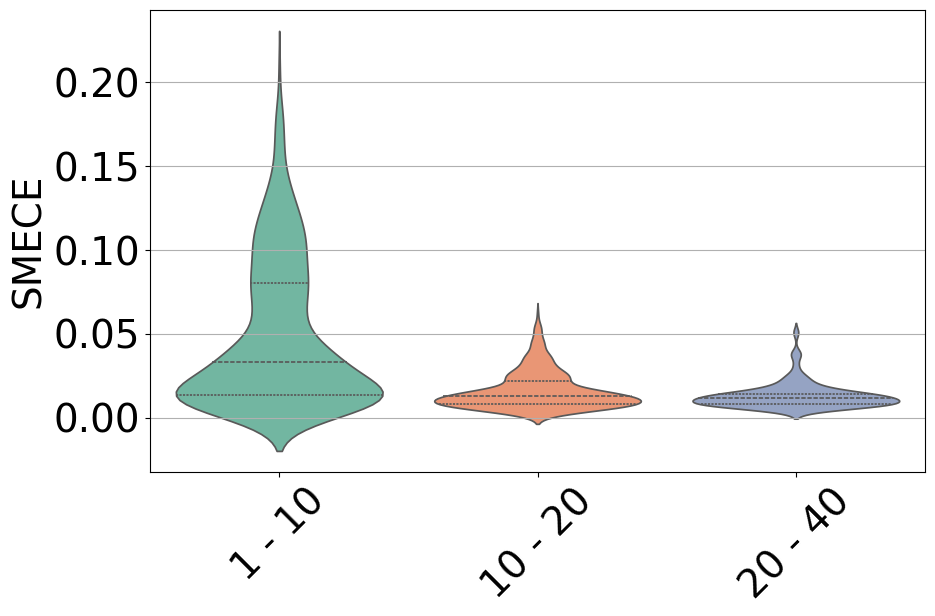}
    }
\end{tabular}
}
\caption{KAN Key Parameters vs. Calibration Metrics. Each row corresponds to a different calibration metric: AdaECE (top row), Classwise-ECE (middle row), and SmoothECE (bottom row). Columns represent the effects of Layer Width, Grid Range, Grid Order, Shortcut, and Number of Parameters ($10^4$) on calibration performance.}
\Description{tbc}
\label{fig:multiple_factors_combined}
\end{center}
\vskip -0.2in
\end{figure*}

\section{KANs Hyperparameter vs. Different Losses on ECE}

\begin{figure*}[ht]
\vskip 0.2in
\begin{center}
\resizebox{\textwidth}{!}{%
\begin{tabular}{ccccc}
    \subfigure[CE: Layer Width]{
        \includegraphics[width=0.18\linewidth]{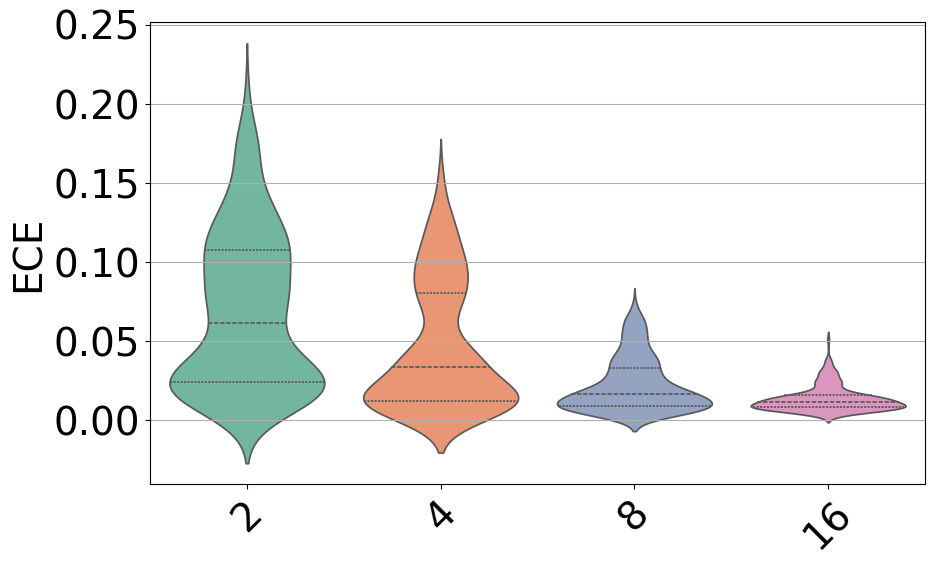}
    } &
    \subfigure[CE: Grid Range]{
        \includegraphics[width=0.18\linewidth]{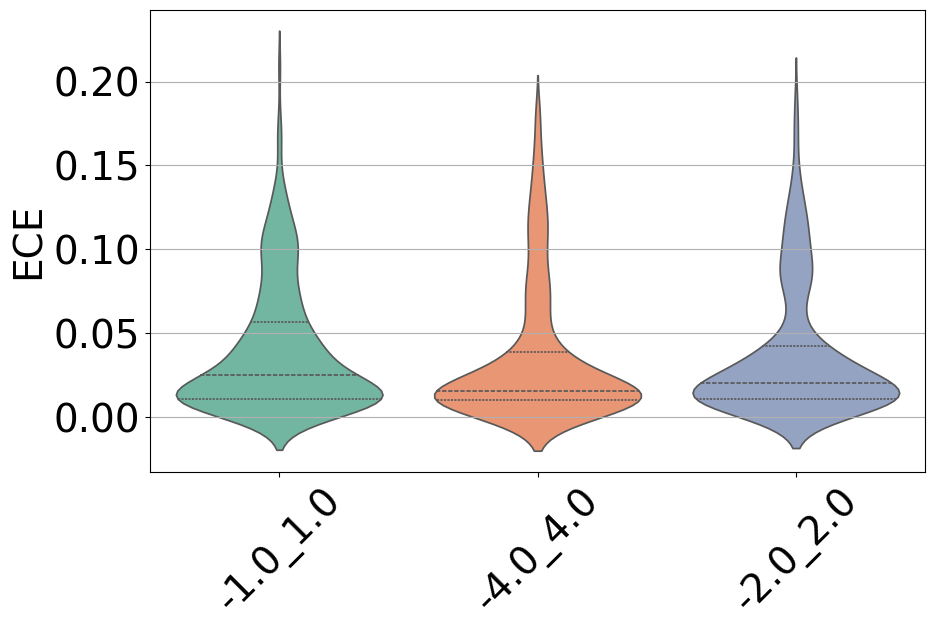}
    } &
    \subfigure[CE: Grid Order]{
        \includegraphics[width=0.18\linewidth]{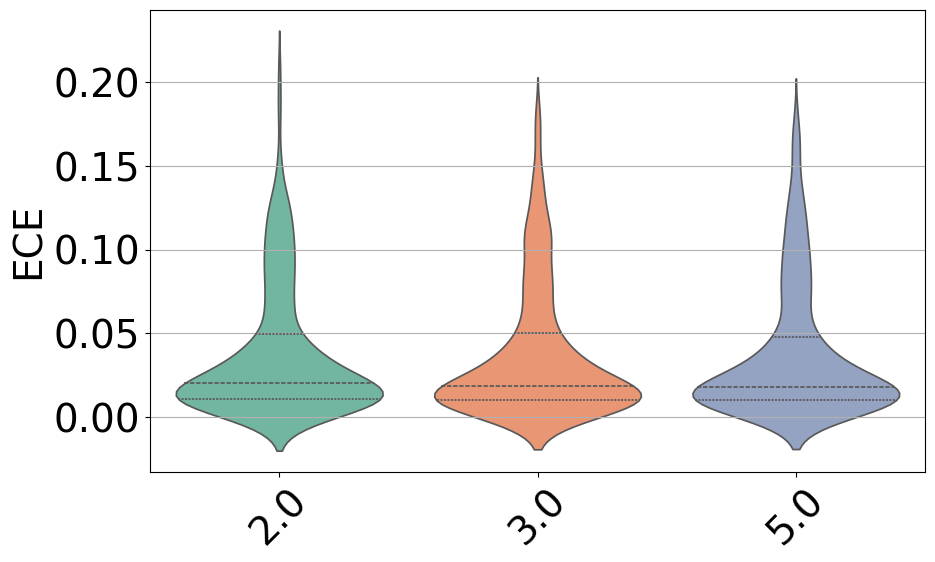}
    } &
    \subfigure[CE: Shortcut]{
        \includegraphics[width=0.18\linewidth]{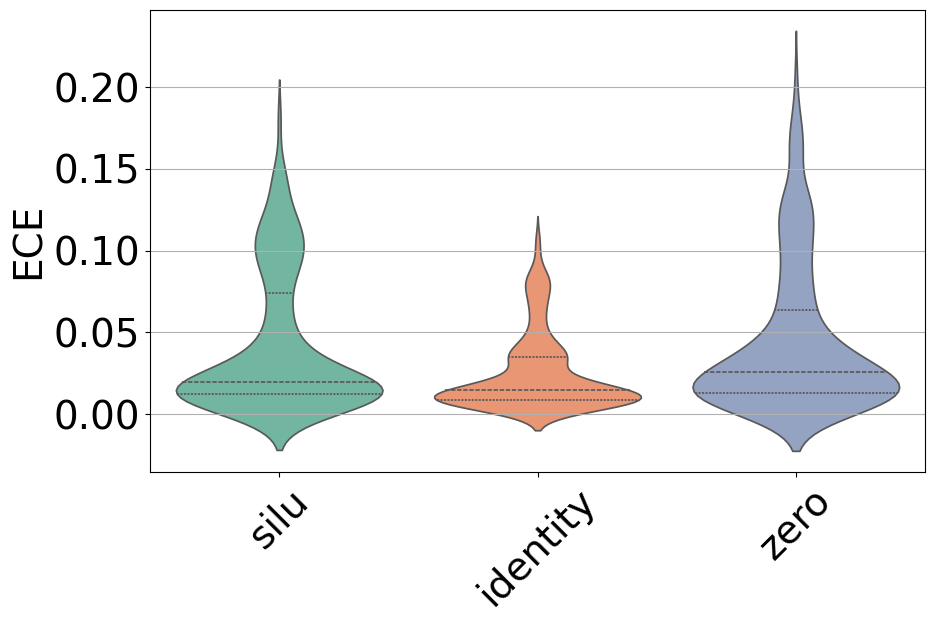}
    } &
    \subfigure[CE:Params]{
        \includegraphics[width=0.18\linewidth]{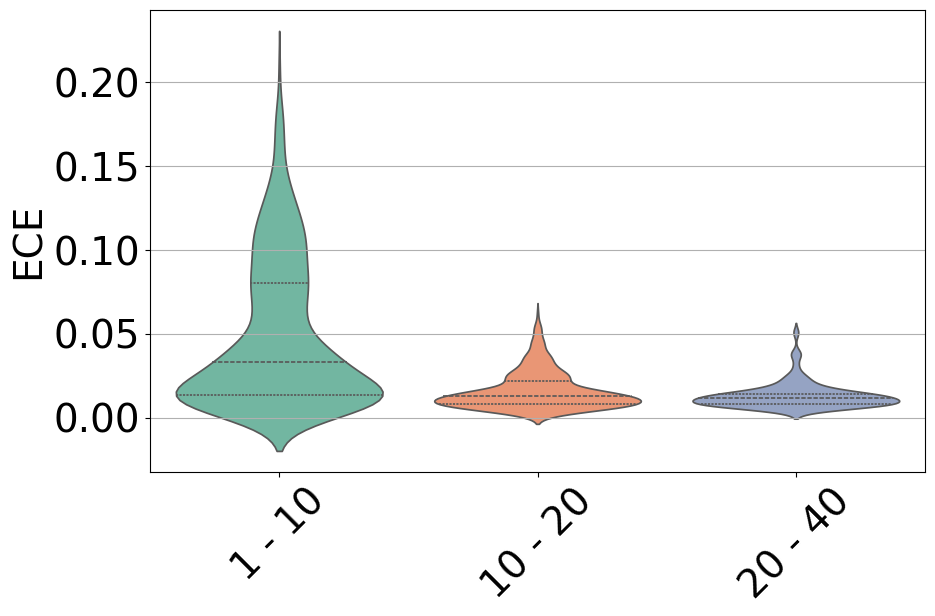}
    } \\
    \subfigure[TSL(CE): Layer Width]{
        \includegraphics[width=0.18\linewidth]{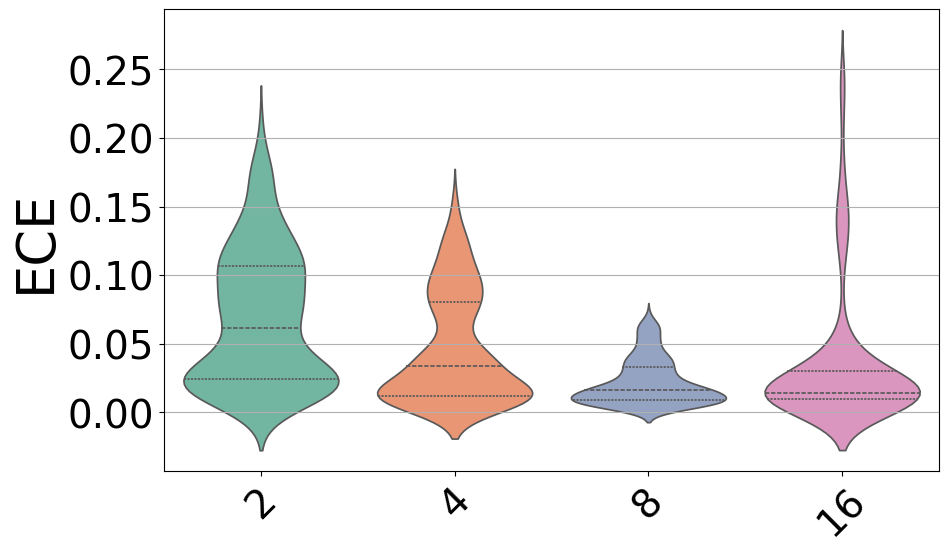}
    } &
    \subfigure[TSL(CE): Grid Range]{
        \includegraphics[width=0.18\linewidth]{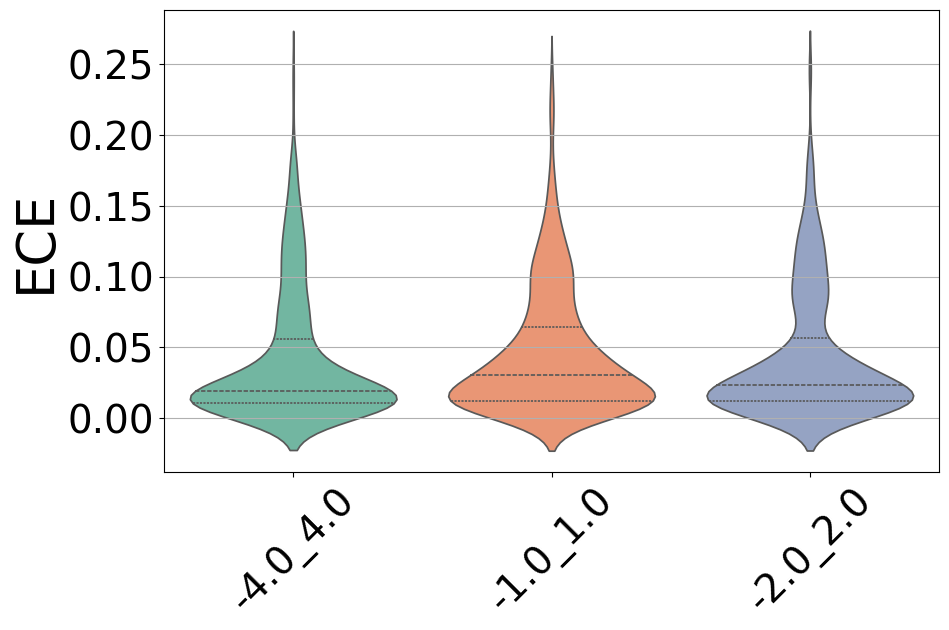}
    } &
    \subfigure[TSL(CE): Grid Order]{
        \includegraphics[width=0.18\linewidth]{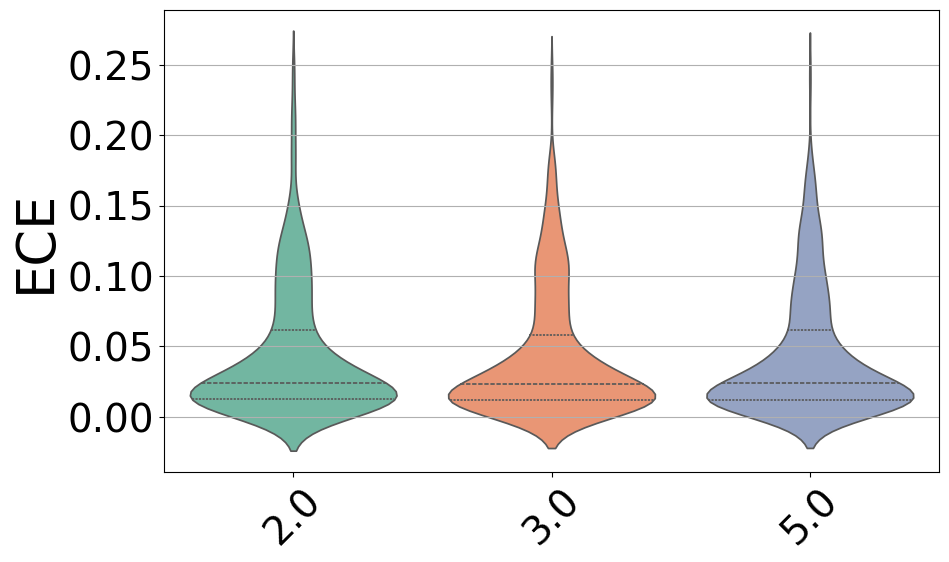}
    } &
    \subfigure[TSL(CE): Shortcut]{
        \includegraphics[width=0.18\linewidth]{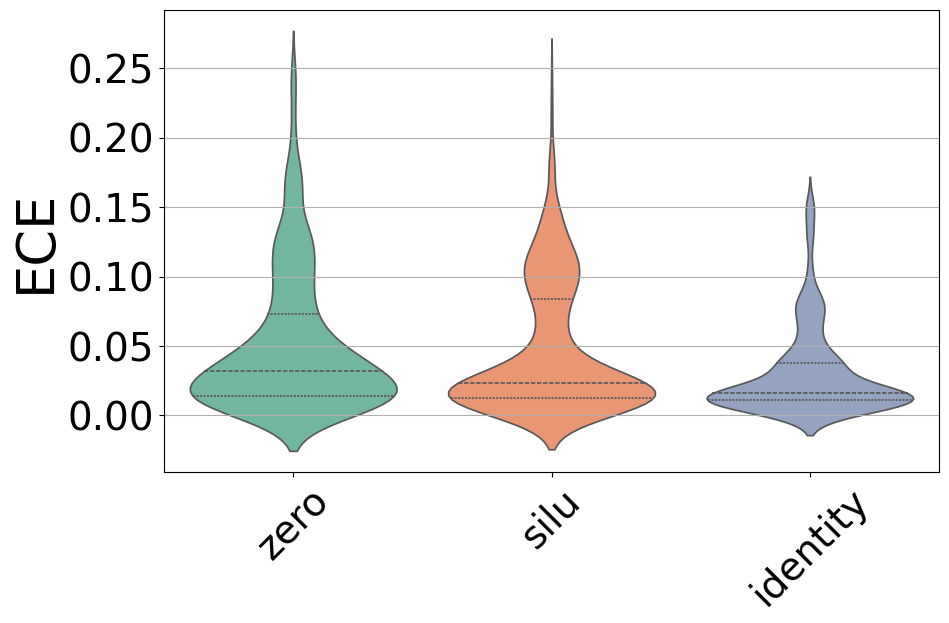}
    } &
    \subfigure[TSL(CE):Params]{
        \includegraphics[width=0.18\linewidth]{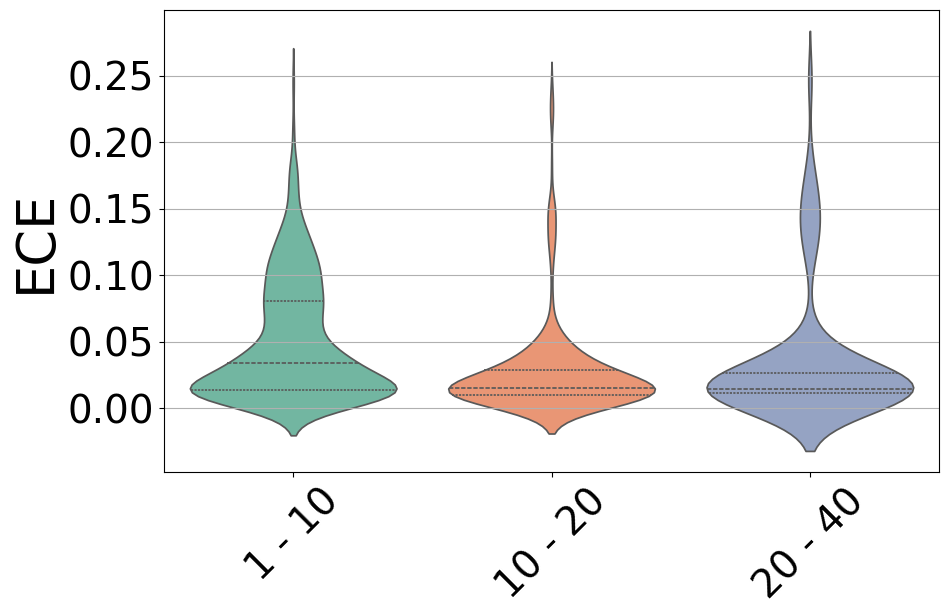}
    } \\
\end{tabular}
}
\caption{Comparison of KAN calibration metrics with and without Temperature-Scaled Loss (TSL). The top row shows Cross-Entropy (CE) results, while the bottom row shows TSL applied to CE. Columns illustrate the effects of Layer Width, Grid Range, Grid Order, Shortcut, and Number of Parameters ($10^4$) on calibration performance.}
\Description{tbc}
\label{fig:multiple_factors_tsl_ce}
\end{center}
\vskip -0.2in
\end{figure*}

\begin{figure*}[ht]
\vskip 0.2in
\begin{center}
\resizebox{\textwidth}{!}{%
\begin{tabular}{ccccc}
    \subfigure[BS: Layer Width]{
        \includegraphics[width=0.18\linewidth]{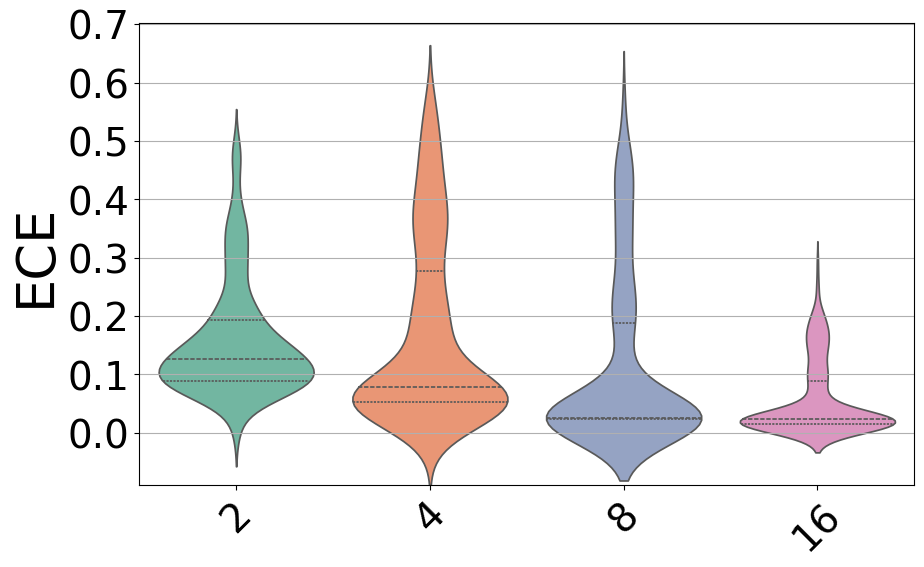}
    } &
    \subfigure[BS: Grid Range]{
        \includegraphics[width=0.18\linewidth]{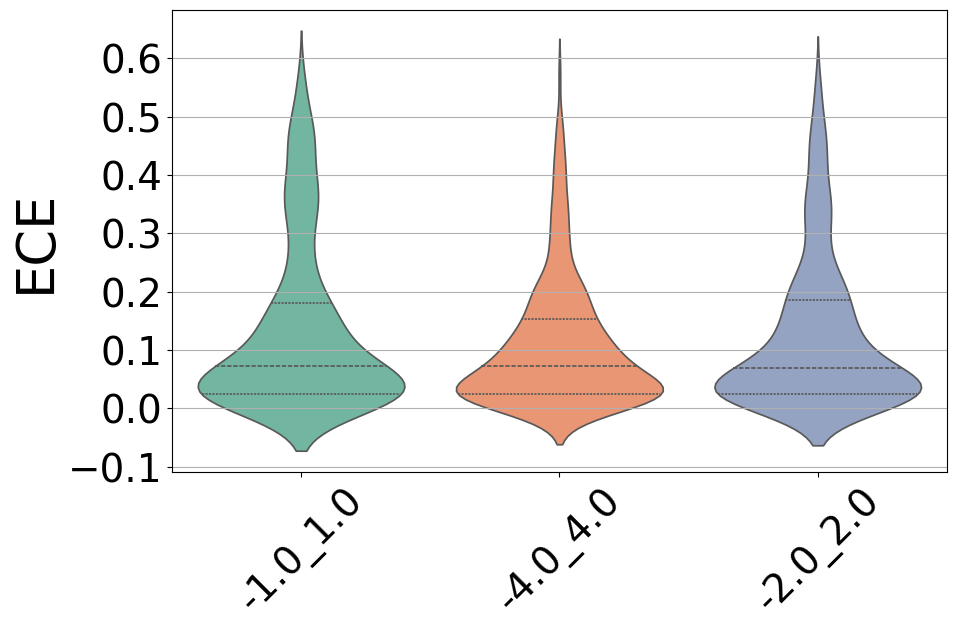}
    } &
    \subfigure[BS: Grid Order]{
        \includegraphics[width=0.18\linewidth]{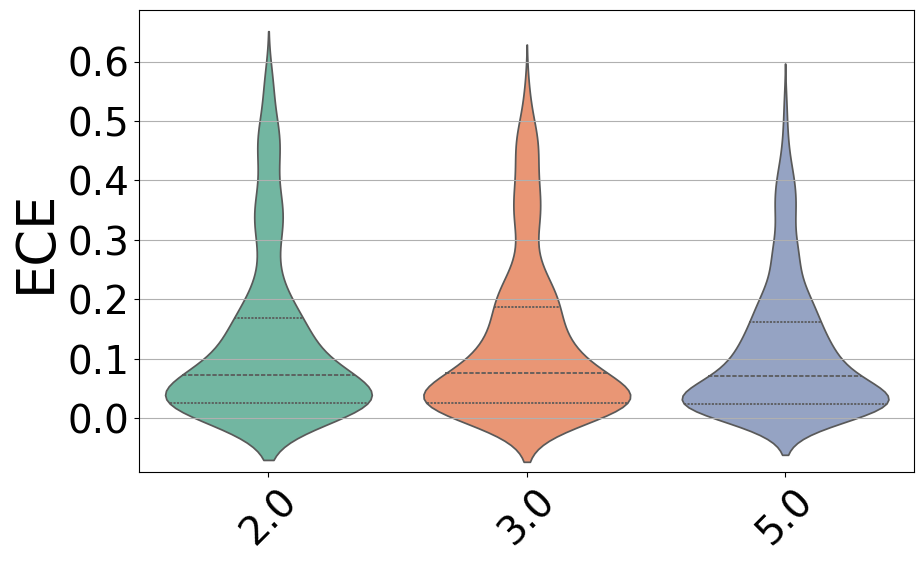}
    } &
    \subfigure[BS: Shortcut]{
        \includegraphics[width=0.18\linewidth]{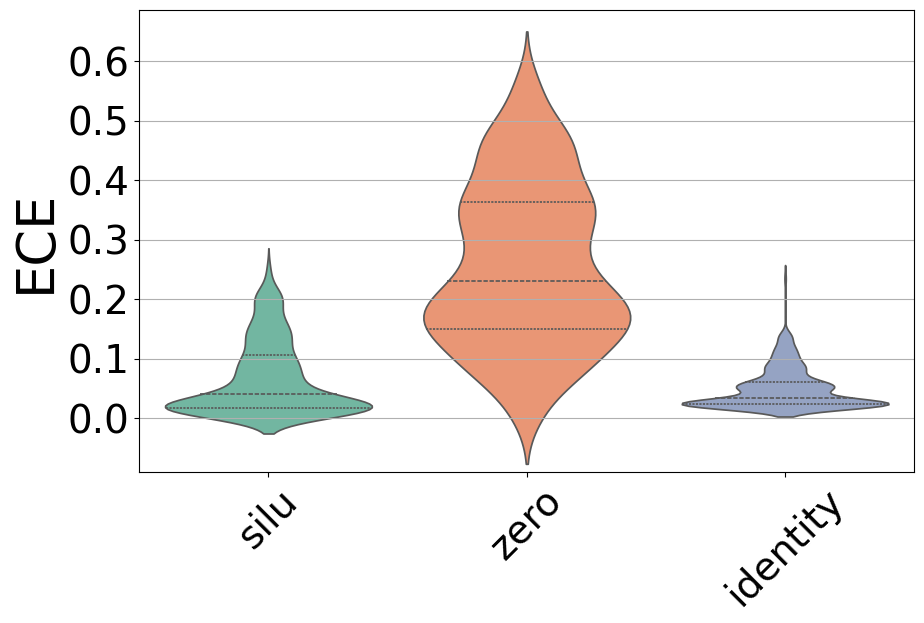}
    } &
    \subfigure[BS:Params]{
        \includegraphics[width=0.18\linewidth]{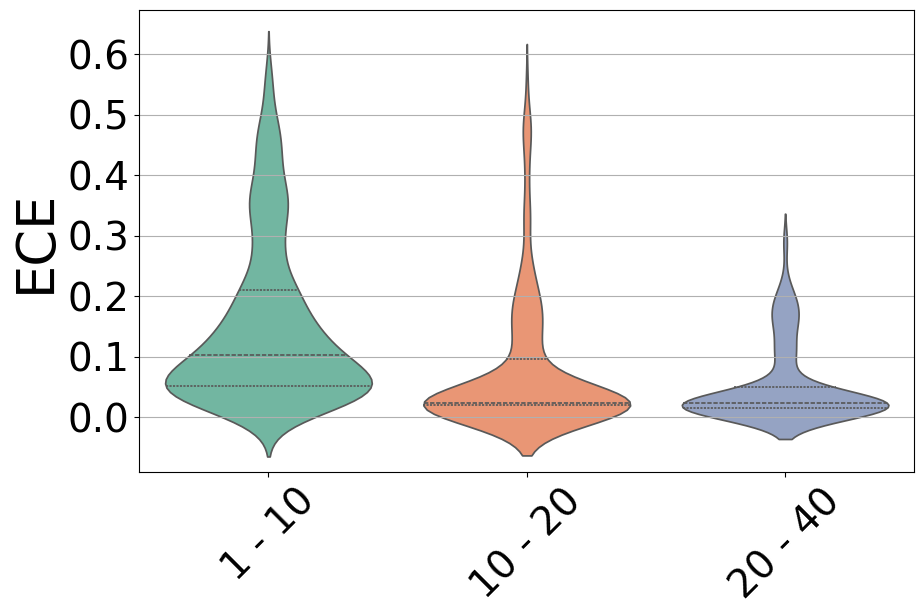}
    } \\
    \subfigure[TSL(BS): Layer Width]{
        \includegraphics[width=0.18\linewidth]{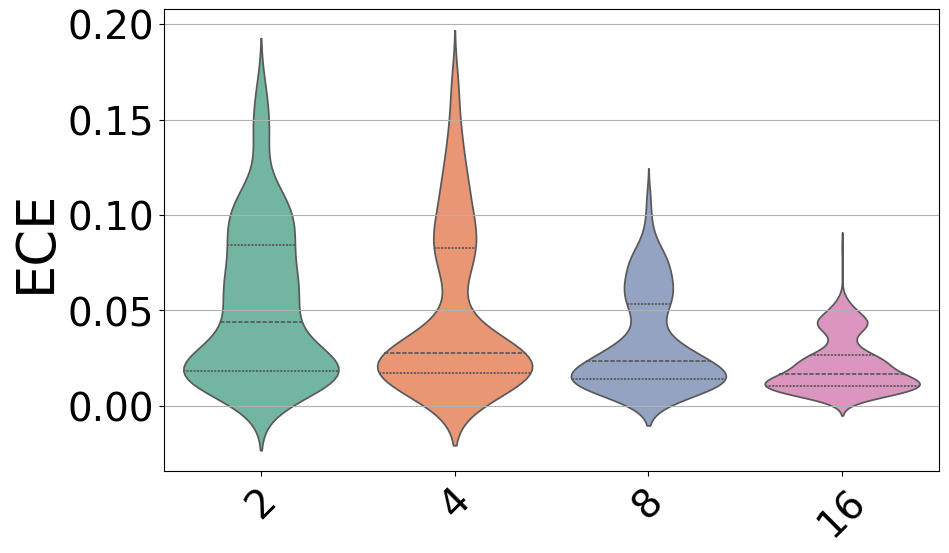}
    } &
    \subfigure[TSL(BS): Grid Range]{
        \includegraphics[width=0.18\linewidth]{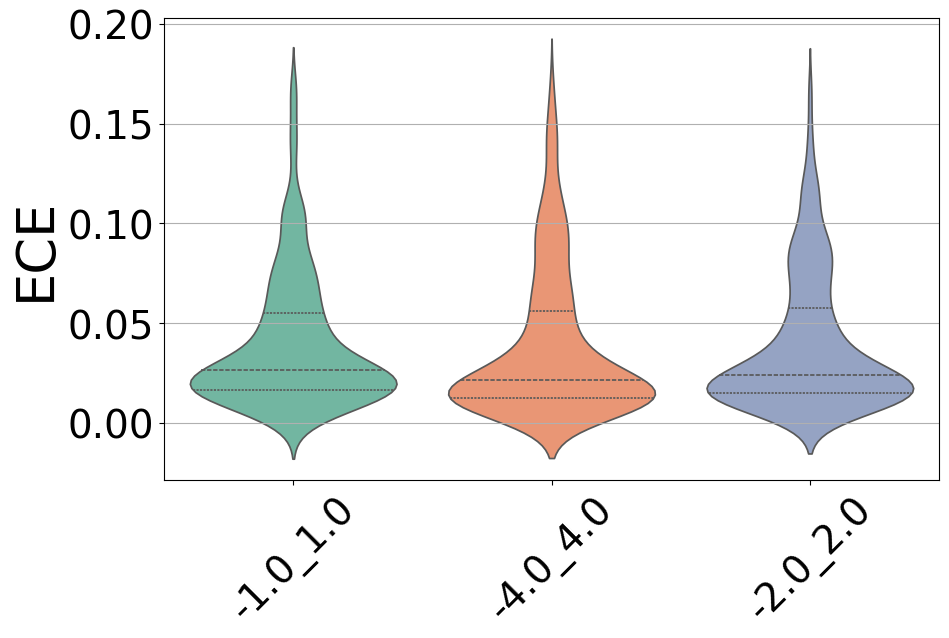}
    } &
    \subfigure[TSL(BS): Grid Order]{
        \includegraphics[width=0.18\linewidth]{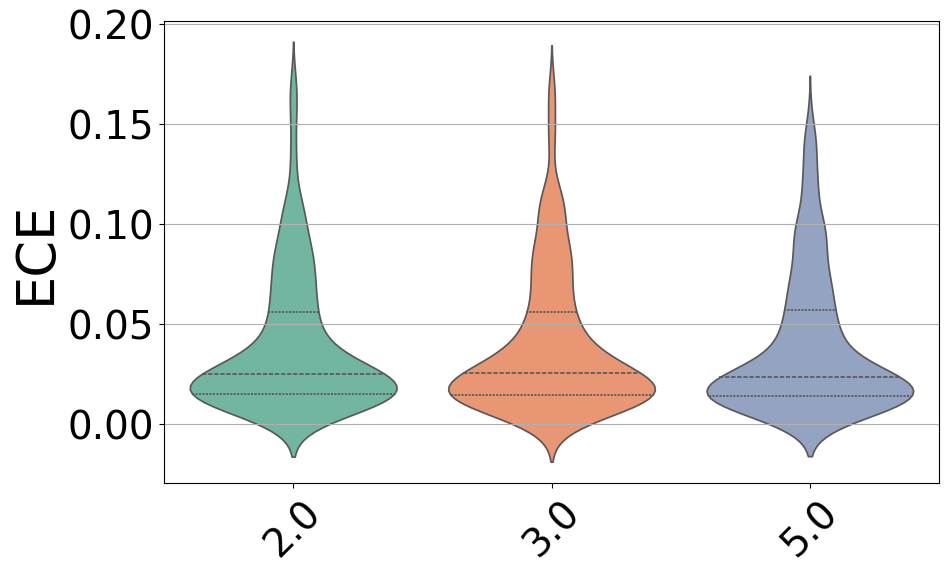}
    } &
    \subfigure[TSL(BS): Shortcut]{
        \includegraphics[width=0.18\linewidth]{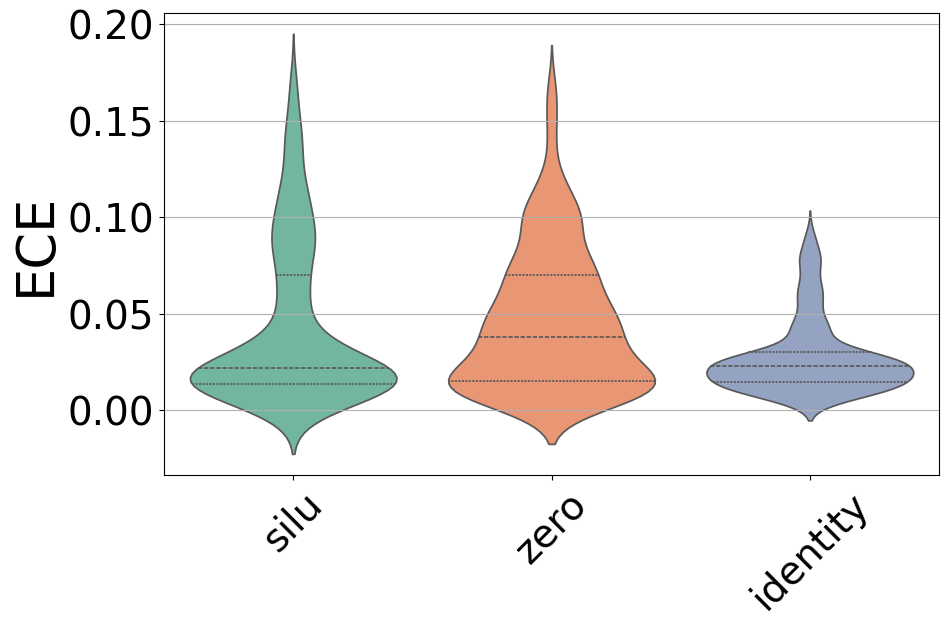}
    } &
    \subfigure[TSL(BS):Params]{
        \includegraphics[width=0.18\linewidth]{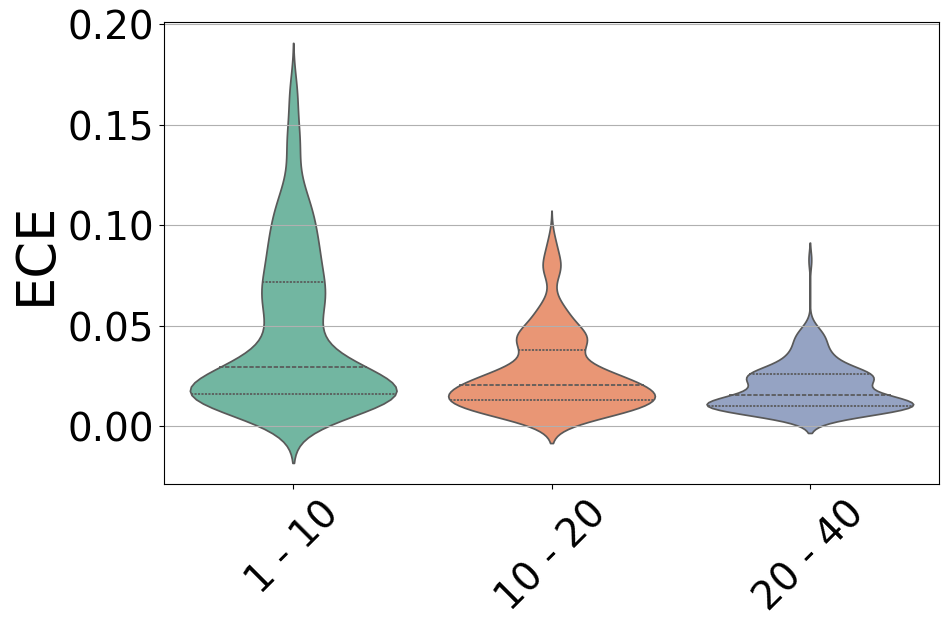}
    } \\
\end{tabular}
}
\caption{Comparison of KAN calibration metrics with and without Temperature-Scaled Loss (TSL). The top row shows Brier Score (BS) results, while the bottom row shows TSL applied to BS. Columns illustrate the effects of Layer Width, Grid Range, Grid Order, Shortcut, and Number of Parameters ($10^4$) on calibration performance.}
\Description{tbc}
\label{fig:multiple_factors_tsl_bs}
\end{center}
\vskip -0.2in
\end{figure*}

\begin{figure*}[ht]
\vskip 0.2in
\begin{center}
\resizebox{\textwidth}{!}{%
\begin{tabular}{ccccc}
    \subfigure[LS: Layer Width]{
        \includegraphics[width=0.18\linewidth]{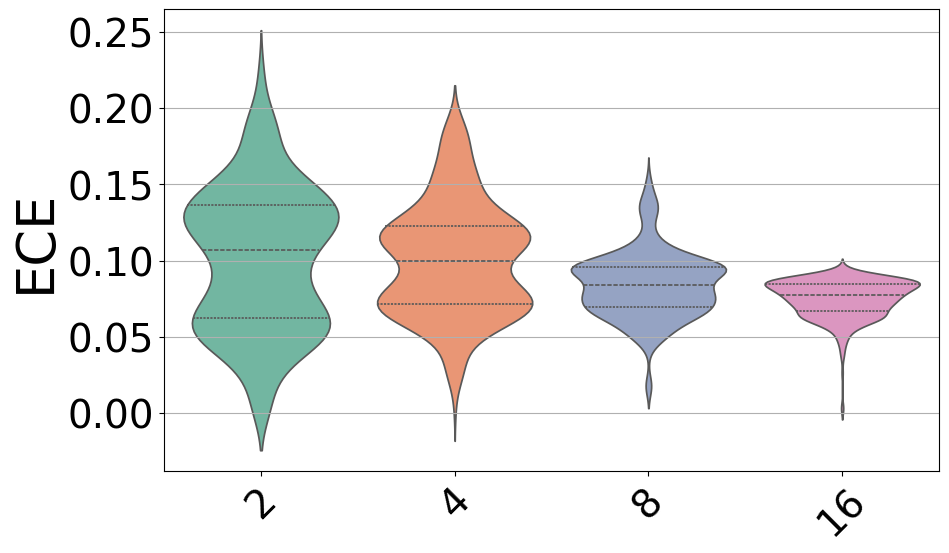}
    } &
    \subfigure[LS: Grid Range]{
        \includegraphics[width=0.18\linewidth]{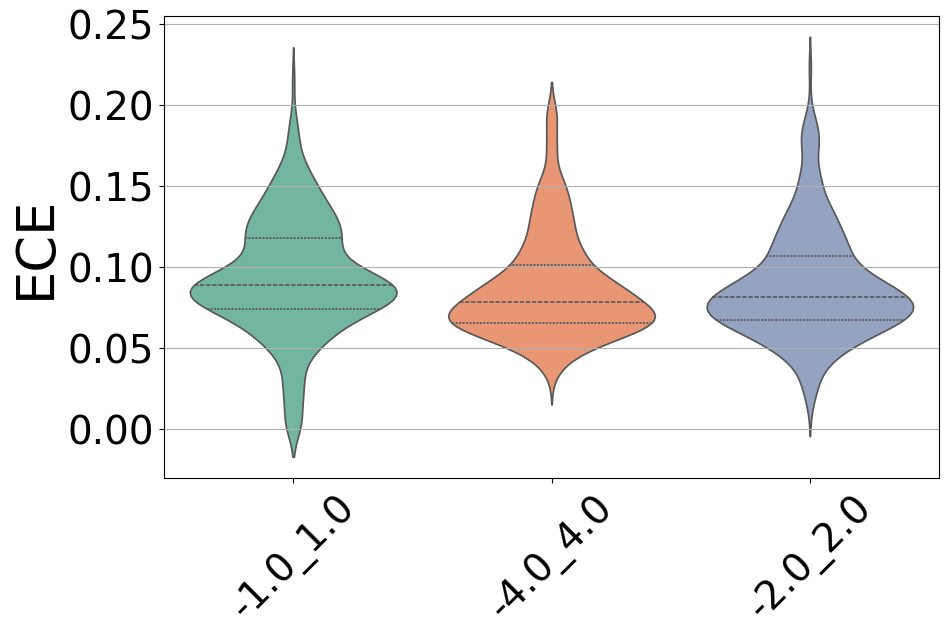}
    } &
    \subfigure[LS: Grid Order]{
        \includegraphics[width=0.18\linewidth]{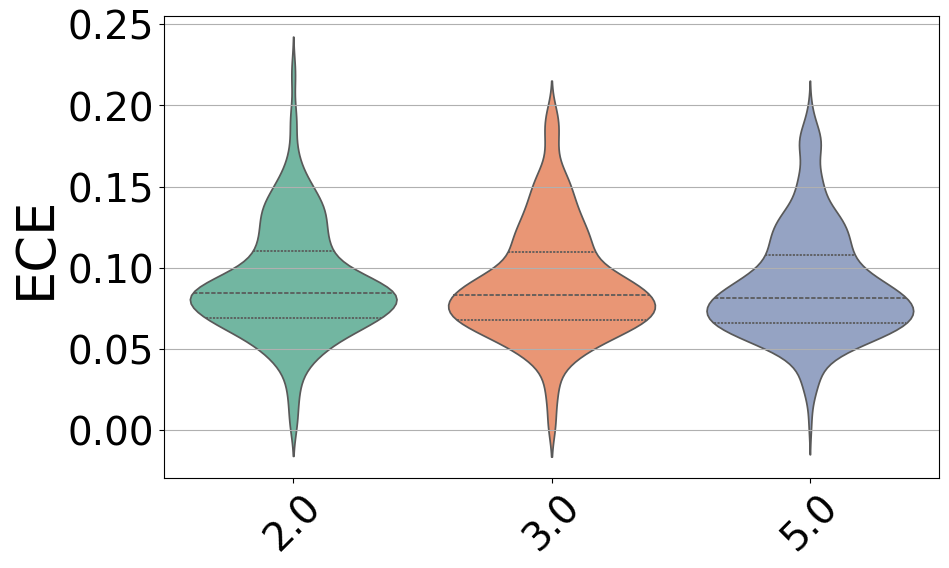}
    } &
    \subfigure[LS: Shortcut]{
        \includegraphics[width=0.18\linewidth]{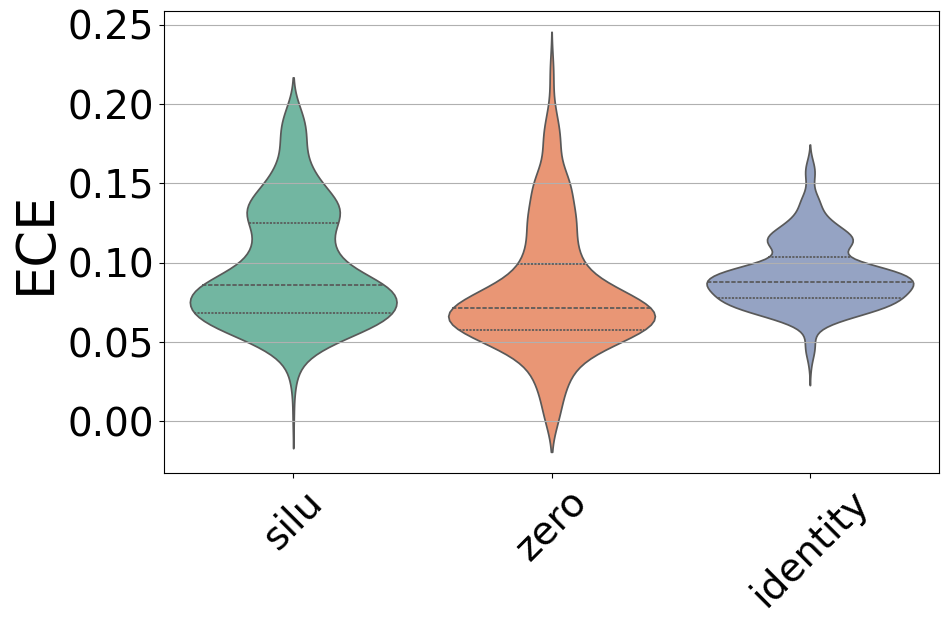}
    } &
    \subfigure[LS:Params]{
        \includegraphics[width=0.18\linewidth]{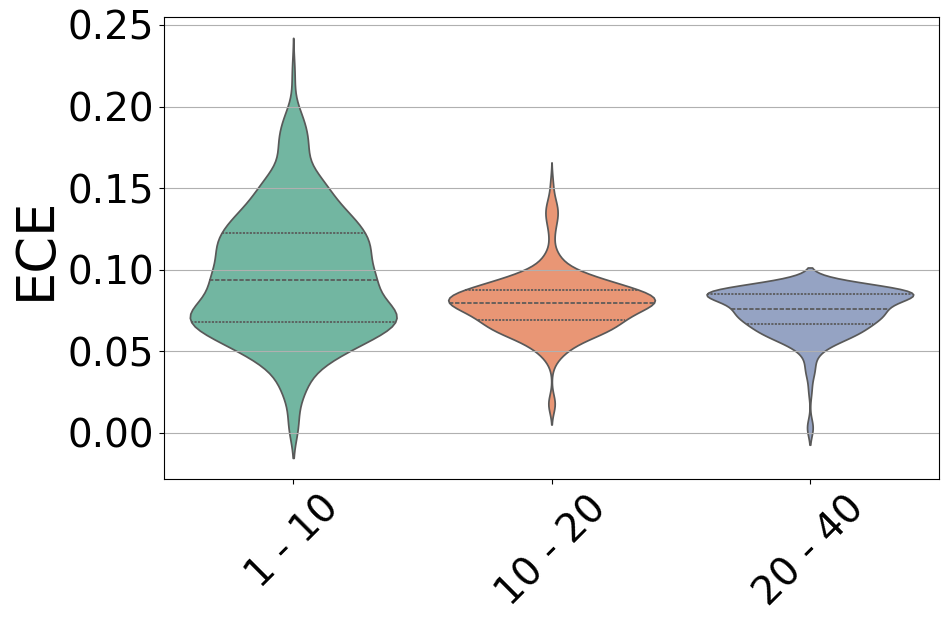}
    } \\
    \subfigure[TSL(LS): Layer Width]{
        \includegraphics[width=0.18\linewidth]{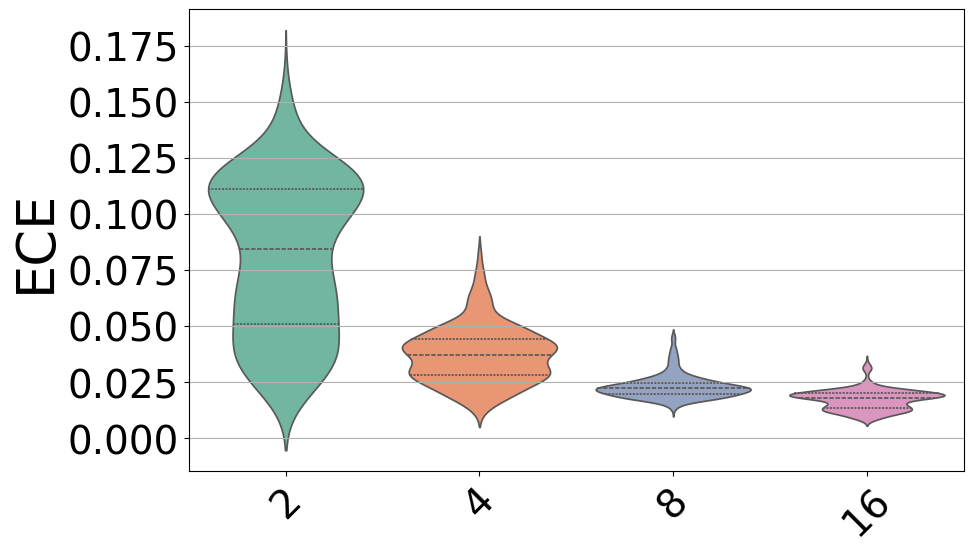}
    } &
    \subfigure[TSL(LS): Grid Range]{
        \includegraphics[width=0.18\linewidth]{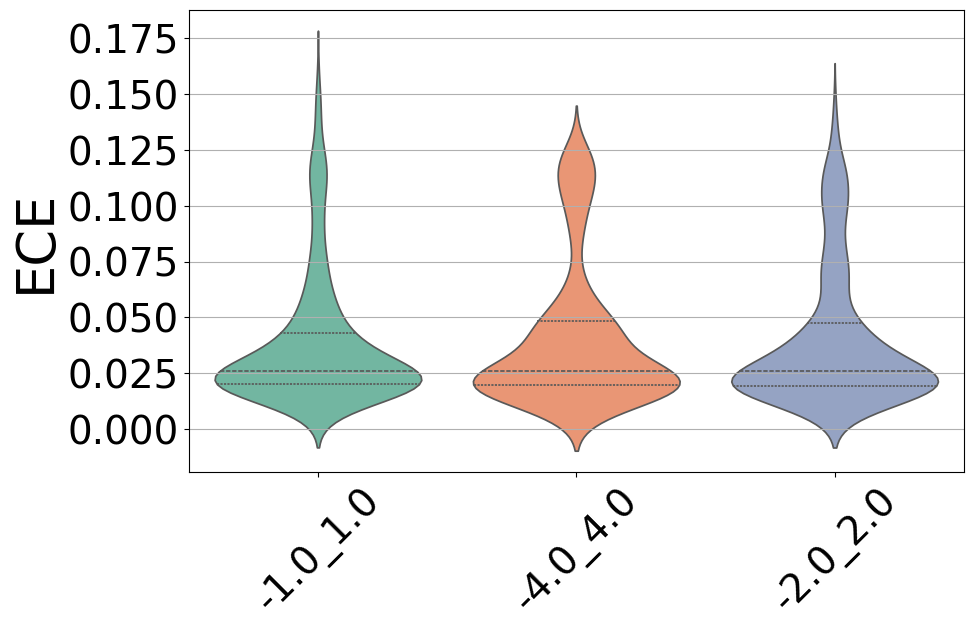}
    } &
    \subfigure[TSL(LS): Grid Order]{
        \includegraphics[width=0.18\linewidth]{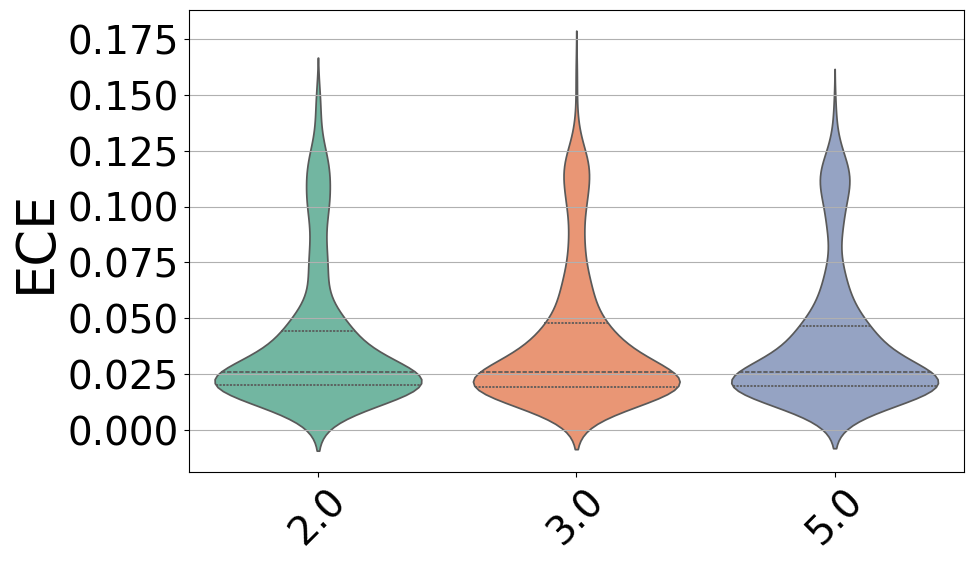}
    } &
    \subfigure[TSL(LS): Shortcut]{
        \includegraphics[width=0.18\linewidth]{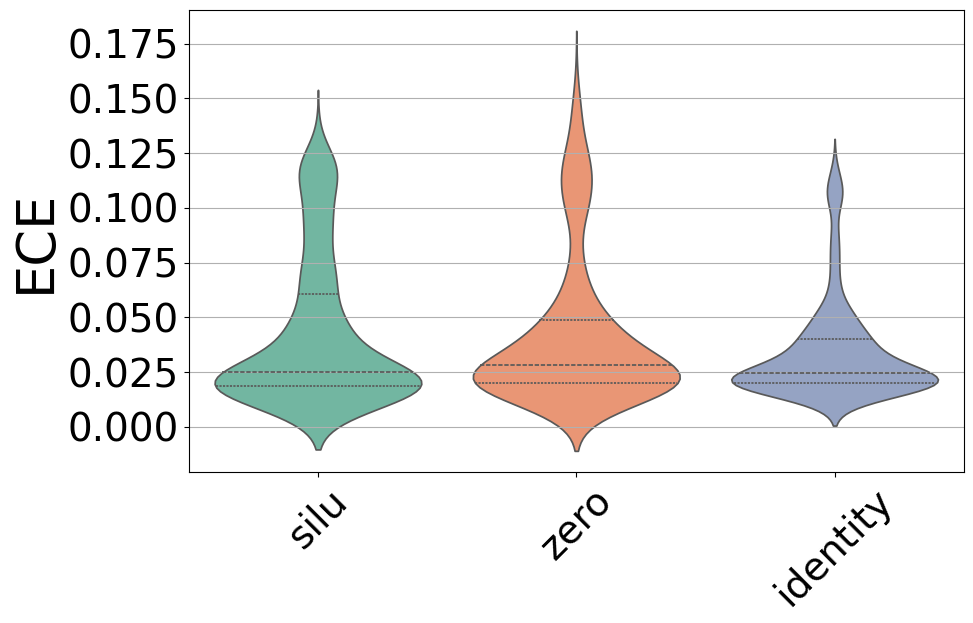}
    } &
    \subfigure[TSL(LS):Params]{
        \includegraphics[width=0.18\linewidth]{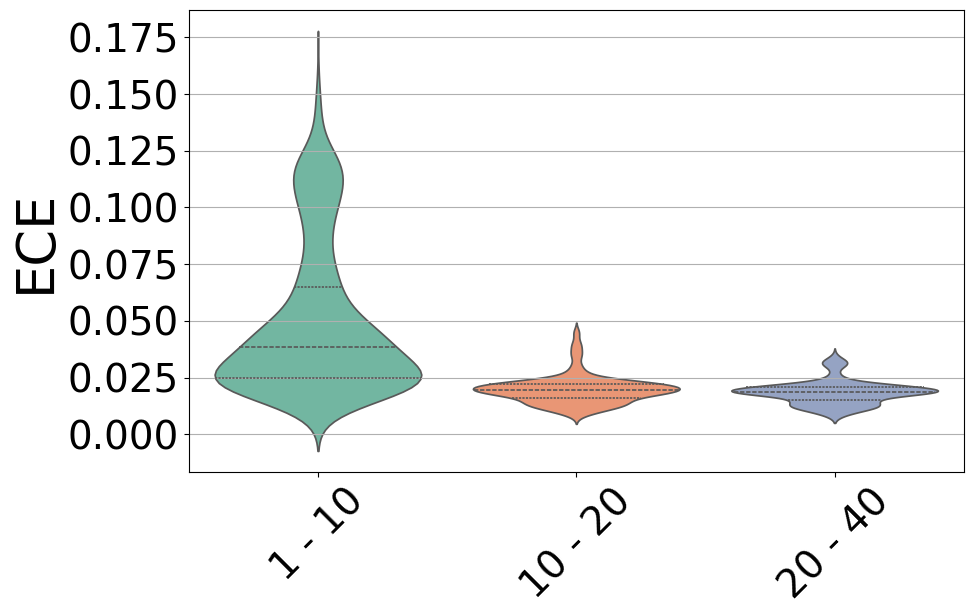}
    } \\
\end{tabular}
}
\caption{Comparison of KAN calibration metrics with and without Temperature-Scaled Loss (TSL). The top row shows Lable Soomth (LS) results, while the bottom row shows TSL applied to LS. Columns illustrate the effects of Layer Width, Grid Range, Grid Order, Shortcut, and Number of Parameters ($10^4$) on calibration performance.}
\Description{tbc}
\label{fig:multiple_factors_tsl_ls}
\end{center}
\vskip -0.2in
\end{figure*}

\begin{figure*}[ht]
\vskip 0.2in
\begin{center}
\resizebox{\textwidth}{!}{%
\begin{tabular}{ccccc}
    \subfigure[FL: Layer Width]{
        \includegraphics[width=0.18\linewidth]{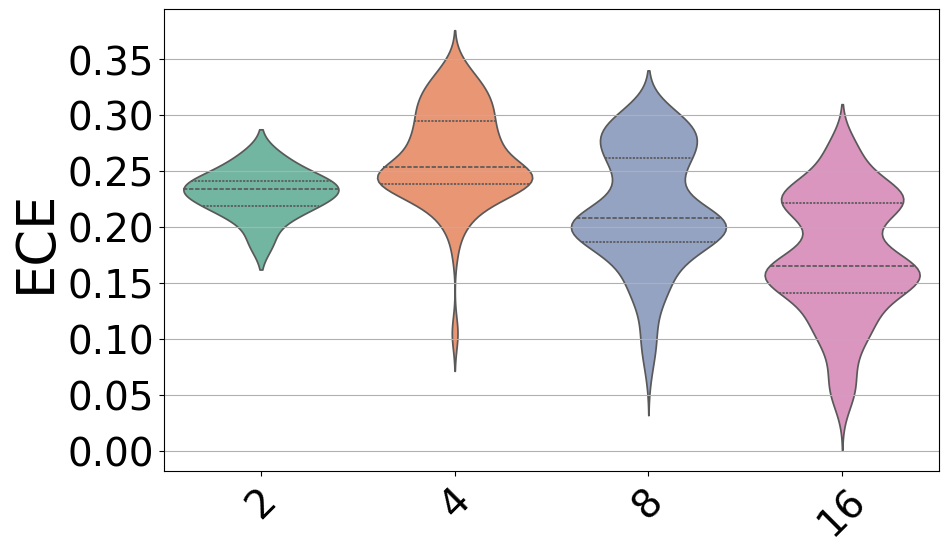}
    } &
    \subfigure[FL: Grid Range]{
        \includegraphics[width=0.18\linewidth]{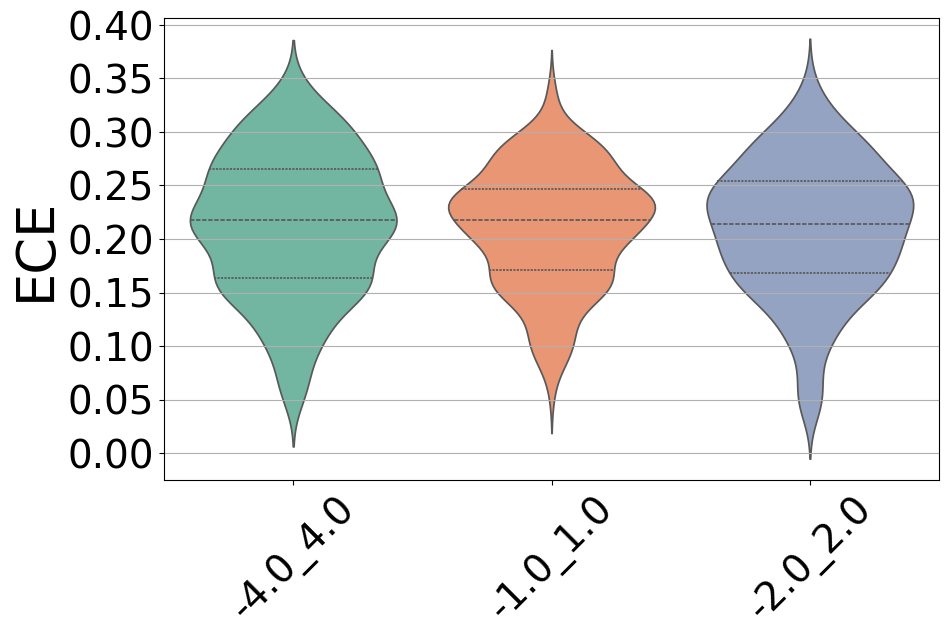}
    } &
    \subfigure[FL: Grid Order]{
        \includegraphics[width=0.18\linewidth]{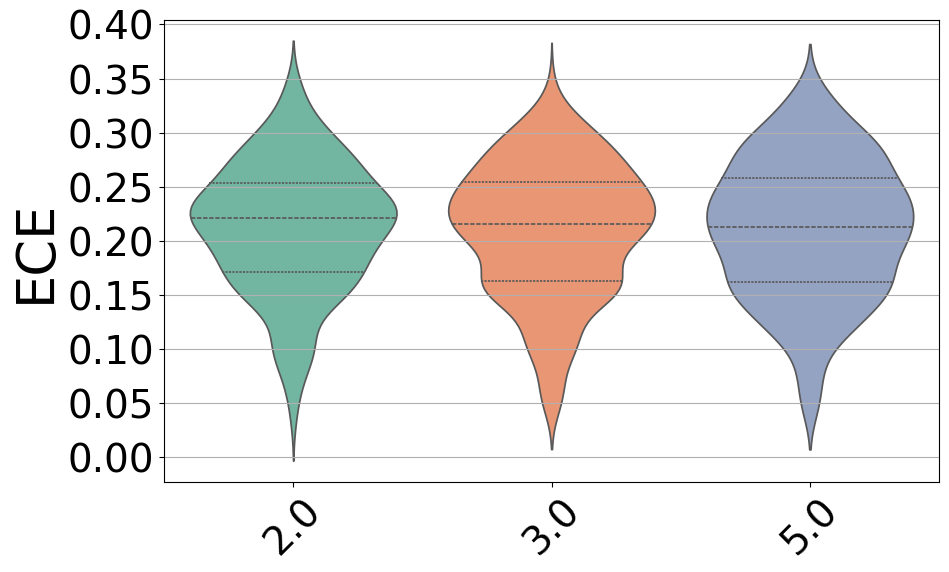}
    } &
    \subfigure[FL: Shortcut]{
        \includegraphics[width=0.18\linewidth]{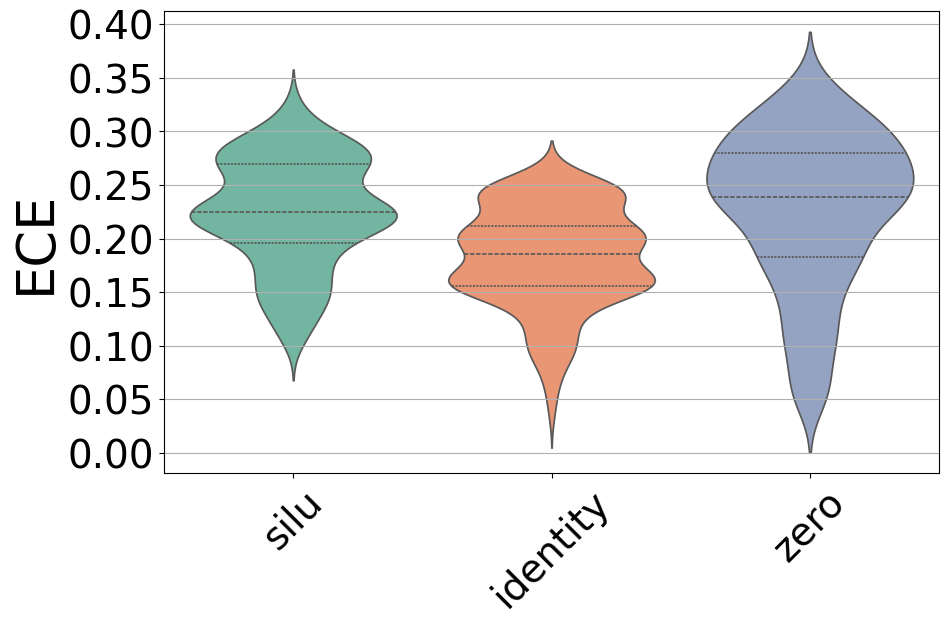}
    } &
    \subfigure[FL:Params]{
        \includegraphics[width=0.18\linewidth]{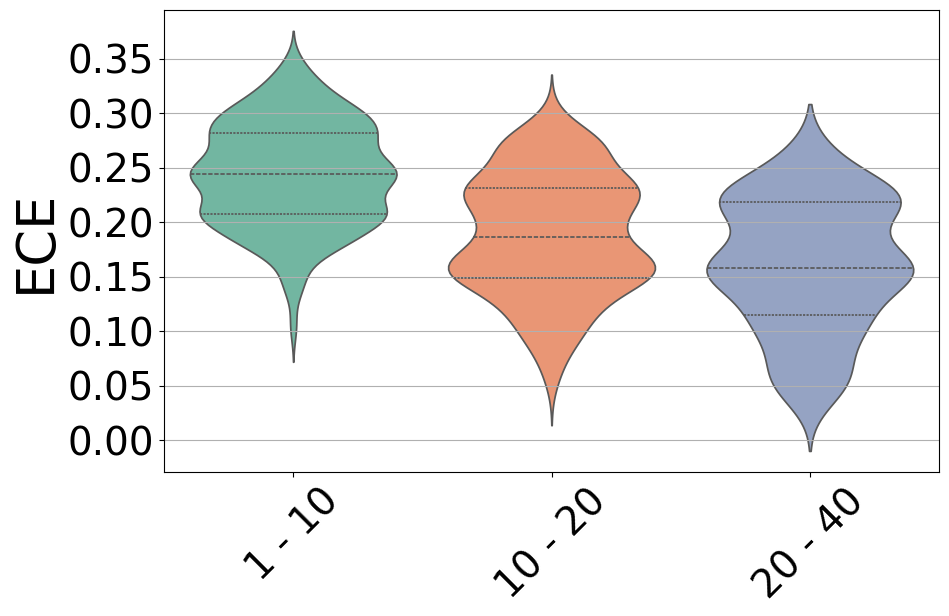}
    } \\
    \subfigure[TSL(FL): Layer Width]{
        \includegraphics[width=0.18\linewidth]{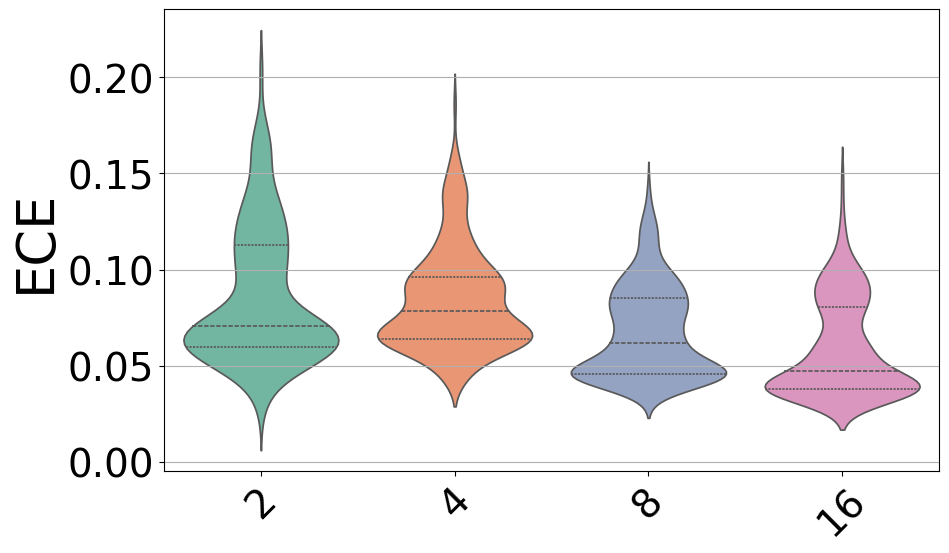}
    } &
    \subfigure[TSL(FL): Grid Range]{
        \includegraphics[width=0.18\linewidth]{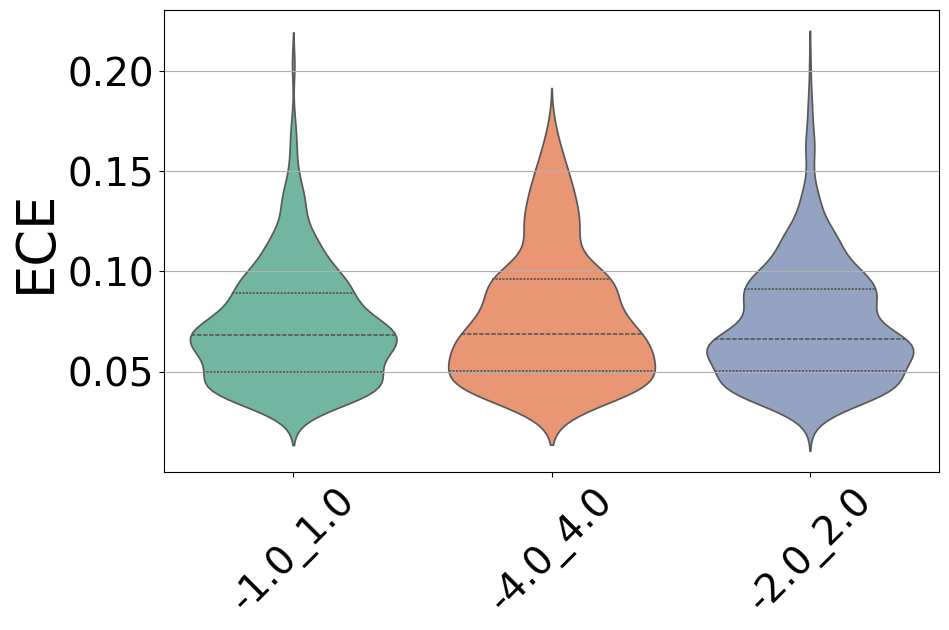}
    } &
    \subfigure[TSL(FL): Grid Order]{
        \includegraphics[width=0.18\linewidth]{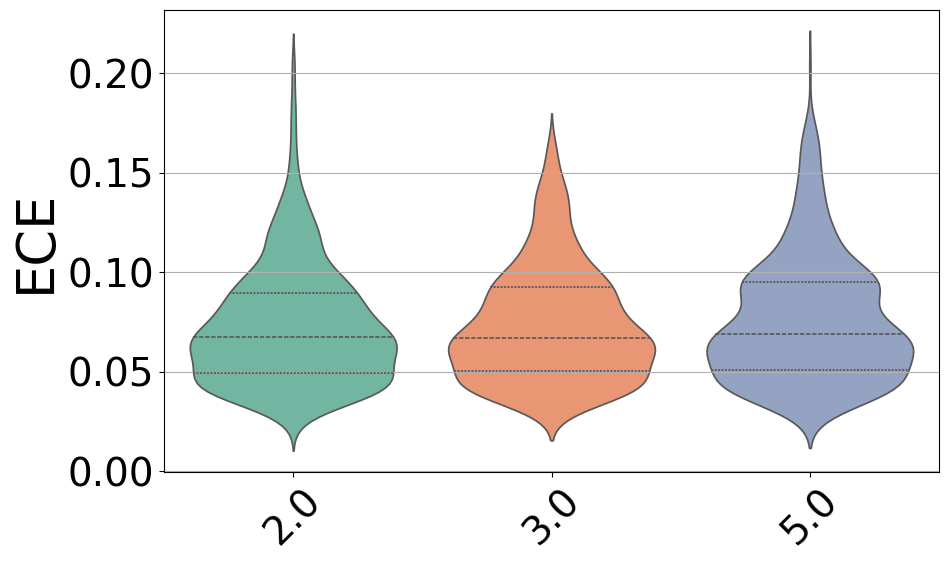}
    } &
    \subfigure[TSL(FL): Shortcut]{
        \includegraphics[width=0.18\linewidth]{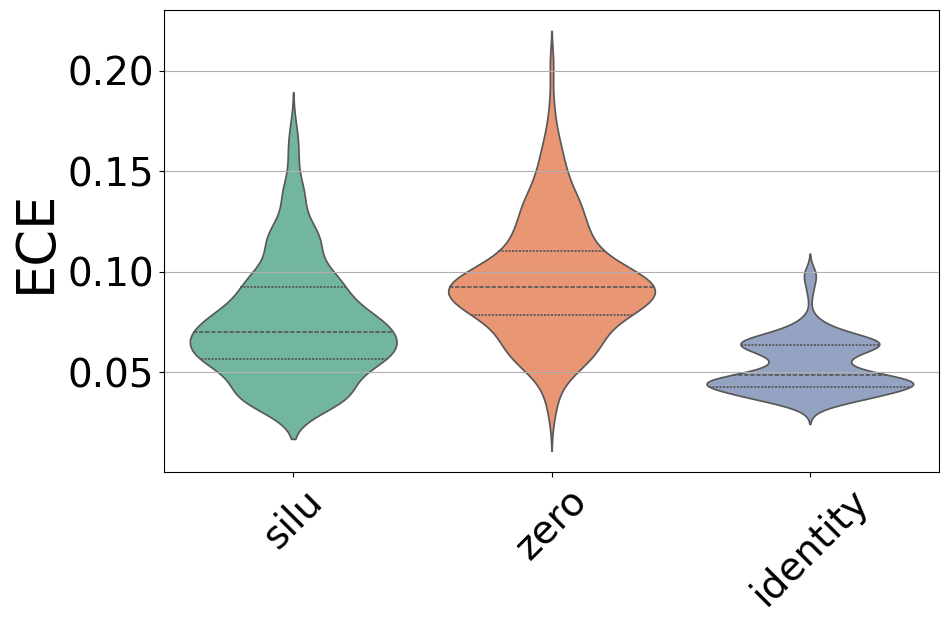}
    } &
    \subfigure[TSL(FL):Params]{
        \includegraphics[width=0.18\linewidth]{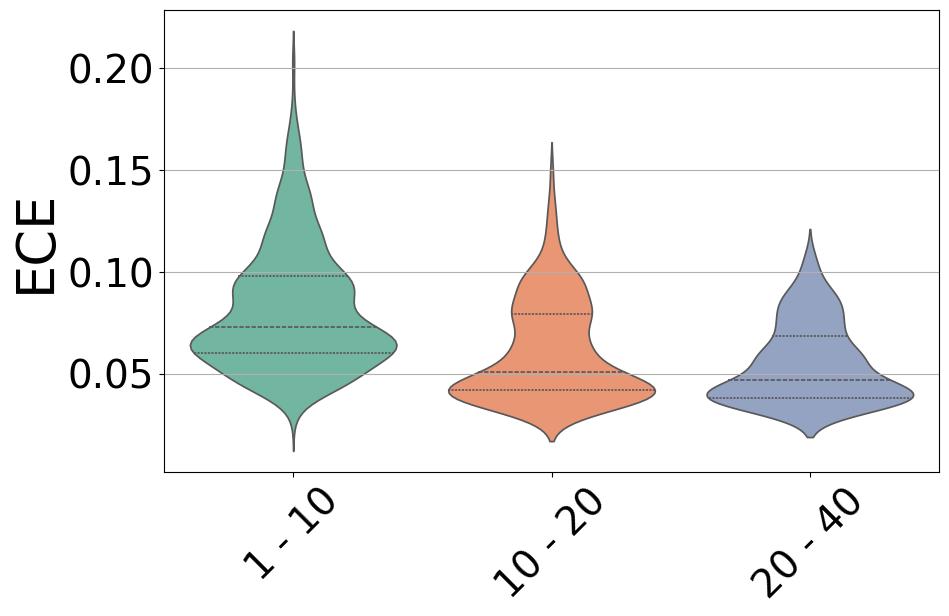}
    } \\
\end{tabular}
}
\caption{Comparison of KAN calibration metrics with and without Temperature-Scaled Loss (TSL). The top row shows Focal Loss (FL) results, while the bottom row shows TSL applied to FL. Columns illustrate the effects of Layer Width, Grid Range, Grid Order, Shortcut, and Number of Parameters ($10^4$) on calibration performance.}
\Description{tbc}
\label{fig:multiple_factors_tsl_fl}
\end{center}
\vskip -0.2in
\end{figure*}

\begin{figure*}[ht]
\vskip 0.2in
\begin{center}
\resizebox{\textwidth}{!}{%
\begin{tabular}{ccccc}
    \subfigure[DFL: Layer Width]{
        \includegraphics[width=0.18\linewidth]{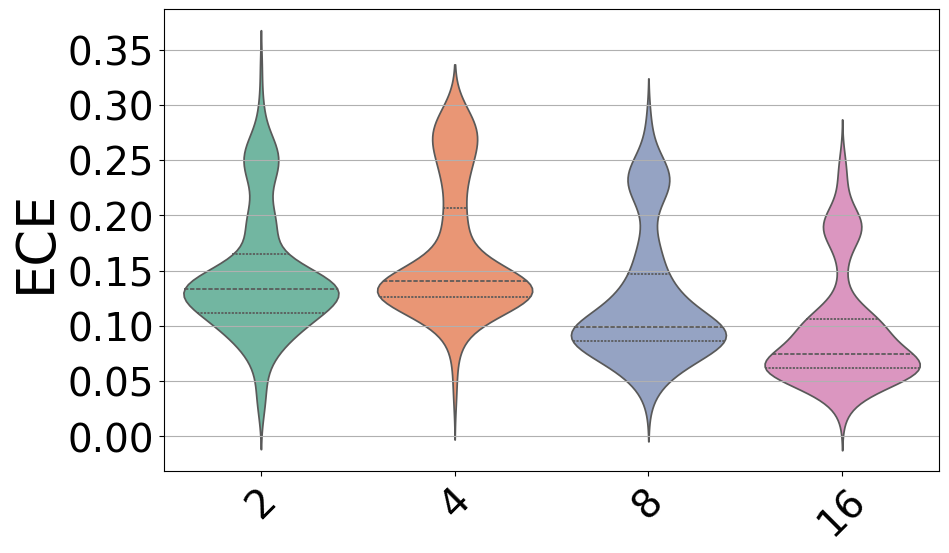}
    } &
    \subfigure[DFL: Grid Range]{
        \includegraphics[width=0.18\linewidth]{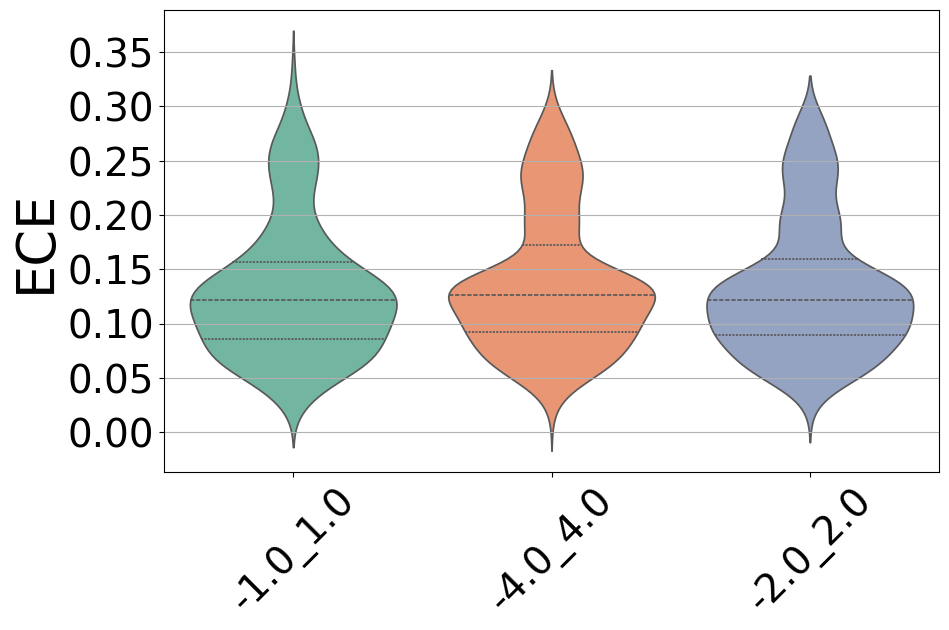}
    } &
    \subfigure[DFL: Grid Order]{
        \includegraphics[width=0.18\linewidth]{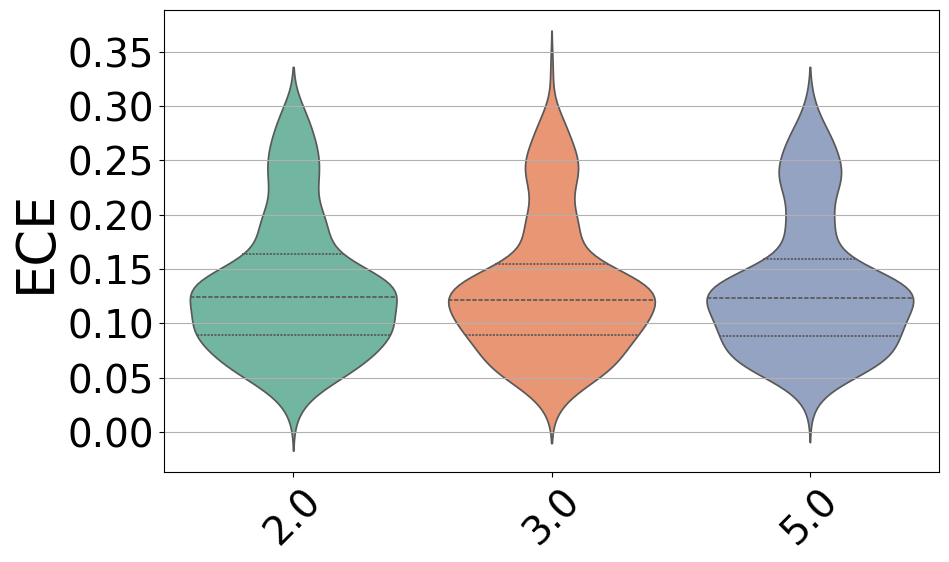}
    } &
    \subfigure[DFL: Shortcut]{
        \includegraphics[width=0.18\linewidth]{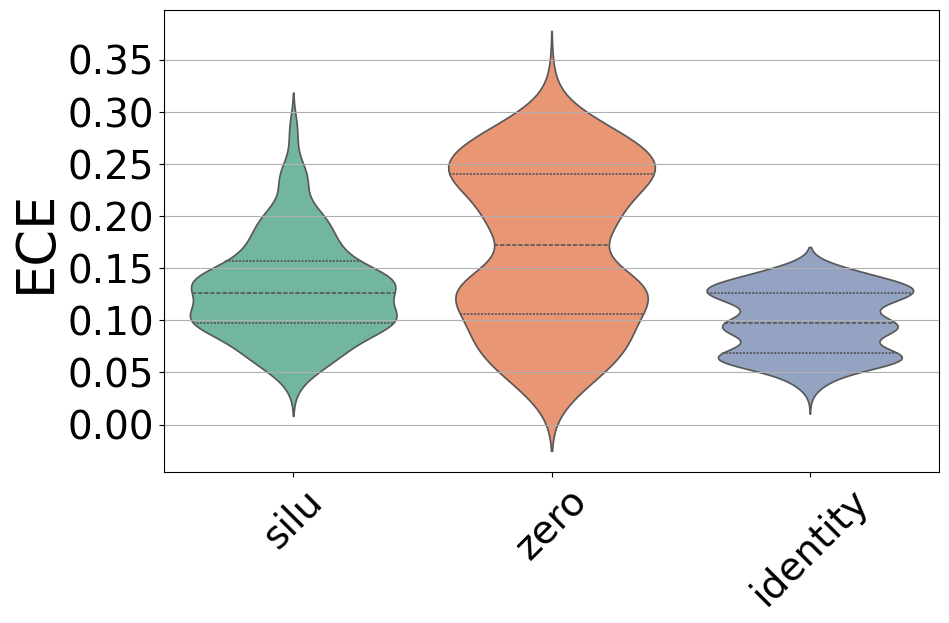}
    } &
    \subfigure[DFL:Params]{
        \includegraphics[width=0.18\linewidth]{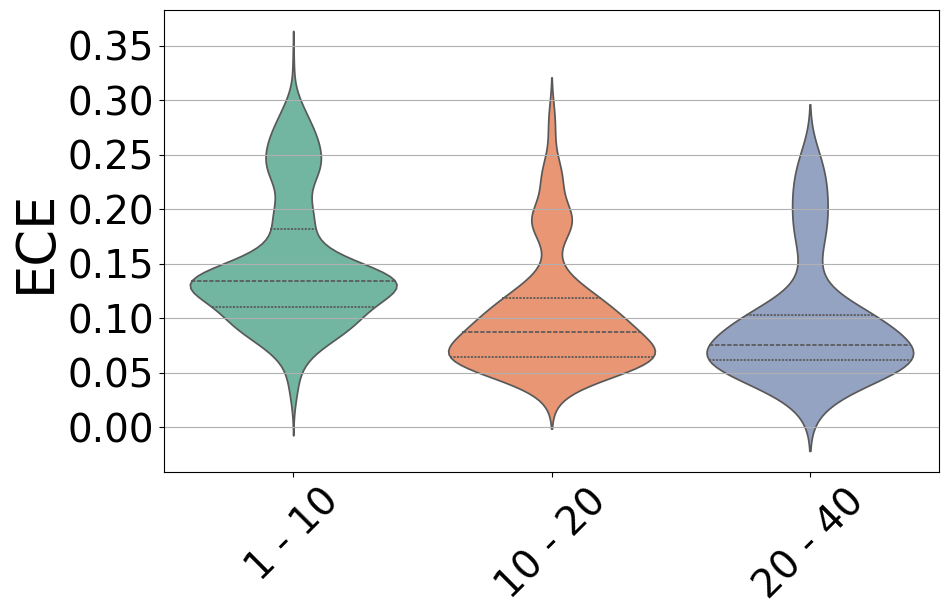}
    } \\
    \subfigure[TSL(DFL): Layer Width]{
        \includegraphics[width=0.18\linewidth]{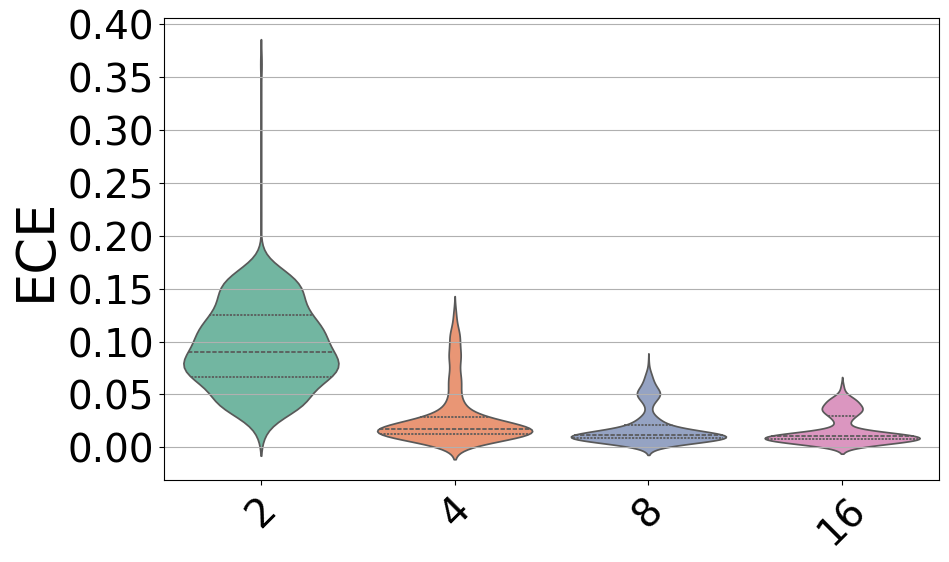}
    } &
    \subfigure[TSL(DFL): Grid Range]{
        \includegraphics[width=0.18\linewidth]{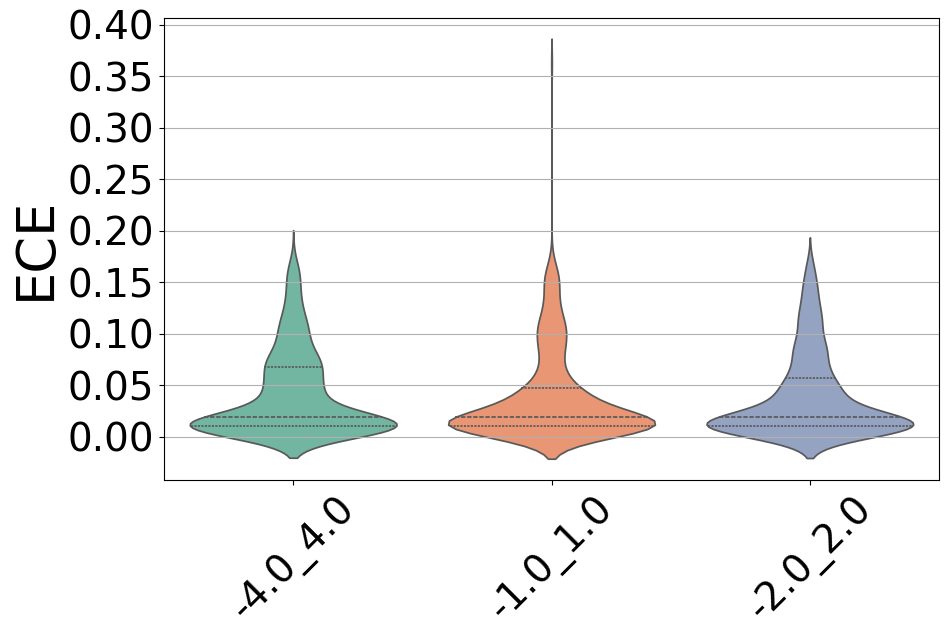}
    } &
    \subfigure[TSL(DFL): Grid Order]{
        \includegraphics[width=0.18\linewidth]{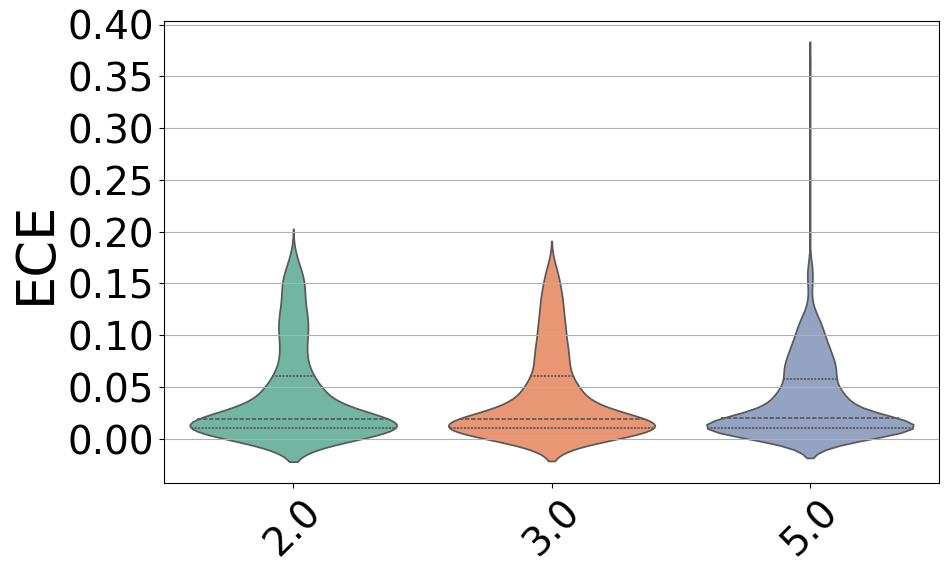}
    } &
    \subfigure[TSL(DFL): Shortcut]{
        \includegraphics[width=0.18\linewidth]{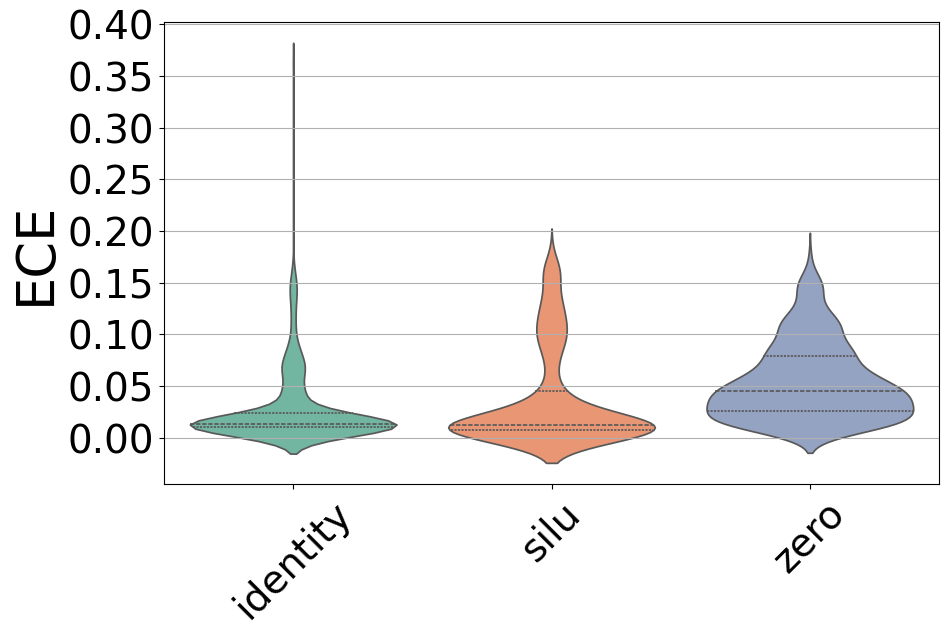}
    } &
    \subfigure[TSL(DFL):Params]{
        \includegraphics[width=0.18\linewidth]{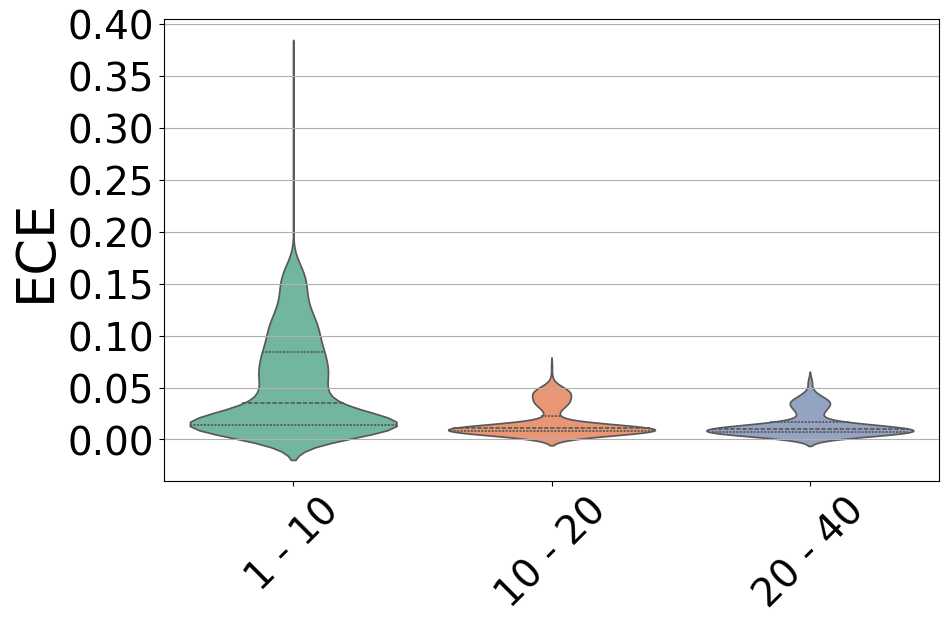}
    } \\
\end{tabular}
}
\caption{Comparison of KAN calibration metrics with and without Temperature-Scaled Loss (TSL). The top row shows Dual Focal Loss (DFL) results, while the bottom row shows TSL applied to DFL. Columns illustrate the effects of Layer Width, Grid Range, Grid Order, Shortcut, and Number of Parameters ($10^4$) on calibration performance.}
\Description{tbc}
\label{fig:multiple_factors_tsl_dfl}
\end{center}
\vskip -0.2in
\end{figure*}

\begin{figure*}[ht]
\vskip 0.2in
\begin{center}
\resizebox{\textwidth}{!}{%
\begin{tabular}{ccccc}
    \subfigure[FCL: Layer Width]{
        \includegraphics[width=0.18\linewidth]{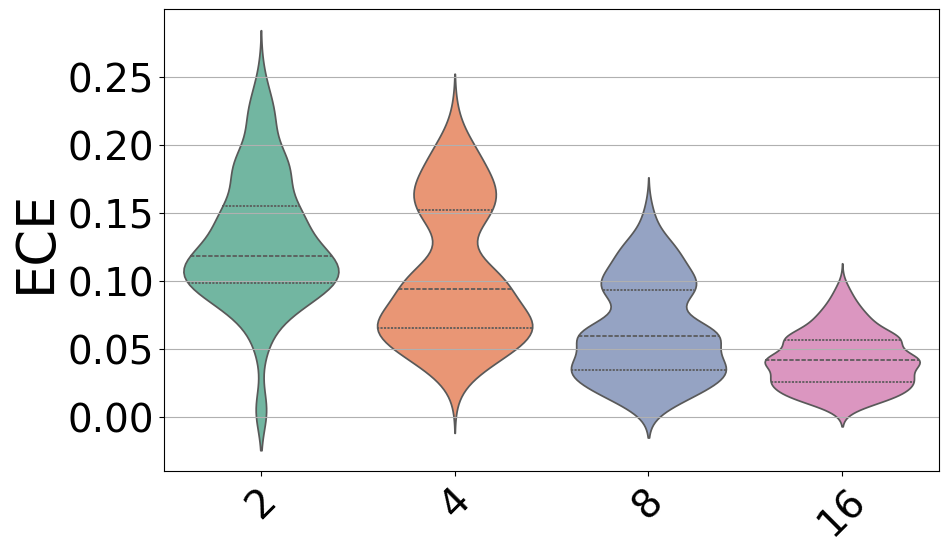}
    } &
    \subfigure[FCL: Grid Range]{
        \includegraphics[width=0.18\linewidth]{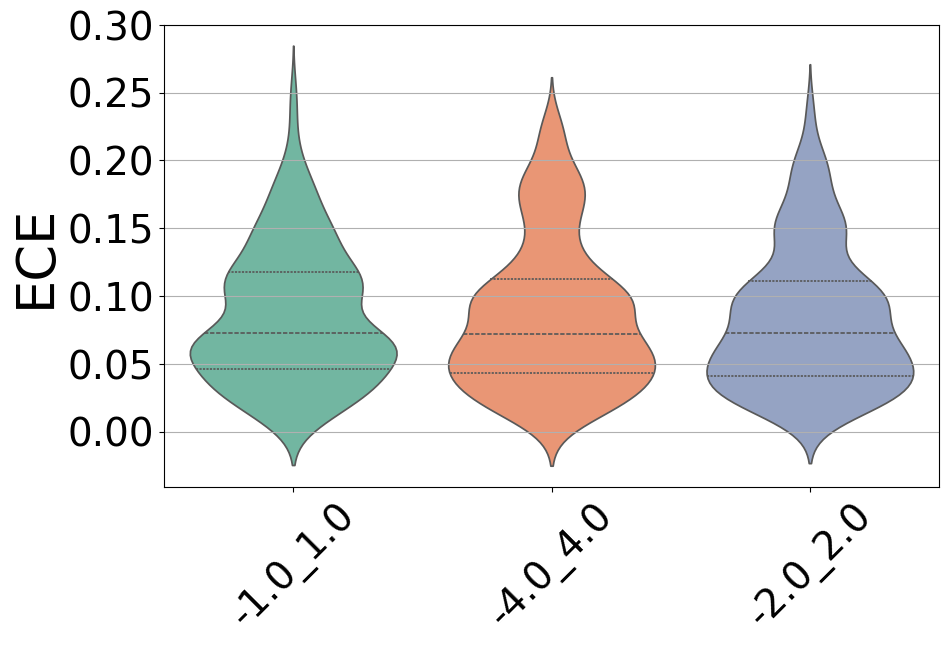}
    } &
    \subfigure[FCL: Grid Order]{
        \includegraphics[width=0.18\linewidth]{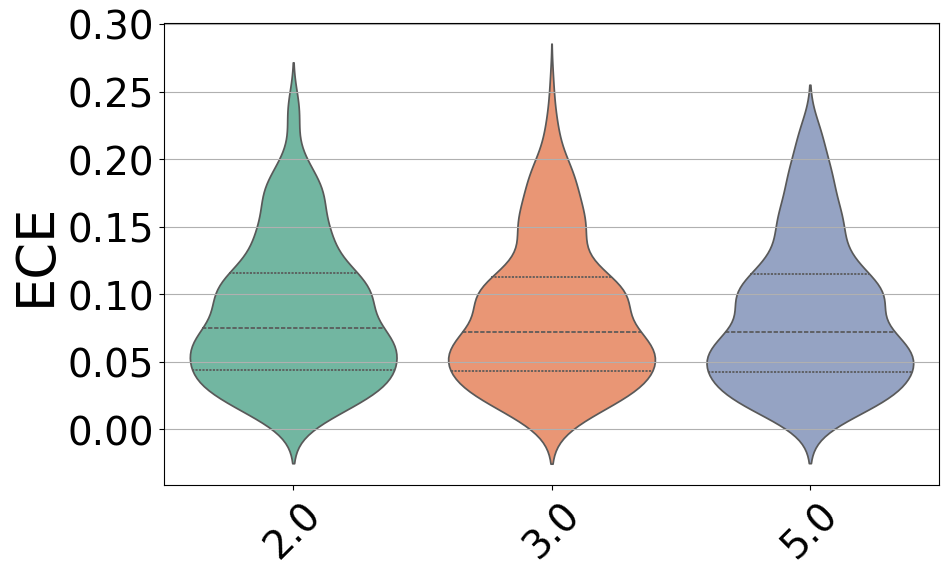}
    } &
    \subfigure[FCL: Shortcut]{
        \includegraphics[width=0.18\linewidth]{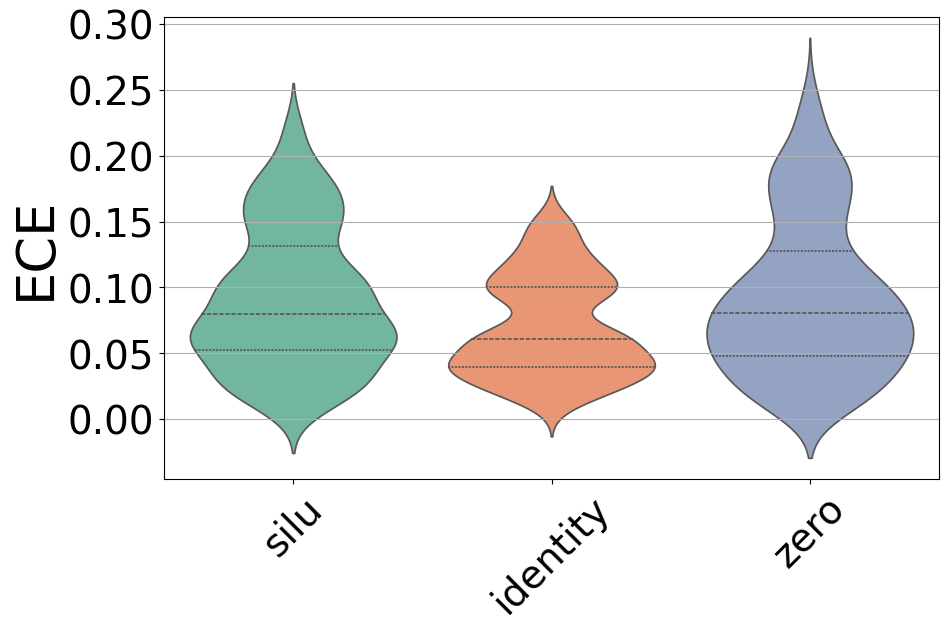}
    } &
    \subfigure[FCL:Params]{
        \includegraphics[width=0.18\linewidth]{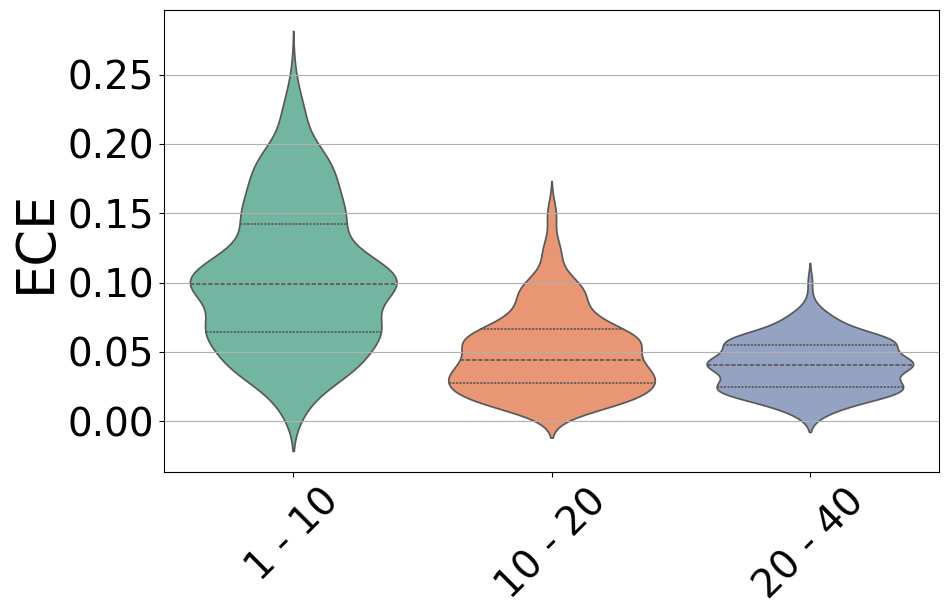}
    } \\
    \subfigure[TSL(FCL): Layer Width]{
        \includegraphics[width=0.18\linewidth]{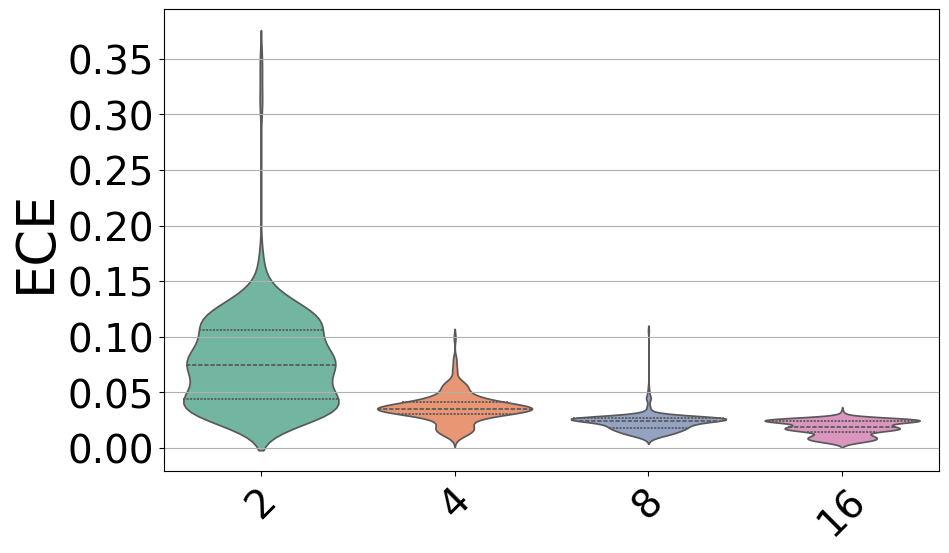}
    } &
    \subfigure[TSL(FCL): Grid Range]{
        \includegraphics[width=0.18\linewidth]{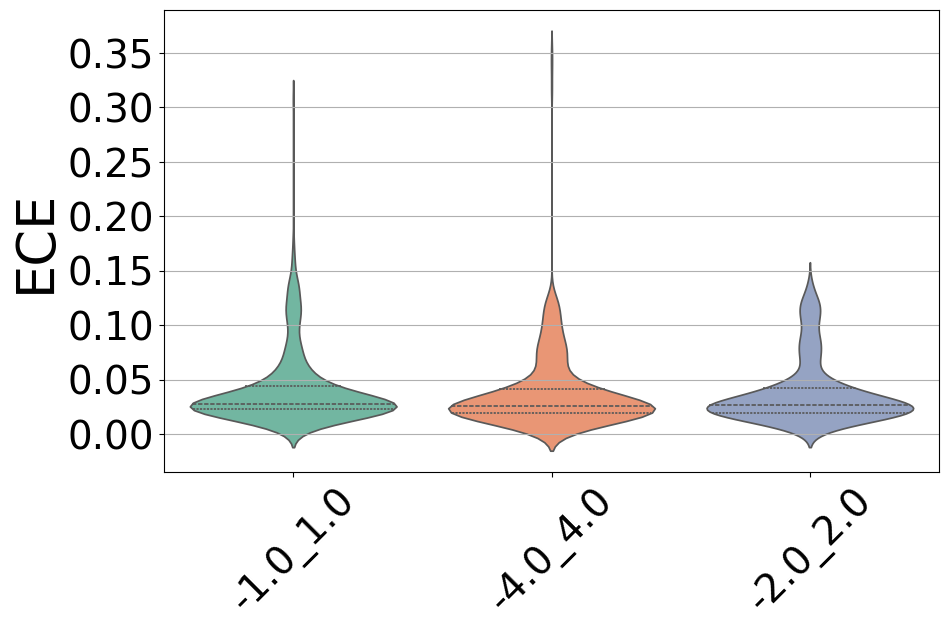}
    } &
    \subfigure[TSL(FCL): Grid Order]{
        \includegraphics[width=0.18\linewidth]{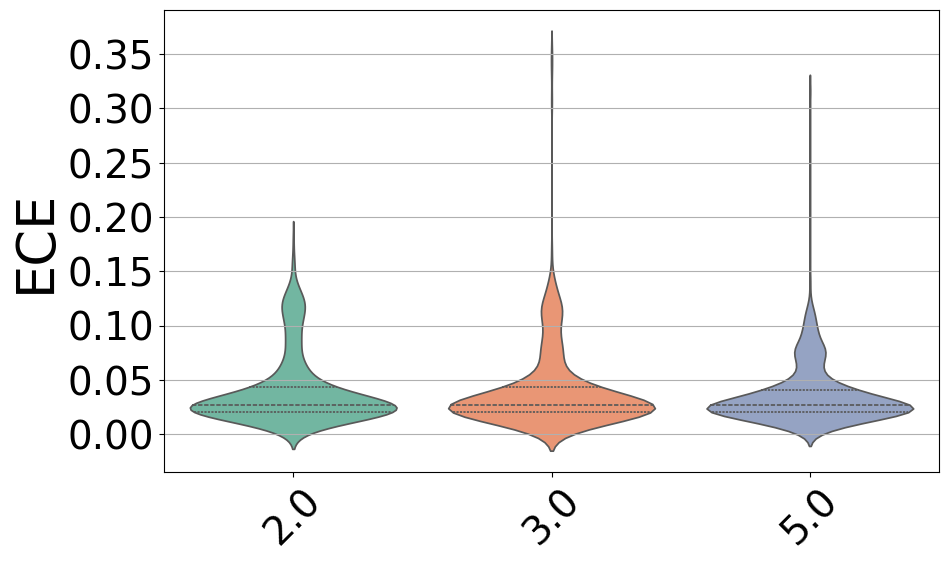}
    } &
    \subfigure[TSL(FCL): Shortcut]{
        \includegraphics[width=0.18\linewidth]{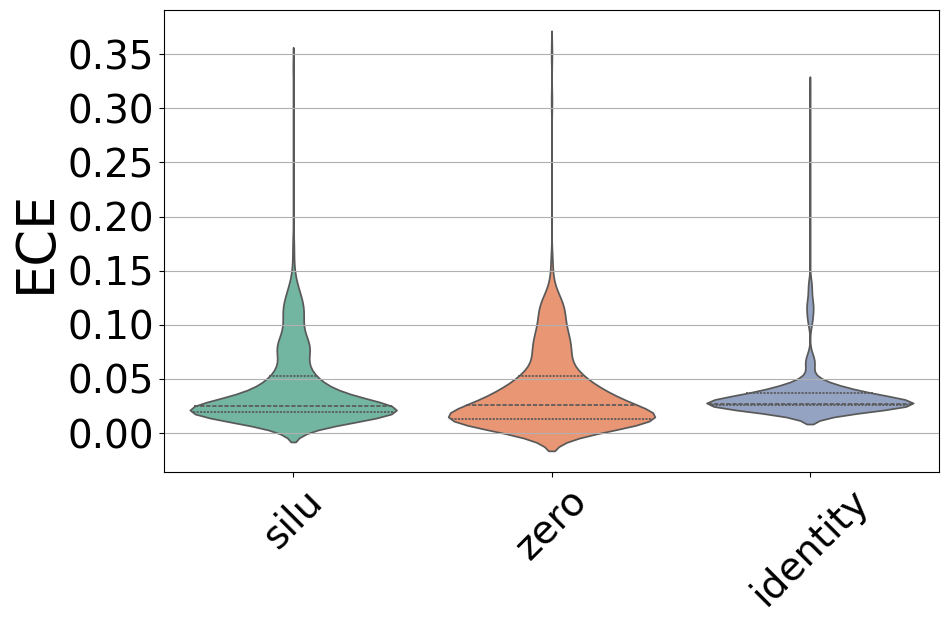}
    } &
    \subfigure[TSL(FCL):Params]{
        \includegraphics[width=0.18\linewidth]{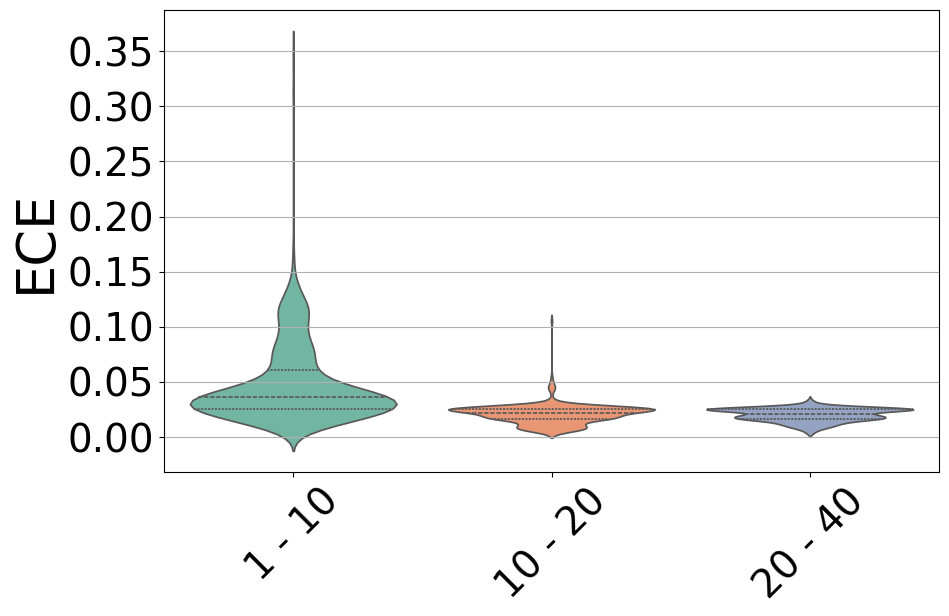}
    } \\
\end{tabular}
}
\caption{Comparison of KAN calibration metrics with and without Temperature-Scaled Loss (TSL). The top row shows Focal Calibration Loss (FCL) results, while the bottom row shows TSL applied to FCL. Columns illustrate the effects of Layer Width, Grid Range, Grid Order, Shortcut, and Number of Parameters ($10^4$) on calibration performance.}
\Description{tbc}
\label{fig:multiple_factors_tsl_fcl}
\end{center}
\vskip -0.2in
\end{figure*}

\end{document}